\documentclass[conference]{IEEEtran}
\IEEEoverridecommandlockouts
\usepackage{xurl}
\usepackage{cite}
\usepackage{amsmath,amssymb,amsfonts}
\usepackage{textcomp}
\usepackage{xcolor}
\usepackage{balance}

\usepackage{booktabs} % For professional-looking tables
\usepackage{makecell} % Crucial for multi-line cells
\usepackage{subcaption}
\usepackage{tcolorbox}
\usepackage{enumitem}
\usepackage{multirow}
\usepackage{tabularx}
\usepackage{algorithmicx}
\usepackage{algpseudocode}
\usepackage{algorithm}
\newcommand{\Func}[1]{\textsc{#1}}
\usepackage{graphicx}
\usepackage[inkscapelatex=false]{svg}
\usepackage[colorinlistoftodos]{todonotes}
\usepackage{tikz}\usetikzlibrary{shapes.geometric}
\usepackage{todonotes}
\usepackage{xspace}
\usepackage[skip=5pt]{caption}
\tcbset{
  colback=gray!3!white,
  colframe=black!20,
  left=1mm, right=1mm, top=1mm, bottom=1mm,
  boxrule=0.3pt,
  fonttitle=\bfseries,
  arc=2pt
}

\newcommand{\system}{\textsc{Opti-Q}\xspace}

\newcommand{\pseudosection}[1]{\vspace{1.4ex}\noindent \textbf{{#1}}~~}

\newcommand{\hide}[1]{}
\newif\ifextended
 \extendedfalse % VLDB version
\def\BibTeX{{\rm B\kern-.05em{\sc i\kern-.025em b}\kern-.08em
    T\kern-.1667em\lower.7ex\hbox{E}\kern-.125emX}}
\begin{document}

\title{Opti-Q: A Constraint-Based Optimization Framework for Multi-LLM Question Planning}

\author{
  Aamir Hamid$^1$, Bharg Barot$^1$, Satvik Racharla$^1$, Tim Finin$^1$, Primal Pappachan$^2$, and Roberto Yus$^1$ \\
  {\small \hspace{-0.2cm} $^1$University of Maryland, Baltimore County, USA. \quad $^2$Portland State University, USA.} \\
  {\small \texttt{\{ahamid2, bhargvb1, yv04378, finin, ryus\}@umbc.edu, primal@pdx.edu}}
}
\maketitle

\begin{abstract}
While large language models (LLMs) enable strong question answering (QA), budgeted deployment is complicated by nondeterminism and heterogeneous resource profiles (cost, latency, and energy). We present \system, a database-inspired, cost-based optimizer that implements a \emph{plan-before-execute} paradigm for multi-LLM orchestration. \system models LLM invocations as physical operators in an execution DAG and, for each question, searches for plans that optimize answer quality (QoA) while trading off financial cost, latency, and energy under user-specified resource constraints. Plans can include sequential operators that pass intermediate answers as context and parallel/blend operators that run models concurrently and merge their outputs. To search this space without executing each candidate plan, \system uses \textsc{PerfDB}, a statistics catalog populated and refreshed from benchmarks and execution traces, to estimate the QoA and resource costs of both individual operators and composed subplans. Using these estimates, \system performs Pareto-frontier search and selects a final plan based on user preferences. On MMLU-Pro and SimpleQA under user-specified budgets, \system improves average QoA by $\approx 58\%$ and $\approx 41\%$ over baselines at comparable cost, demonstrating that database-style planning yields better quality--resource trade-offs for multi-LLM QA.
\end{abstract}

\begin{IEEEkeywords}
Large language models, multi-LLM systems, query planning, multi-objective optimization.
\end{IEEEkeywords}

\section{Introduction}
\label{sect:intro}
Large Language Models (LLMs) such as GPT-5~\cite{openai2025gpt5}, Llama~3~\cite{grattafiori2024llama}, and DeepSeek-R1~\cite{deepseekai2025deepseekr1incentivizingreasoningcapability} have transformed natural language processing (NLP)~\cite{devlin2019bertpretrainingdeepbidirectional}, driving advances in question-answering (QA)~\cite{chen2017readingwikipediaansweropendomain}, text summarization~\cite{lewis2021retrievalaugmentedgenerationknowledgeintensivenlp} and domain-specific reasoning~\cite{karpukhin2020densepassageretrievalopendomain}. Their success has fueled adoption across industries, from customer service to healthcare~\cite{8649787,Bickmore2005EstablishingAM}. However, real-world deployment remains challenging due to nondeterminism, leading to hallucinations~\cite{hamid2023genaipabench}, high computational costs~\cite{cottier2024rising}, substantial energy consumption~\cite{strubell-etal-2019-energy}, and  latency trade-offs~\cite{patterson2022carbonfootprintmachinelearning}. Considerable effort has focused on these issues, including increasing accuracy and reliability through advanced prompting strategies (e.g., Chain-of-Thought prompting) or extensive fine-tuning~\cite{wei2022chain}. However, a fundamental challenge persists: different LLMs exhibit varying performance across question/task types~\cite{wei2024measuringshortformfactualitylarge}.  
% For example, OpenAI's GPT-5 achieves a top score on the American Invitational Mathematics Examination benchmark (100\% accuracy)~\cite{openai2025gpt5}, whereas Google DeepMind’s Gemini~3.1~Pro reports strong coding performance (LiveCodeBench Pro: Elo 2887; SWE-bench Verified: 80.6\%) and achieves 44.4\% on Humanity's Last Exam (no tools)~\cite{deepmind2026gemini31pro}.
Moreover, recent evidence suggests that a single high-capability LLM is not always optimal; coordinated multi-LLM collaboration can outperform the single ``best'' model on real-world tasks~\cite{feng2025llmdroolsmultillmcollaboration}.

Inspired by machine learning (ML) ensembles~\cite{10.5555/648054.743935}, recent work explores combining multiple LLMs to improve reliability and answer quality~\cite{amatriain2024promptdesignengineeringintroduction, suzuoki2024reducing}, as cross-model validation helps mitigate hallucinations and raise response quality~\cite{dey2025uncertainty}. However, naive execution strategies (e.g., always querying all models and combining (\textit{blending})~\cite{jiang2023llmblenderensemblinglargelanguage} their outputs, or always choosing a fixed ``strong'' subset of models) introduce new challenges: higher financial costs and latency, inefficient resource use, and sometimes lower answer quality. Moreover, while many systems rely on dynamic sequential orchestration, where later model choices depend on earlier model outputs (e.g., via LangChain-style chains)~\cite{LangChainDocs} or execution-time optimization for cost/throughput~\cite{palimpzestCIDR}. Such decisions can be myopic when made without planning for downstream consequences. For example, in our experiments, a history question was best answered by a plan using Gemma-3:27B, Qwen-2.5:14B, and Phi-4:14B (outperforming other plans by a 1.24x  to 1.48× factor), whereas a sports question was best handled by a simple parallel execution of Gemma-3:27B and Qwen-2.5:14B. These variations show that the optimal model combination and execution strategy are question-dependent. A robust multi-LLM QA system must therefore determine the best execution plan dynamically for each question, while accounting for end-to-end cost, latency, energy, and quality.

% For example, selecting a cheap model first can yield low-quality or verbose intermediate outputs that force later escalation, increase downstream token usage, and ultimately worsen end-to-end cost, latency, and energy compared to a globally optimized plan. 
Database systems face a similar problem where the same declarative query can be executed by many alternative physical plans, and the optimizer uses a cost model and statistics to select a plan before execution~\cite{10.1145/3448016.3452838}.
In Multi-LLM QA, many alternative workflows may be possible for a user question, each trading off monetary cost, latency, energy, and answer quality.
We therefore cast multi-LLM orchestration as a cost-based (and multi-objective) query-planning problem, where the question serves as the query, each LLM invocation is a physical operator, and an end-to-end workflow is a physical plan whose selection is guided by a model ``catalog'' of historical performance statistics.
Unlike classical query optimization, multi-LLM planning must account for stochastic and semantic operator outputs (quality is uncertain and non-monotonic), output-dependent downstream costs (e.g., token lengths and content can amplify later latency/cost/energy), and inherently multi-dimensional objectives (quality–cost–latency–energy) under user constraints. This necessitates new cost estimation models and multi-objective plan selection beyond traditional single-metric cost minimization.

To tackle these challenges, we introduce \system, a cost-based optimizer for multi-LLM QA that selects an execution plan \emph{before} issuing any model calls, mirroring how DBMS optimizers choose physical query plans prior to execution. As illustrated in Figure \ref{fig:Arch_one}, the plans generated by \system involve combinations of parallel and sequential multi-LLM operations. 
% A plan specifies (a) which models to invoke and (b) how to compose them using these operations. A sequential plan (e.g., Plan~1) composes LLMs into a pipeline, where the output of one model ($L_1$) serves as a context for a subsequent model ($L_2$) for refinement or verification. In contrast, a parallel plan (e.g., Plan~n) executes multiple models ($L_3, L_4$) concurrently, aggregating their output \hide{independent} to create a single high-fidelity answer.
To select among candidate plans, we propose a technique to estimate, before execution, a plan's cost (financial, latency, energy) and benefit (answer quality) from historical performance statistics on LLMs and their combinations across question types. Using these estimates, \system identifies plans representing optimal trade-offs among competing objectives under optional user constraints via multi-objective optimization (MOO). Given the potentially large plan space, which depends on the number of LLMs considered and the complexity of the plan structure, \system employs a ``pluggable optimization engine'' that supports diverse planning strategies, including dynamic programming (DP), Hill Climbing, and Non-dominated Sorting Genetic Algorithm (NSGA-II) to explore candidate plans and select one for execution. This enables \system to dynamically balance performance and resource efficiency before committing to costly LLM invocations.
The main contributions of this work are:
\raggedbottom
\begin{itemize}
    \item A cost/benefit formulation of multi-LLM QA planning to select a workflow under budget and resource constraints.
    \item A statistics-driven optimizer that enumerates and prunes sequential/parallel/hybrid workflows and estimates quality and resource costs prior to execution.
    \item A system that integrates the optimizer with an execution engine for real-time routing across open-source LLMs.
\end{itemize}

We evaluate our approach on two QA benchmarks, one open-ended and one multiple-choice, and compare it against four state-of-the-art baselines.
In both, \system achieves superior QoA-cost efficiency, improving the average QoA by $\approx 58\%$ and $\approx 41\%$ over the strongest budget-aware baseline at matched per-question cost (budget level b=3). We further analyze the impact of LLM diversity, plan complexity, and budget levels on the \hide{quality--cost--latency--energy} trade-offs \system discovers.

\section{Related Work}
\label{sect:related-work}

Recent work at the intersection of data management and LLMs spans two largely orthogonal directions. The first studies how LLMs can assist database query optimization (e.g., improving cardinality estimation~\cite{10.14778/3503585.3503586}, cost modeling~\cite{yao2025query}, or physical operator selection~\cite{10.14778/3583140.3583160}). This topic focuses on optimizing database queries over structured data which differs from our setting, where the challenge is to select and compose multiple LLM calls to answer a natural-language question under quality–cost–latency–energy trade-offs.
The second direction is more related to our work and treats LLM calls as expensive, probabilistic operators in query-like workflows, asking how to optimize these AI-powered workflows using system and optimizer principles. Orthogonal to this, orchestration frameworks such as LangChain\footnote{\url{https://github.com/langchain-ai/langchain}}
 and DSPy~\cite{khattab2024dspy} support multi-step flows but leave plan structure and physical choices largely developer-scripted, making optimization implicit.

Early approaches to LLM call orchestration adopt \emph{cascading}, routing requests through models of increasing capacity. FrugalGPT~\cite{chen2023frugalgptuselargelanguage} typifies this paradigm by learning a budget-aware cascade that reduces cost while maintaining accuracy. However, cascades often impose a largely fixed structure and can accumulate sequential latency; moreover, their intermediate generations can inflate downstream token usage (and thus cost/latency), making greedy ``start-cheap'' strategies globally suboptimal.
Other ensemble methods, such as LLM-Blender~\cite{jiang2023llmblenderensemblinglargelanguage} and LLM-TOPLA~\cite{tekin2024llmtoplaefficientllmensemble}, improve quality by fusing multiple outputs, often assuming that additional inference cost is acceptable. 
Building on these insights, adaptive \emph{question-time} orchestration chooses models conditioned on the input. ThriftLLM~\cite{10.14778/3749646.3749702} learns per-question ensemble selection within a budget, while Shekhar et al.~\cite{shekhar2024optimizingcostsllmusage} predict quality pre-inference to choose cost-effective models under latency constraints for summarization. These methods align computation with question difficulty, but they typically (i) optimize a single dominant objective and/or (ii) treat model invocations as largely independent decisions rather than as a structured plan with interactions among heterogeneous LLMs.

Complementary recent work formalizes LLM calls as operators in declarative data systems, enabling optimizer-style reasoning over AI pipelines. Galois~\cite{Saeed2023QueryingL} emphasizes logical optimizations, such as pushing down filters to reduce expensive LLM calls, while frameworks such as PALIMPZEST~\cite{palimpzestCIDR}, Abacus~\cite{Russo2025AbacusAC}, and Stretto~\cite{sanmartino2026strettoexecutionenginellmaugmented} develop cost-based optimization, constrained objectives, and execution-time trade-offs. 
% Benchmarks such as SemBench~\cite{zubillaga2026sembenchuniversalsemanticframework} similarly motivate this systems direction by evaluating semantic query-processing engines across scenarios and modalities.
\system differs from these declarative systems along two key architectural dimensions. First, prior systems primarily optimize data-processing pipelines over records, documents, or multimodal corpora by selecting physical implementations for a given declarative program or workflow. \system instead shifts the optimization boundary to online, per-question planning, dynamically generating and selecting sequential, parallel, or hybrid execution graphs tailored to an individual incoming question. Second, while several existing frameworks handle resource-quality trade-offs through scalarized objectives or fixed workflow templates, \system treats QoA, cost, latency, and energy as distinct first-class objectives and explicitly explores Pareto trade-offs at question time to support user-specified priorities and resource constraints.

%We need to cut something so I'm cutting this part out
% \textcolor{blue}{Conceptually, \system follows a classical cost-based query optimization by enumerating alternative physical plans, estimating their behavior using statistics, and selecting plans that satisfy resource constraints~\cite{10.1145/3448016.3452838}. The database community has extensively studied multi-objective query optimization (MOQO), developing randomized and approximation techniques to compute Pareto-optimal frontiers, particularly in cloud environments mapping runtime against monetary cost~\cite{trummer2014approximation,trummer2016fast}. Multi-LLM planning departs from traditional MOQO in that candidate plans are not semantics-preserving in the database sense: answer quality is an explicit, stochastic objective variable that depends directly on output characteristics (e.g., intermediate verbosity impacting downstream cost). \system translates traditional MOQO principles to this non-deterministic setting by combining historical trace statistics with multi-objective optimization algorithms to plan under uncertainty in the stochastic multi-LLM execution space.}

% \input{sections/ourapproach}
\section{Opti-Q Multi-LLM Planning}
\label{sect:overview}
This section formalizes user questions, LLMs, and multi-LLM question plans. Then, it postulates the selection of an optimal plan for a given question under QoA, financial, latency, and energy constraints as a multi-objective optimization (MOO) problem. Finally, it sketches our approach. 
\ifextended
Frequently used notation appears in Table~\ref{table:notation_table}.
\else
% Frequently used notation appears in the extended version of this work~\cite{}.
\fi

% \vspace{-.3cm}
\ifextended
\begin{table}[!htb]
\centering
\caption{Frequently used notation.}

\label{table:notation_table}
\renewcommand{\arraystretch}{1.2}  % Adjust row spacing for better readability
\setlength{\tabcolsep}{8pt}  % Increase column spacing (default is ~6pt)
\footnotesize  % Reduce font size slightly to maintain compactness
\begin{tabular}{p{2.6cm} p{4.9cm}}  % Increase column widths for better spacing
\toprule
\textbf{Symbol} & \textbf{Definition} \\
\midrule
$Q$ = ($p_t$, $T$, $F_{\max}$, $L_{\max}$, $E_{\max}$, $QoA_{\min}$, $W$) & User question; its textual content $p_t$; topic(s) $T$; cost, latency, energy, and QoA constraints $F_{\max}, L_{\max}, E_{\max}, QoA_{\min}$; and weights per constraint $W$. \\
\hline
$L = \{L_1, \dots, L_n\}$ & Set of available LLMs. \\
\hline
$\mathit{Tok}_{L_i}(\cdot), T^{\mathrm{out}}_i$ & 
Input and output token lengths associated with model $L_i$. 
$\mathit{Tok}_{L_i}(\cdot)$ is the token-counting function using the tokenizer of $L_i$, while $T^{\mathrm{out}}_i$ is the model’s average output length. \\
\hline
$L_i = (f_i, l_i, e_i, \mathit{qoa}_i)$ 
& Model descriptor; placeholders for expected financial cost (USD/token), latency (s/token), energy cost (J/token), and QoA estimate used during planning. These quantities are instantiated from PerfDB based on the execution context, including question type, operator, prompt configuration, and model. \\
\hline

$f_i^{fix}, f_i^{var}$ & Fixed (per-invocation) cost and variable (per-token) cost of invoking $L_i$. \\
\hline

$\xi$ & Execution context (conditioning key for \textsc{PerfDB}); captures the conditions (e.g., topic, operator type, model) under which QoA/cost/latency/energy estimates are valid. \\
\hline

$\pi = (\{op_1, \dots, op_m\}, \prec$) & Question plan; its operations ($op_i$ is an LLM call or composition) and their partial order defining dependencies $\prec$ . \\
\hline
{$T_{op_i}^{out}, \; T_{op_i}^{proc}$} 
& Number of tokens generated, and processed by an operation.  \\
\hline
$\text{Seq}(Q, L_1, \dots,L_k)$ & Sequential operation: $L_2$ takes $L_1$'s answer to $Q$, $Q$ itself, and a context prompt $pt_{ctx}$ that instructs it on what to do.\\
\hline
$\text{Par}(Q, \{L_1, \dots, L_k\})$ & Parallel operation: each model process $Q$ independently.\\
\hline
$\text{Blend}(\text{Q, Par}(\dots), L_b)$ & Blending operation: model $L_b$ combines outputs from a parallel execution, guided by blending prompt $pt_{bld}$. \\
\hline
$Financial_{\pi}$, $Latency_{\pi}$, $Energy_{\pi}$, $QoA_{\pi}$ & Expected financial cost, latency, energy, and QoA of plan $\pi$. \\
\bottomrule
\end{tabular}
\end{table}
\fi
\subsection{Modeling Multi-LLM Planning}
\label{sect:model}

\pseudosection{Question model.} \label{subsection:Q-model}
We model each user question as a tuple
\(
Q = (p_t,\, T,\, F_{\text{max}},\, L_{\text{max}},\, E_{\text{max}},\, \mathrm{QoA}_{\text{min}},\, W),
\)
where $p_t$ is the textual prompt, explicitly formulated as a question (e.g., ``Who are currently the oldest and youngest presidents of an EU country?"), \(T\) represents its topic(s) (e.g., ``politics''). The parameters $F_{\text{max}}$, $L_{\text{max}}$, $E_{\text{max}}$, and $QoA_{\text{min}}$ specify user constraints that represent the maximum allowable total financial cost (USD/question), maximum total latency (seconds/question), maximum energy usage (J/question), and minimum acceptable QoA\footnote{We use the generic term \textit{QoA} to denote any metric to assess model performance depending on the task and dataset, such as accuracy, BLEU, F1, or embedding-based similarity measures (e.g., cosine similarity).} (on a scale 0-1, 1 being the highest). We collectively refer to these as the \textit{budget} $\mathcal{B}$. The weight vector $W = (w_F, w_L, w_E, w_{\mathrm{QoA}})$, with $\sum_i w_i = 1$, encodes relative importance across objectives.
Importantly, $\mathcal{B}$ defines \emph{feasibility} (candidate plans must satisfy the user-constraints), whereas $W$ is applied only
after planning to select one plan from the Pareto-optimal set; $W$ is not used to scalarize objectives during plan search. In practice, users do not need to specify all constraints manually. \system can expose presets such as Cost-saving, Balanced, High-quality, Low-latency, and Energy-aware, which translate internally into default bounds and objective weights. We distinguish between two question types: (1) those with \textit{binary correctness} (e.g., multiple-choice or yes/no questions), where \(QoA \in \{0,1\}\); and (2) those with \textit{open-ended or free-text} answers, where \(QoA \in [0,1]\) reflects graded partial correctness (e.g., ``John Smith III'' vs.\ ``J.\ Smith''). \system's implementation provides estimators for QoA in both cases.

\pseudosection{LLM Model.} \label{subsection:L-model}
Let $L = \{L_1, L_2, \dots, L_n\}$ be the available LLMs. Each $L_i$ is represented by the tuple
$(f_i, l_i, e_i, \mathit{qoa}_i, \mathit{Tok}_i, T_i^{\text{out}})$, where $f_i$, $l_i$, and $e_i$
denote the expected financial cost, latency, and energy consumption of invoking $L_i$ under an execution context;
$\mathit{Tok}_i(\cdot)$ is the model-specific tokenizer that maps a prompt $p_t$ to its input-token
count $\mathit{Tok}_i(p_t)$ (e.g., GPT-4 vs.\ Llama-3 use different tokenizers); and
$T_i^{\text{out}}$ is the expected number of output tokens generated by $L_i$ under the same context.
We use $\xi$ to denote the \emph{execution context} (e.g., topic, operator type, and model).
At planning time, Opti-Q instantiates the context-conditioned metrics
$\mathit{qoa}_i(\xi)$, $f_i(\xi)$, $l_i(\xi)$, $e_i(\xi)$, and $T_i^{\text{out}}(\xi)$~\footnote{We may omit explicitly mentioning $\xi$
when it is clear from context.} using the corresponding
empirical estimates retrieved from \textsc{PerfDB}, rather than
assuming a single global quality value per model. These metrics characterize the expected cost and behavior of model invocations within a plan. Each invocation cost includes a fixed and a variable component dependent on the number of input and output tokens under $\xi$ (e.g.,
\(
f_i(\xi, p_t) = f_i^{\text{fix}}(\xi) + f_i^{\text{var}}(\xi)\cdot \bigl(\mathit{Tok}_i(p_t) + T_i^{\text{out}}(\xi)\bigr).
\)
Here, $f_i^{\text{fix}}(\xi)$ captures the minimum per-invocation cost, regardless of the specific question. To handle stochastic intermediate output lengths, \textsc{PerfDB} stores the expected output length
$T_i^{\text{out}}(\xi)$ estimated from historical traces under the same context $\xi$, which Opti-Q
uses to estimate token-dependent plan costs during planning.

\pseudosection{Plan Model.} \label{subsection:p-model}\label{subsection:L-model}
For a question $Q$, let $\Pi$ denote the set of admissible plans. A plan $\pi \in \Pi$ specifies how one or more LLMs from $L$ are invoked to process $Q$. Each plan comprises operations $ \{\text{op}_1, \dots, \text{op}_n\}$, where each $\text{op}_i$ represents either an LLM invocation or a composition of LLMs, and a partial order $\prec$ defining execution dependencies. If $\text{op}_i \prec \text{op}_j$, then $\text{op}_j$ consumes the output of $\text{op}_i$. Initial operations, those with no predecessors, consume $Q$; final operations produce the plan's answer. Operations without order constraints will execute in parallel, and all operation outputs must either feed into subsequent operations or produce final answers. For example, a plan involving operations $\text{op}_1, \text{op}_2, \text{op}_3$ may specify $\text{op}_1 \prec \text{op}_2$, where both $\text{op}_1$ and $\text{op}_3$ execute in parallel and $\text{op}_2$ consumes $\text{op}_1$'s output.
A \emph{sequential operation}, denoted \(\text{Seq}(Q, L_1, \dots, L_k) = L_k \left( Q \oplus p_{ctx} \oplus L_{k-1} \left( Q \oplus p_{ctx} \oplus \dots \oplus L_1(Q) \right) \right)\),
concatenates ($\oplus$) the question, a context prompt $pt_{ctx}$ (that instructs the LLM on how to use it as a context; see Section~\ref{sect:framework}), and $L_1$'s output for input to $L_2$.
A \emph{parallel operation}, denoted 
\(\text{Par}(Q, \{L_1, \dots, L_k\}) = L_1(Q) \odot L_2(Q) \odot \dots \odot L_k(Q)\), poses $Q$ to each model independently and uses structured output fusion ($\odot$) to combine their outputs. $\odot$ signifies a ``concatenation'' in which the output of each model is prefixed with its identifier for traceability (e.g., ``LLM1: answer1; LLM2: answer2'').
% A \emph{parallel operation}, denoted \(\text{Par}(Q, \{L_1, \dots, L_k\}) = L_1(Q) \odot L_2(Q) \odot \dots \odot L_k(Q)\)r, invokes models independently and concatenates ($\odot$) their outputs.
A \emph{blending operation} combines multiple model outputs into a single answer~\cite{jiang2023llmblenderensemblinglargelanguage}. This operation, denoted $\text{Blend}(Q, \text{Par}(Q, \dots), L_b) =  L_b( Q\ \oplus pt_{bld} \ \oplus \text{Par}(Q, \dots))$, uses a dedicated ``blending'' model $L_b$, which takes the original question, a blending prompt $p_{bld}$, and the concatenated outputs of the parallel operation as input (Figure~\ref{fig:Arch_one} shows an example; more details on the prompt in Section~\ref{sect:framework}). Blending operations may appear \emph{only} after parallel operations, and each parallel operation must be immediately followed by exactly one blending operation. Note that the same model may appear multiple times in both sequential and parallel operations, so $\text{Seq}(Q, L_1, L_1)$ and $\text{Par}(Q, \{L_2, L_2\})$ are valid.

\pseudosection{Multi-LLM QA Planning Problem.}\label{subsection:p-model}\label{subsection:O-model}
Let $\pi \in \Pi$ be a plan using a subset of LLMs of $L$ to process $Q$. This plan induces total financial, latency, and energy consumption costs and obtains a QoA based on the chosen models and their composition. Our goal is to find Pareto-optimal plans balancing QoA, financial cost, latency, and energy under user constraints. Formally, we frame this as a MOO problem:
\[
\scalebox{0.85}{$
\begin{aligned}
\text{Maximize} \quad & \big[\mathrm{QoA}(\pi), -\mathrm{Financial}(\pi), -\mathrm{Latency}(\pi), -\mathrm{Energy}(\pi)\big] \\
\text{Subject to:} \quad & 
\begin{cases} 
\mathrm{Financial}(\pi) \leq F_{\max} \\ 
\mathrm{Latency}(\pi) \leq L_{\max} \\ 
\mathrm{Energy}(\pi) \leq E_{\max} \\ 
\mathrm{QoA}(\pi) \geq \mathrm{QoA}_{\min}
\end{cases}
\end{aligned}
$}
\]

% To solve this MOO problem, we employ (Non-dominated Sorting Genetic Algorithm), an evolutionary multi-objective optimization algorithm (NSGA-II) \cite{996017}, which: (1) maintains a population of candidate solutions, (2) applies non-dominated sorting to identify Pareto fronts, (3) computes crowding distance $d_i$ for solution diversity preservation, and (4) handles constraints through constrained-domination principles. The algorithm outputs a Pareto frontier of optimal trade-offs between the objectives $\mathrm{QoA}(\pi)$, $\mathrm{Financial}(\pi)$, $\mathrm{Latency}(\pi)$, and $\mathrm{Energy}(\pi)$. These objective values are computed from the model costs and execution patterns defined in the plan (see implementation in the next section). This approach enables \emph{weight-free a posteriori} decision making, allowing users to select solutions from $\mathcal{P}^*$ according to their preferred operational constraints without requiring predefined objective weights.

\subsection{Overview of \system Approach}
\begin{figure}[ht]
    \centering
\includegraphics[width=\linewidth]{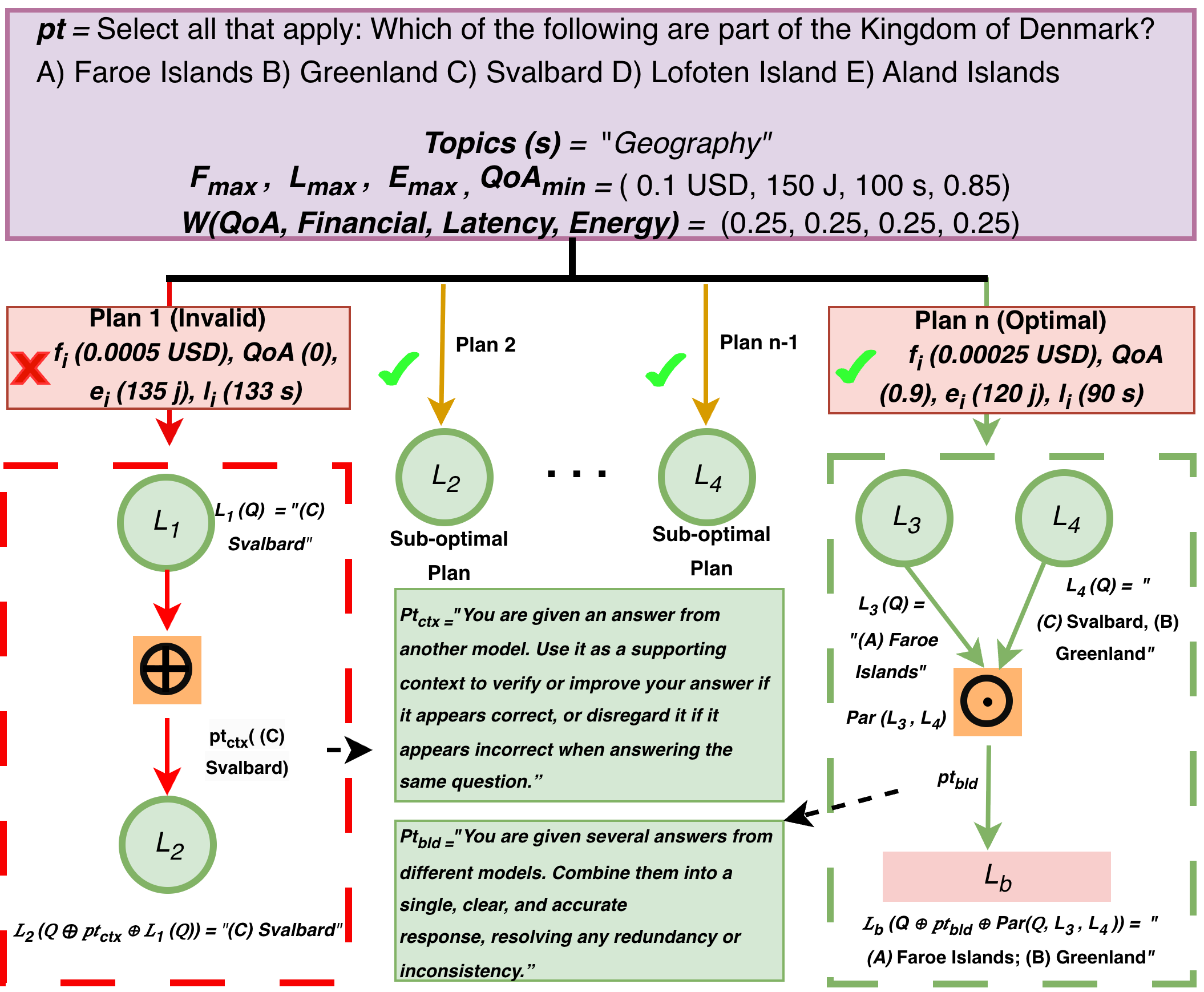}
    \caption{Comparison of candidate multi-LLM execution plans for a question.}
   \label{fig:Arch_one}
\end{figure}

For a given question $Q$, our approach selects the plan $\pi \in \Pi$ that best balances QoA against financial, latency, and energy costs. As shown in Figure~\ref{fig:Arch_one}, consider the example question, \textit{``Select all that apply: Which of the following are part of the Kingdom of Denmark? 
A) Faroe Islands
B) Greenland
C) Svalbard
D) Lofoten Islands
E) Aland Islands
''}. 
First, \system performs a \textbf{question parsing} phase to extract the prompt text $pt$, and topic(s) $T$ (detailed in Section~\ref{sect:framework}), along with user-defined constraints and priorities (budget $\mathcal{B}$ and weight vector $W$). 
% For the critical task of initial resource estimation (which drives planning and model selection), we compute an estimated input token length ($T^{in}_{\text{est}}(pt)$) using a generic tokenizer ( $\texttt{GPT-2}$). This estimate is sufficient for populating the search space, but the final, precise token count and resulting cost are calculated dynamically per model during execution using the model's native tokenizer.
We require the user to specify $\mathcal{B}$; maximizing QoA ``at any cost'' is supported but not our focus.
If $\mathbf{W}$ is omitted,  we default to uniform weights.

% If the user does not specify $\mathcal{B}$ explicitly, the framework assigns default values of \(\infty\) for financial, latency, and energy costs, and \(0\) for the minimum required quality. Similarly, if \(W\) is not specified, the weights are distributed equally by default.
% Next, \system determines the maximum number of LLM invocations allowed per plan, $k$. This value is derived directly from a user-defined budget, $\mathcal{B}$, which we assume is a prerequisite for any optimization. Consequently, scenarios aimed at maximizing QoA ``at any cost'' without budgetary constraints are considered outside the scope of this work. 
\pseudosection{Bounding the plan size.}
Next, \system determines the maximum number of operations allowed per plan, $k$, based on $\mathcal{B}$.
For each $L_i$, \system computes the maximum model invocations $n_i$ allowed under each constraint (dividing the relevant budget by model's per-constraint cost) and taking the minimum across constraints. For example, the budget in Figure~\ref{fig:Arch_one} constrains a resource-heavy LLM, $L_1$, to one call while allowing a lighter $L_2$ up to three calls. Limits are not mutually exclusive: a plan may call $L_1$ once and still use $L_2$ one or more times within the remaining budget. \system samples $k$ uniformly from $\left[min_i(n_i), max_i(n_i)\right]$; as shown in our evaluation and in traditional ML ensemble size selection methods~\cite{10.1145/3583133.3590562}, this randomized choice within the interval does not harm performance.

\pseudosection{Performance Statistics (\textsc{PerfDB}).} 
\system uses \textsc{PerfDB}, a statistical catalog that stores quality and resource statistics for individual LLM invocations and composed plans. \textsc{PerfDB} enables the planner to estimate plan metrics (i.e., latency, financial cost, QoA, and energy) without executing the plan. \textsc{PerfDB} is organized into levels of compositional depth (i.e., operator-nesting depth in the plan representation), which determines the granularity of stored statistics. Level~0 contains statistics for single LLMs obtained from public benchmarks (including ML.Energy~\cite{ml-energy-leaderboard} and Artificial Analysis\footnote{\url{https://artificialanalysis.ai/}}. Level~1 aggregates results of single operations with ``depth~1'', e.g., $\text{Seq}(Q, L_{1}, L_{2})$ or $\text{Par}(Q, \{L_{1}, L_{2}, L_{3}\})$. Level~2 and higher, store measurements for subsets of plans with ``depth~2'' or higher (e.g., \(\text{Blend}(Q, \text{Seq}(Q, L_{1}, L_{2}), \text{Par}(Q, \{L_{3}, L_{4}\}), L_{b})\)). All entries are keyed by execution context $\xi$, including topic, operator type, model, prompt configuration, and serving backend, thus capturing both workload-dependent behavior and backend-specific latency, energy, and throughput profiles.
\textsc{PerfDB} is populated offline and incrementally after execution. For each executed plan, OPTI-Q appends a compact trace containing the execution context, plan structure, invocation configuration, token usage, and \emph{observed} resource metrics. Cost and latency are obtained from API responses and execution logs. Energy is recorded when exposed by the backend, or \emph{estimated} from token counts and prior measurements otherwise. QoA is updated only when quality feedback is available, such as benchmark labels, known reference answers, analyst judgments, user feedback, or automatic quality estimators. Thus, \textsc{PerfDB} stores \emph{observed} values when available and \emph{estimated} values otherwise. Since each execution contributes one compact trace, storage grows linearly with the number of logged executions, and \textsc{PerfDB} does not need to be rebuilt from scratch. New models, fine-tuned checkpoints, prompt variants, and serving backends are represented as distinct configured LLM invocation instances with initially sparse statistics; \system bootstraps these instances from metadata or similar invocations and refines their estimates as new benchmark runs and live traces accumulate. Because LLM outputs and serving behavior may vary across runs, \textsc{PerfDB} also stores variance estimates and confidence intervals, enabling risk-aware planning when reliability is important. Section~VI analyzes how varying \textsc{PerfDB} coverage affects planning performance.

\pseudosection{Plan Generation.} In this phase, \system generates plans with up to $k$ operations. Because the  plan space grows exponentially in $|L|$ and $k$ (as we will detail in the following section), an exhaustive search is infeasible. \system therefore supports a  ``pluggable optimization engine" that includes diverse planning strategies (Section~\ref{sect:plan-gen}): (i) Dynamic Programming (DP), an exact solver that can recover the Pareto frontier for small $\left| L \right|$ and k but suffers from state-space explosion; (ii) stochastic Hill Climbing (HC), a lightweight greedy heuristic with low planning latency that can
stagnate in local optima in our non-convex multi-objective landscape; and (iii)
NSGA-II~\cite{996017}, a multi-objective evolutionary algorithm that balances exploration
and exploitation. This design supports exact optimization when feasible, lightweight heuristics otherwise, and \textsc{NSGA-II} as the default scalable Pareto-front approximator.

% NSGA-II initializes a diverse population by sampling ``distinct'' candidate plans (e.g., treating $Par(Q,L_1,L_2)$ and $Par(Q,L_2,L_1)$ as equivalent). Each chromosome in the NSGA-II population encodes a candidate plan (Figure~\ref{fig:Arch_one} shows sequential \emph{(Plan 1)}, single-LLM \emph{(Plan 2)}, and parallel \emph{(Plan $n$)} forms). NSGA-II evaluates plans on four competing objectives: maximizing QoA while minimizing financial cost, latency, and energy consumption, with fitness estimates from PerfDB. Iterative non-dominated sorting with crowding-distance preservation maintains diversity while converging toward the Pareto frontier. After the maximum generations, \system selects an optimal plan. In our running example, \emph{(Plan $n$)} is selected  because it is expected to achieve a better trade-off:  high QoA (0.9) while controlling financial cost (0.00025 USD), latency (90 s) and energy consumption (120 J). 

\pseudosection{Plan Execution Phase.} To execute the selected plan, operations are carried out according to the plan topology, applying context propagation and blending where applicable. For example, consider the execution of \emph{Plan $n$} in Figure~\ref{fig:Arch_one}, which includes a parallel call followed by blending, i.e., $\operatorname{Blend}(Q,\operatorname{Par}(Q,L_3,L_4),L_b)$, and is selected in this case. Models $L_3$ and $L_4$ run in parallel to answer the multi-answer user prompt. Suppose $L_3$ returns a partial but correct subset (e.g., (A) Faroe Islands), while $L_4$ returns (B) Greenland but also includes a plausible yet incorrect extra option (e.g., (C) Svalbard). The blender $L_b$ aggregates the candidate sets, cross-checks them against the question constraints, removes unsupported options, and produces a consolidated final prediction (A, B) (Faroe Islands; Greenland). This simple example illustrates the strength of multi-objective planning: a single-model plan (only $L_4$) may be cheaper but less robust, risking error propagation. We quantify variability due to nondeterminism by repeated executions and confidence intervals in Section~\ref{sect:expr}.

\section{Plan Generation}
\label{sect:plan-gen}
To answer a question using multiple LLMs, we must (i) represent candidate execution plans in a compact form that supports efficient manipulation, (ii) estimate a plan's cost--benefit \emph{without} executing it, and (iii) search the resulting plan space to find high-quality tradeoffs. 

\subsection{Plan Encoding} 
\label{sect:planrep}

\begin{figure}[htb!]
\centering
\includegraphics[width=\linewidth]{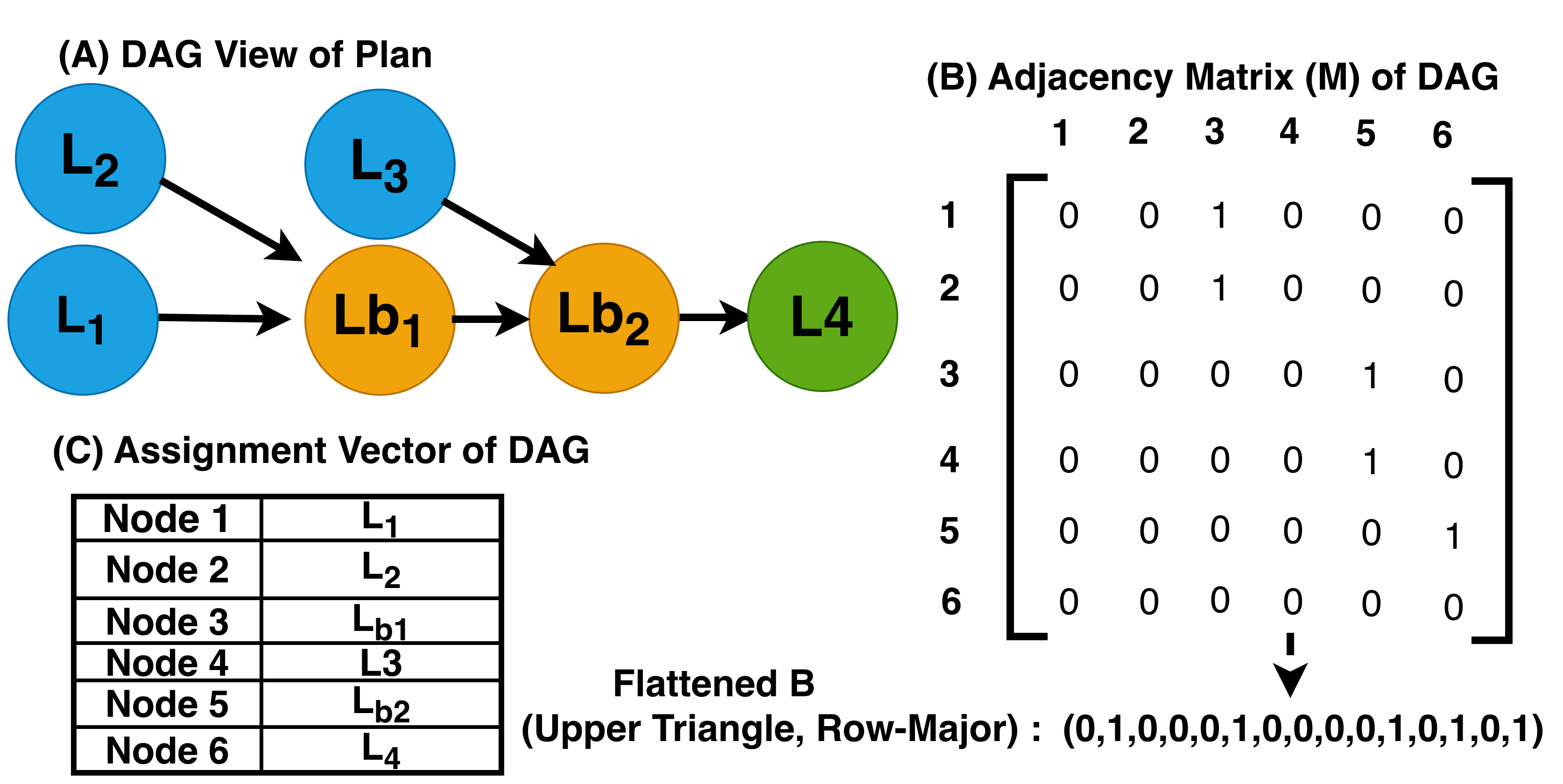}
\caption{Sample question plan encoding.}
\label{fig:complex-dag}
\end{figure}

All planners in Opti-Q
operate over the same compact, yet expressive encoding candidate plan~$\pi$. We represent $\pi$ as a tuple $(\textit{connectivity\_map}, \mathbf{m})$, where $\textit{connectivity\_map}\) canonically encodes the plan's directed acyclic execution topology and  $\mathbf{m}$ assigns one model to each of the $k$ nodes. Figure~\ref{fig:complex-dag} shows an example plan with its encoding. The DAG topology over $k$ operations is encoded by the upper-triangular adjacency matrix $M \in \{0,1\}^{k\times k}$, with $M_{ij} = 1$ for $(1 \leq i < j \leq k)$ iff there is a data-flow edge from $L_i$ to $L_j$. For computational efficiency, we flatten $M$ into a $k(k-1)/2$-bit  vector $B = (b_{12}, b_{13}, \dots, b_{(k-1)k})$, where each bit $b_{ij} \in \{0,1\}$ indicates whether $L_i$ feeds its output to $L_j$. Blending operations are implicit: a designated blending model ($L_b$) is invoked to combine inputs if a node $L_j$ has an in-degree $\deg^-(L_j) > 1$. To ensure invariance under node relabeling (structural isomorphism), we enumerate all valid topological orderings of the DAG and select the lexicographically maximal bit vector, $B_{\text{max}}$, defining the canonical integer identifier as $\textit{connectivity\_map} = \operatorname{integer}(B_{\text{max}})$. This encoding method guarantees acyclicity by construction, facilitates efficient manipulation, and provides a unique, isomorphism-aware representation of complex execution flows. We enforce the constraint that the final node must have an in-degree $\geq 1$, ensuring the plan produces a valid output.

\vspace{-0.3cm}
Figure~\ref{fig:complex-dag}(A) illustrates an execution plan composed of $k = 5$ operations constructed from Sequential, Parallel, and Blending operations, while Figure~\ref{fig:complex-dag}(B) and (C) show its connectivity matrix and model assignments, respectively. The plan begins with a parallel operation where models $L_1$ and $L_2$ process the question $Q$, defined as $op_1 = \text{Par}(Q, (L_1, L_2))$. Their outputs are then blended using model $L_{b_1}$, forming $op_2 = \text{Blend}(Q, op_1, L_{b_1})$. Concurrently, an independent parallel operation executes model $L_3$, defined as $op_3 = \text{Par}(Q, (L_3))$. The results from the first blend ($op_2$) and the execution of $L_3$ ($op_3$) are synthesized in a hierarchical blending step, $op_4 = \text{Blend}(Q, op_2, op_3, L_{b_2}$). Finally, in a sequential operation, $op_5 = \text{Seq}(Q, op_4, L_4)$, model $L_4$ processes the original question $Q$ again using the output of $op_4$ as context to produce the final answer. This encoding captures the execution structure compactly and canonically, requiring $\mathcal{O}(k^2)$ space.

\subsection{Cost-Benefit Estimation}
\label{sect:estimators}

For each candidate plan, we estimate (pre-execution) its expected QoA, financial cost, energy, and latency\ifextended
(see Algorithm~\ref{alg:nsga2})\fi.

\pseudosection{Token Estimation.}
Cost-benefit metrics depend on the number of tokens processed and generated, which we must estimate pre-execution. For each model $L$, let $\mathit{Tok}_{L}(\cdot)$ denote its tokenizer (models may segment the same string differently) and let $T^{\mathrm{out}}_{L}$ be the model's expected output length (e.g., average response tokens from PerfDB). We estimate two quantities for a plan $\pi$: total generated tokens $T^{\mathrm{out}}_{\pi}$ and total processed tokens $T^{\mathrm{proc}}_{\pi}$ (input + output across model invocations).

\textit{Sequential.} In $\mathrm{Seq}(Q,L_1,L_2)$, $L_1$ consumes $\mathit{Tok}_{L_1}(Q)$ tokens and produces $T^{\mathrm{out}}_{L_1}$. The second model consumes the original question plus a context prompt $pt_{\mathrm{ctx}}$ and the \emph{immediate predecessor's} output, i.e.,
$\mathit{Tok}_{L_2}\!\bigl(Q + pt_{\mathrm{ctx}} + T^{\mathrm{out}}_{L_1}\bigr)$, and produces $T^{\mathrm{out}}_{L_2}$. We conservatively sum outputs,
$T^{\mathrm{out}}_{\mathrm{Seq}} = T^{\mathrm{out}}_{L_1} + T^{\mathrm{out}}_{L_2}$, allowing for potential redundancy where later models restate prior text. For a length-$k$ chain $\mathrm{Seq}(Q,L_1,\ldots,L_k)$ we avoid exponential growth by appending only the immediate predecessor’s output at each step:
\(
T_{\pi}^{\mathrm{proc}}
= \bigl(\mathit{Tok}_{L_1}(Q) + T^{\mathrm{out}}_{L_1}\bigr)
+ \sum_{i=2}^{k} \Bigl(\mathit{Tok}_{L_i}\!\bigl(Q + pt_{\mathrm{ctx}} + T^{\mathrm{out}}_{L_{i-1}}\bigr)
+ T^{\mathrm{out}}_{L_i}\Bigr).
\)
Equivalently, chains can be viewed as nested compositions, e.g., $\mathrm{Seq}(Q,L_1,L_2,L_3)=\mathrm{Seq}(Q,\mathrm{Seq}(Q,L_1,L_2),L_3)$, with $pt_{\mathrm{ctx}}$ applied from the second step onward.

\text{Parallel + blending.} In $\mathrm{Blend} \!\bigl(Q, \mathrm{Par}(Q,(L_1,\ldots,L_k)), L_b\bigr)$, each $L_i$ processes $\mathit{Tok}_{L_i}(Q)$ input tokens and produces $T^{\mathrm{out}}_{L_i}$ output tokens. The blending model $L_b$ consumes the question, a blending prompt $pt_{\mathrm{bld}}$, and the concatenated outputs:
\(
\mathit{Tok}_{L_b}\!\Bigl(Q + pt_{\mathrm{bld}} + \sum_{i=1}^{k} T^{\mathrm{out}}_{L_i}\Bigr)
\)
and produces $T^{\mathrm{out}}_{L_b}$. Thus,
\(
T^{\mathrm{proc}}_{\pi}
=
\sum_{i=1}^{k} \bigl(\mathit{Tok}_{L_i}(Q) + T^{\mathrm{out}}_{L_i}\bigr)
\;+\;
\mathit{Tok}_{L_b}\!\Bigl(Q + pt_{\mathrm{bld}} + \sum_{i=1}^{k} T^{\mathrm{out}}_{L_i}\Bigr)
\;+\;
T^{\mathrm{out}}_{L_b},
\qquad
T^{\mathrm{out}}_{\pi}=T^{\mathrm{out}}_{L_b}.
\)

% For a parallel operation followed by blending, each model in the parallel set processes the same prompt
% concurrently. Specifically, in $\mathrm{Blend} \!\bigl(Q, \mathrm{Par}(Q,(L_1,\ldots,L_k)), L_b\bigr)$, each
% $L_i$ processes $\mathit{Tok}_{L_i}(Q)$ input tokens and produces $T^{\mathrm{out}}_{L_i}$ output tokens.
% The blending model $L_b$ then combines these results using a blending prompt $pt_{\mathrm{bld}}$, receiving
% \(
% \mathit{Tok}_{L_b}\!\Bigl(Q + pt_{\mathrm{bld}} + \sum_{i=1}^{k} T^{\mathrm{out}}_{L_i}\Bigr)
% \)
% input tokens and producing $T^{\mathrm{out}}_{L_b}$ output tokens. Hence, the total number of processed
% tokens in a parallel+blending operation is
% \(
% \sum_{i=1}^{k} \bigl(\mathit{Tok}_{L_i}(Q) + T^{\mathrm{out}}_{L_i}\bigr)
% \;+\;
% \mathit{Tok}_{L_b}\!\Bigl(Q + pt_{\mathrm{bld}} + \sum_{i=1}^{k} T^{\mathrm{out}}_{L_i}\Bigr)
% \;+\;
% T^{\mathrm{out}}_{L_b},
% \)
% producing $T^{\mathrm{out}}_{\mathrm{Par}} = T^{\mathrm{out}}_{L_b}$ output tokens.

\textit{General plans.} For an arbitrary (nested) plan $\pi$, we compute $(T^{\mathrm{proc}}_{\pi},T^{\mathrm{out}}_{\pi})$ by recursively applying the above rules bottom-up over the plan tree/DAG and summing the processed tokens over all invoked models. This yields a conservative, tokenizer-aware estimate of token volume that propagates through both sequential and parallel compositions. 
\ifextended
 
 For the nested plan (Figure~\ref{fig:complex-dag}),
\[\small
\begin{aligned}
\mathrm{Seq}\big(&Q,\,
\mathrm{Blend}\big(Q,\,
\mathrm{Par}\big(Q,\,
(L_3,\,
\mathrm{Blend}\big(Q,\,
\mathrm{Par}(Q,(L_1,L_2)),\,
L_{b_1}\big)\big),\,
L_{b_2}\big),\,
L_4\big)\big)
\end{aligned}
\]
tokens are accumulated from the innermost operation outward. For clarity, we assume that all models share the same tokenizer. We begin with the parallel operation $\mathrm{Par}(Q,(L_1,L_2))$, where both $L_1$ and $L_2$ process the question $Q$ ($\mathit{Tok}_{L_1}(Q) = \mathit{Tok}_{L_2}(Q) = 100$) and produce $T^{\mathrm{out}}_{L_1} = 50$ and $T^{\mathrm{out}}_{L_2} = 60$ tokens, respectively. Their combined outputs (110 tokens) are passed to the blending model $L_{b_1}$, which also receives the original question and a blending prompt ($pt_{\mathrm{bld}} = 20$), yielding a total input of 
\(
\mathit{Tok}_{L_{b_1}}(Q + pt_{\mathrm{bld}} + T^{\mathrm{out}}_{L_1} + T^{\mathrm{out}}_{L_2}) = 230
\)
tokens and producing $T^{\mathrm{out}}_{L_{b_1}} = 100$ tokens. This blended output then feeds into another blending operation $L_{b_2}$ together with model $L_3$, which processes $Q$ independently 
($\mathit{Tok}_{L_3}(Q) = 100$ and $T^{\mathrm{out}}_{L_3} = 150$, giving $T^{\mathrm{proc}}_{L_3} = 250$). 
The two outputs—$L_{b_1}$’s 100 tokens and $L_3$’s 150 tokens—are then fused by $L_{b_2}$, which processes
\(
\mathit{Tok}_{L_{b_2}}(Q + pt_{\mathrm{bld}} + T^{\mathrm{out}}_{L_{b_1}} + T^{\mathrm{out}}_{L_3}) = 490
\)
tokens and produces $T^{\mathrm{out}}_{L_{b_2}} = 120$ tokens. 
Finally, in the outer sequential operation, model $L_4$ processes the question $Q$, the output from $L_{b_2}$, and a context prompt ($pt_{\mathrm{ctx}} = 30$), for a total of 
\(
\mathit{Tok}_{L_4}(Q + pt_{\mathrm{ctx}} + T^{\mathrm{out}}_{L_{b_2}}) = 400
\)
tokens, and produces $T_{L_4}^{\mathrm{out}} = 150$ tokens.  Summing all components, the plan processes $\approx\!1{,}780$ tokens, illustrating how token counts propagate through composition.
\else
The extended version~\cite{optiqextended2026} includes a complete example computation for the plan in Figure~\ref{fig:complex-dag}.
\fi

\pseudosection{QoA Estimation.}
Predicting a plan's QoA before execution is difficult due to model nondeterminism and the combinatorial space of multi-LLM plans. We therefore estimate QoA using \textsc{PerfDB}, a performance database populated from offline benchmarks and execution traces that stores topic-conditioned QoA statistics for (i) atomic invocations (e.g., $(Q,L_i)$) and (ii) common local compositions  (e.g., $Seq(Q, L_i, L_j)$ and $Blend(Q,Par(Q,\{L_1,\ldots,L_k\}),L_b)$). The goal is not perfect calibration, but a conservative estimate that preserves \emph{relative} differences between candidate plans, which is sufficient for Pareto-based search.

\textit{Topic conditioning and lookup.} Let $\mathcal{T}=\{T_1,\ldots,T_m\}$ be the inferred topics of $Q$. For any operator or substructure $op$, we aggregate topic-specific entries as $QoA(op) = \frac{1}{m} \sum_{i=1}^{m} QoA_{op, T_i}$.
To initialize estimation, we first attempt an exact \textsc{PerfDB} match for the whole plan (or large substructures). If not available, we perform a fuzzy match using a weighted similarity that combines structural overlap with a normalized edit-distance over a canonicalized plan string; we accept a fuzzy match only if the score exceeds a fixed threshold (e.g., 0.75 in our experiments). Importantly, all QoA values used during planning are retrieved from \textsc{PerfDB} under a single schema (topic + operator/substructure key), even when the retrieval itself is fuzzy.
\ifextended
The complete process is summarized in Algorithm~\ref{alg:process-dag}.
\else
The complete procedure appears in the extended version of the paper~\cite{optiqextended2026}.
\fi

\textit{Sequential refinement.}
For a sequential dependency where $op_j$ consumes the output of $op_i$ (e.g., a local $\text{Seq}(\cdot,L_i,L_j)$), we update the running estimate using a pairwise relative-effect factor derived from \textsc{PerfDB}:
\(
QoA_{\text{next}} =
\operatorname{clip}_{[0,1]}\!\left(
QoA_{\text{curr}} \cdot
\frac{QoA(Q,\text{Seq}(L_i,L_j))}{\max(\epsilon,\, QoA(Q,L_j))}
\right).
\)
where $\epsilon$ is a small constant and $\operatorname{clip}_{[0,1]}(\cdot)$ bounds the result to $[0,1]$.
Intuitively, the ratio compares the composed step against the successor alone, capturing the marginal impact of inserting $L_i$ before $L_j$ under the same topic-conditioned context. 
We adopt a multiplicative relative-effect formulation because composition effects in our traces scale with the successor's baseline quality (i.e., prepending a context-providing model tends to improve a weak successor proportionally more than a strong one) which an additive model does not capture. The $\operatorname{clip}_{[0,1]}(\cdot)$ guards the small fraction of cases where sparse-coverage ratios extrapolate beyond the valid range.
\emph{Example:} if $QoA(Q,L_j)=0.50$ and $QoA(Q,\text{Seq}(L_i,L_j))=0.60$, the factor is $1.2$; with $QoA_{\text{curr}}=0.55$, we obtain $QoA_{\text{next}}\approx \operatorname{clip}_{[0,1]}(0.55 \times 1.2)=0.66$.

\textit{Parallel + blending.}
For a blend node \(op_{L_b}\), where the blending operation is defined as \(\mathrm{Blend}(Q,\mathrm{Par}(Q,\{L_1,\ldots,L_k\}),L_b)\),
we combine the incoming branch estimates using \textsc{PerfDB} references for (i) the blend operator itself and (ii) each predecessor run standalone.
Let
$QoA^{\text{ref}}_{op_{L_b}}=QoA(Q,op_{L_b})$ and
$QoA^{\text{ref}}_{L_s}=QoA(Q,L_s)$.
Given the branch estimates computed in the current traversal, $QoA^{\text{new}}_{L_s}$, we scale the reference blend quality by the average relative change of its inputs:
{\small\[
QoA^{\text{new}}_{op_{L_b}} =
\operatorname{clip}_{[0,1]}\!\left(
QoA^{\text{ref}}_{op_{L_b}}
\cdot\left(
1 + \frac{1}{k} \sum_{s=1}^{k}
\frac{QoA^{\text{new}}_{L_s} - QoA^{\text{ref}}_{L_s}}{\max(\epsilon,\, QoA^{\text{ref}}_{L_s})}
\right)
\right).
\]
}

This estimator is robust to a single weak branch (average rather than product) while still reflecting systematic improvements/degradations across branches.
Note that the averaging form estimates the expected blend quality and is deliberately agnostic to which branch is correct; it does not model the error-correction mechanism illustrated in Figure~\ref{fig:Arch_one}, which is realized at execution time. This is a conservative choice that avoids over-crediting blends on the basis of a single strong branch at planning time.
\emph{Example:} suppose $k=2$, $QoA^{\text{ref}}_{op_{L_b}}=0.60$, and references $QoA^{\text{ref}}_{L_1}=0.50$, $QoA^{\text{ref}}_{L_2}=0.40$. If the current plan yields $QoA^{\text{new}}_{L_1}=0.60$ and $QoA^{\text{new}}_{L_2}=0.44$, then the average relative change is
$\frac{1}{2}\left(\frac{0.60 - 0.50}{0.50}+\frac{0.44 - 0.40}{0.40}\right)=0.15$,
so $QoA^{\text{new}}_{op_{L_b}}=\operatorname{clip}(0.60\times(1+0.15))=0.69$.

\textit{Plan-level propagation and fallbacks.}
\ifextended
We compute the plan QoA by traversing the plan DAG in a topological order that respects precedence constraints (see Algorithm~\ref{alg:process-dag}).
\else
We compute the plan QoA by traversing the plan DAG in a topological order that respects precedence constraints (the complete algorithm is in the extended version~\cite{optiqextended2026}).
\fi
So, an operator is processed only after all prerequisites are estimated. 
Along edges corresponding to sequential consumption we apply the sequential update rule; at blend nodes we wait for all incoming branches and apply the blending rule. When \textsc{PerfDB} lacks a required composed entry (e.g., missing $QoA(Q,\text{Seq}(L_i,L_j))$), we fall back to the closest available statistics (e.g., use other known predecessors of $L_j$ to estimate an average relative-effect factor, or default to a neutral factor of $1$ when no pairwise information exists). For nested blend structures where direct references are sparse, we normalize the structure by flattening nested blends before performing lookups, then apply the same blend rule on the resulting fan-in.
These fallbacks ensure estimation completes under incomplete coverage; in Section~\ref{sect:expr} we quantify how estimation error varies with \textsc{PerfDB} coverage.

\ifextended
Algorithm~\ref{alg:process-dag} summarizes the preprocessing stage used before traversal-based QoA estimation. Its purpose is to populate two maps from \textsc{PerfDB}: \textit{Matched}, which stores metrics for whole-plan or subplan matches keyed by sink node, and \textit{BlendRef}, which stores the reference metrics required by the blend estimator. The procedure first attempts an exact or fuzzy lookup of the entire plan. If no full-plan match is found, it searches for maximal sequential sub-chains and stores any successful matches in \textit{Matched}. It then constructs \textit{BlendRef} by collecting, for each blend node, both the reference metrics of the blend operator itself and the standalone reference metrics of its predecessor inputs. These outputs are later consumed by the traversal procedure in Appendix~\ref{app:trvsl}.

\begin{algorithm}[!htb]
\footnotesize
\caption{\textsc{Process Plan}}
\label{alg:process-dag}
\begin{algorithmic}[1]
\Require DAG $G=(V,E)$,\; question $Q$ (for topic $T$),\; fuzzy threshold $s$,\; historical performance
database \textsc{PerfDB}
\Ensure Two maps: $\mathit{Matched}$ (sink node $\mapsto$ metrics object) and $\mathit{BlendRef}$ (blend node $\mapsto$ map[input node $\mapsto$ metrics object])
\Statex\textbf{\textit{Phase 0 – validation}}
\If{\textbf{not} \Call{IsValidDAG}{$G$}}
    \State \textbf{raise} \textsc{InvalidDagError}
\EndIf
\State $\mathit{Matched}\gets\varnothing$;\; $\mathit{BlendRef}\gets\varnothing$
\Statex\textbf{\textit{Phase 1 – whole-graph lookup}}
\State $\mathcal{X} \gets$ \Call{PerfDBLookupExact}{$G,Q$} \Comment{Lookup based on graph structure and question topics}
\If{$\mathcal{X}=\varnothing$}
    \State $\mathcal{X} \gets$ \Call{PerfDBLookupFuzzy}{$G,Q,s$} \Comment{Fuzzy match based on structure, semantics, threshold $s$}
\EndIf
\If{$\mathcal{X}\neq\varnothing$}
    \State $\mathit{Matched}[\Call{Sink}{$G$}]\gets \mathcal{X}$ \Comment{Store metrics object}
    \State \Return $(\mathit{Matched},\mathit{BlendRef})$ \Comment{Return early if full match found}
\EndIf
\Statex\textbf{\textit{Phase 2 – sequential sub-chains}}
\State $S \gets$ \Call{StartNodes}{$G$}
\ForAll{$C \in$ \Call{FindMaximalSeqChains}{$G,S$}}  \Comment{$C$ is a DAG representing one chain}
    \State $\mathcal{X} \gets$ \Call{PerfDBLookupExact}{$C,Q$}
    \If{$\mathcal{X}=\varnothing$}
        \State $\mathcal{X} \gets$ \Call{PerfDBLookupFuzzy}{$C,Q,s$} \Comment{Recursive fuzzy match on sub-chains if needed}
    \EndIf
    \If{$\mathcal{X} \neq \varnothing$}
        \State $\mathit{Matched}[\Call{Sink}{$C$}] \gets \mathcal{X}$ \Comment{Store metrics for the matched chain}
    \EndIf
\EndFor
\Statex\textbf{\textit{Phase 3 – blend reference table}}
\State $G'\gets$ \Call{FlattenNestedBlends}{$G$} \Comment{Optional: Simplify structure if needed}
\ForAll{$b \in$ \Call{BlendNodes}{$G'$}}
    \State $\mathcal{X}_{\text{blend}} \gets$ \Call{DBLookupParallelRef}{$b,Q$} \Comment{Reference metrics for blend op $b$}
    \State $\mathit{BlendRef}[b][\text{'ref\_blend'}] \gets \mathcal{X}_{\text{blend}}$
    \ForAll{$u \in$ \Call{Predecessors}{$G',b$}}
        \State $\mathcal{X}_{\text{in}} \gets$ \Call{DBLookupSingleNode}{$u,Q$} \Comment{Reference metrics for input $u$ (standalone)}
        \If{$\mathcal{X}_{\text{in}}\neq\varnothing$}
            \State $\mathit{BlendRef}[b][u]\gets \mathcal{X}_{\text{in}}$ \Comment{Store standalone metrics for input $u$}
        \EndIf
    \EndFor
\EndFor
\State \Return $(\mathit{Matched},\mathit{BlendRef})$
\end{algorithmic}
\end{algorithm}

\fi

\pseudosection{Financial Cost Estimation.}
We estimate the financial cost of a plan as the sum of its operation-level costs~\cite{chen2023frugalgptuselargelanguage}. For a single invocation of model $L_i$, we use a simple fixed-plus-variable pricing model: a fixed per-call charge $f_i^{\text{fix}}$ and a per-token rate $f_i^{\text{var}}$ applied to the total number of billed tokens (prompt + generated). Using our token estimator, the expected billed tokens for $op_i$ equal the model-specific tokenized prompt length plus the predicted output length:
\(
\mathit{Financial}(Q,L_i)= f_i^{\text{fix}} + f_i^{\text{var}}\cdot\bigl(\mathit{Tok}_i(p_t)+T^{\text{out}}_{op_i}\bigr)
\).
For a parallel operator $\text{Par}(Q,\{L_1,\ldots,L_n\})$, invocations are independent, so costs add:
\(
\mathit{Financial}\!\left(\text{Par}(Q,\{L_1,\ldots,L_n\})\right)=\sum_{i=1}^{n}\mathit{Financial}(Q,L_i)
\). 
For a general plan $\pi$, we account for prompt growth in sequential and nested structures by charging each operation for its own prompt plus the outputs of its immediate predecessors:
\(\
\mathit{Financial}(\pi)=\sum_{i=1}^{n}\left(
f_i^{\text{var}}\left(
\mathit{Tok}_i(p_t)+\textstyle\sum_{op_j\in \mathrm{pred}(op_i)}\! T^{\text{out}}_{op_j}+T^{\text{out}}_{op_i}
\right)
+f_i^{\text{fix}}
\right) 
\).  
Here, \(\mathrm{pred}(op_i)\) denotes the set of operators whose outputs are included in $op_i$'s prompt (e.g., the immediate predecessor in a sequence, or all fan-in branches for a blend node).

% First, we estimate the cost of posing a question to any single LLM as a factor of its fixed costs and the variable costs that depend on the number of input and output tokens (i.e., \(
% \mathit{Financial}(Q,L_i)= f_i^{\text{fix}} + f_i^{\text{var}}\cdot\bigl(\mathit{Tok}_i(p_t)+T^{\text{out}}_{op_i}\bigr)
% \)).
% Based on this, the cost of a parallel operation is trivially estimated as the sum of the independent cost estimation per model (i.e., \(
% \mathit{Financial}\!\left(\text{Par}(Q,\{L_1,\ldots,L_n\})\right)=\sum_{i=1}^{n}\mathit{Financial}(Q,L_i)
% \)). The cost of sequential operations has to take into account that the output of one model is used as input for another. The cost accumulates across operations as \(\
% \mathit{Financial}(\pi)=\sum_{i=1}^{n}\left(
% f_i^{\text{var}}\left(
% \mathit{Tok}_i(p_t)+\textstyle\sum_{op_j\in \mathrm{pred}(op_i)}\! T^{\text{out}}_{op_j}+T^{\text{out}}_{op_i}
% \right)
% +f_i^{\text{fix}}
% \right) 
% \).

\pseudosection{Energy Consumption Estimation.}  
LLM energy consumption is approximately proportional to the number of tokens processed and generated~\cite{You2022ZeusUA}. Let \( e_{i} \) (\(\mathrm{J/token}\)) denote the energy per token for operation \( op_i \) obtained from \textsc{PerfDB} or external measurements~\cite{ml-energy-leaderboard}). We estimate plan energy by reusing the same structural accounting as financial cost, but without a fixed per-call term. This keeps energy estimation consistent with how prompts and intermediate outputs propagate through the plan.

% Since energy consumption follows the same structure as financial cost estimation, we use the same equations, replacing financial costs \( f_i \) with energy rates \( e_i \) and removing the fixed cost, which is typically not considered in energy measurements~\cite{ml-energy-leaderboard}. Thus, energy is computed analogously by substituting financial cost factors with \( e_i \), ensuring consistency in estimation between operations.

\pseudosection{Latency Estimation.}
Plan latency depends on both the plan structure and the number of tokens processed and generated~\cite{shekhar2024optimizingcostsllmusage}.
For a single invocation $op_i$, we approximate latency as linear in token volume using a per-token time coefficient $l_i$ (s/token)
\( Latency(Q, L_i) = l_i (\mathit{Tok}_i(p_t)+ T^{\text{out}}_{op_i}) \). 
For sequential (or more generally, precedence-constrained) portions of a plan, latencies add along executed operations, using the same prompt-growth accounting as above. For a pure parallel operator $\text{Par}(Q,\{L_1,\ldots,L_n\})$, branches execute concurrently, so the end-to-end latency is dominated by the slowest branch. For mixed plans with parallelism and joins, we apply these rules along the plan DAG: sum along each critical path and take the maximum over paths that run concurrently.

% Hence, we model the latency estimation of sequential operations in the same way as the energy and financial cost, replacing cost factors with per-token latency values \( l_i \). First, we estimate the latency of a single LLM invocation as \( Latency(Q, L_i) = l_i (\mathit{Tok}_i(p_t)+ T^{\text{out}}_{op_i}) \). For parallel operations, where multiple LLMs process the same input independently, the estimated latency is determined by the slowest operation:
% \( Latency(Par(Q, L_{1},..., L_{n})) = \max_{op_i \in \pi} (Latency(Q, L_i)).\)

\subsection{Pareto Search over Multi-LLM Plans}\label{sect:optimizers}
The plan space is inherently combinatorial. Under our operator-DAG representation, each node selects one of $|L|$ models and each of the $\binom{k}{2}$ forward edges may be present or absent, yielding
\(
|\Pi| = |L|^{k} \cdot 2^{\binom{k}{2}} = |L|^{k} \cdot 2^{\frac{k(k-1)}{2}}
\), with the derivation provided
\ifextended
next. 
\begin{proof} \emph{(Model assignments)} For each of the $k$ ordered stages, we choose any model in $L$, with reuse permitted. By independence across stages, the number of stage-to-model assignments is \[ N_{\text{assign}} \;=\; |L|^{k}. \] \emph{(Acyclic topologies)} With the node order fixed as $1<\cdots<k$, a directed edge may only go from $i$ to $j$ when $i<j$. There are exactly \[ N_{\text{edges}} \;=\; \binom{k}{2} \;=\; \frac{k(k-1)}{2} \] eligible edges, each chosen independently to be present or absent. Hence, the number of admissible upper-triangular (and therefore acyclic) edge sets is \[ N_{\text{conn}} \;=\; 2^{\binom{k}{2}} \;=\; 2^{\frac{k(k-1)}{2}}. \] \emph{(Independence and product rule)} Model choices and edge choices are independent design dimensions, so \[ |\Pi| \;=\; N_{\text{assign}}\cdot N_{\text{conn}} \;=\; |L|^{k}\cdot 2^{\frac{k(k-1)}{2}}. \] \end{proof} \paragraph{Remarks.} (i) This is a labeled count with execution order $1{<}\dots{<}k$ fixed; it therefore upper-bounds implementations that later canonicalize isomorphic topologies or impose additional syntactic constraints (e.g., requiring a single sink or immediate blending at fan-ins). (ii) The expression makes the combinatorial growth explicit: \[ \log |\Pi| = k\log |L| + \frac{k(k-1)}{2}\log 2, \]
\else
in the extended version~\cite{optiqextended2026}. This space becomes large even for modest settings (e.g., $k{=}5$, $|L|{=}10$ gives $10^{5}\cdot 2^{10}\approx 1.024\times 10^{8}$ plans). Moreover, optimizing over $\Pi$ is NP-hard even for simplified variants of our objective structure~\cite{shekhar2024optimizingcostsllmusage}. Consequently, exhaustive enumeration is infeasible except for small instances, and we instead rely on search procedures that can efficiently explore $\Pi$ using inexpensive, pre-execution objective estimates.
In particular, we consider three widely used approaches: a genetic algorithm (NSGA-II), bottom-up dynamic programming (DP), and randomized hill climbing. All three optimizers operate over the same plan encoding (canonical DAG topology + model assignments), use the same \textsc{PerfDB}-based estimators for objective evaluation, and output a set of non-dominated feasible plans that trade off QoA, cost, latency, and energy under user budgets without enumerating all of $\Pi$. We then select a single plan for execution by normalizing objectives and applying the user weights $W$. 
\fi
\ifextended
we present the core mechanics, full plan-generation details and algorithmic variants below.
\else
Due to space constraints, we present only the core mechanics below; full plan-generation details and algorithmic variants appear in the extended version~\cite{optiqextended2026}.
\fi

\pseudosection{Genetic Algorithm (NSGA-II).}
\ifextended
NSGA-II is a robust default for our discrete, non-convex search space spanning both topology and model assignment. As summarized in Algorithm~\ref{alg:nsga2}, each individual encodes (i) a canonicalized DAG topology (\texttt{connectivity\_map}) and (ii) a model-assignment vector $\mathbf{m}$ specifying one LLM per node. Starting from an initial population of size $N$, we first evaluate all individuals using \textsc{QueryOrPredict} over \textsc{PerfDB} and mark infeasible plans based on the user-specified cost, latency, energy, and QoA constraints (lines~1--7). NSGA-II then iterates for $G$ generations using binary tournament selection based on non-dominated rank and crowding distance (lines~8--11). With crossover probability $C_r$, we apply one-point crossover independently to the topology and assignment segments, followed by targeted mutations that mirror common plan edits: (i) edge flips (rewiring), (ii) node insertions or deletions (operator modification), and (iii) model reassignment on non-blending nodes (lines~12--18). After any structural edit, we apply lightweight repair to enforce validity, including a valid sink/output and the blending constraint that any node with in-degree greater than one must use the blending model. We then canonicalize the resulting plan so that isomorphic workflows map to a single encoding. Offspring are re-evaluated through \textsc{QueryOrPredict} and feasibility checking (lines~20--24), after which the parent and offspring populations are merged and ranked using fast non-dominated sorting with constraint handling, where feasible solutions dominate infeasible ones and infeasible solutions are ordered by total violation (lines~25--27). The first non-dominated front $\mathcal{F}_1$ provides an approximation to the Pareto set, from which we select a single execution plan by applying the user preference vector $W$ to normalized objectives (line~28).
\begin{algorithm}[!htb]
  \footnotesize
  \caption{NSGA-II for multi-objective schedule optimization.}
  \label{alg:nsga2}
  \begin{algorithmic}[1]
    \Require Initial population $\mathcal{P}_0$, population size $N = |\mathcal{P}_0|$, generations $G$, crossover probability $C_r$, \\
            \hspace{1.3em}mutation operators $\mu \leftarrow \{\mu_1: \text{edge mod.}, \mu_2: \text{node mod.}, \mu_3: \text{LLM adj.}\}$, \\
            \hspace{1.3em}constraints $(F_{\max}, L_{\max}, E_{\max}, \mathrm{QoA}_{\min})$, performance database $\mathrm{PerfDB}$, \\
            \hspace{1.3em}user preference vector $W = (w_C, w_L, w_E, w_{\mathrm{QoA}})$
    \Ensure Schedule selected from the first Pareto front based on $W$

    \ForAll{$\pi_i \in \mathcal{P}_0$} \Comment{Initial population evaluation}
        \State $(\mathrm{QoA}_i, \text{Cost}_i, \text{Latency}_i, \text{Energy}_i) \gets$ \Func{QueryOrPredict}$(\pi_i, \Func{PerfDB})$
        \If{$\text{Cost}_i > F_{\max}$ \textbf{OR} $\text{Latency}_i > L_{\max}$ \textbf{OR} $\text{Energy}_i > E_{\max}$ \textbf{OR} $\mathrm{QoA}_i < \mathrm{QoA}_{\min}$}
            \State Mark $\pi_i$ as infeasible
        \EndIf
        \State $\pi_i.\mathrm{obj} \gets (-\text{Cost}_i, -\text{Latency}_i, -\text{Energy}_i, \mathrm{QoA}_i)$ \Comment{Store objectives for minimization}
    \EndFor

    \For{$g \gets 0$ \textbf{TO} $G{-}1$} \Comment{Main evolutionary loop}
        \State $\{\mathcal{F}_1, \mathcal{F}_2, \dots\} \gets$ \Func{FastNonDominatedSort}$(\mathcal{P}_g)$ \Comment{Rank based on dominance}
        \State Calculate \Func{CrowdingDistance}$(\mathcal{F}_k)$ for all fronts $\mathcal{F}_k$ \Comment{Calculate diversity}
        \State $\mathcal{M} \gets$ \Func{BinaryTournamentSelection}$(\mathcal{P}_g, N)$ \Comment{Select mating pool}
        \State $\mathcal{Q}_g \gets \emptyset$ \Comment{Initialize offspring population}

        \While{$|\mathcal{Q}_g| < N$} \Comment{Generate offspring}
            \State $(S_p, S_q) \gets$ \Func{SelectParents}$(\mathcal{M})$
            \State $(O_1, O_2) \gets$
                \begin{cases}
                    \Func{Crossover}$(S_p, S_q)$ & \textbf{if } \Func{rand}() < C_r \\
                    (S_p, S_q)                   & \textbf{otherwise}
                \end{cases}
            \State $O_1 \gets$ \Func{Mutate}$(O_1, \mu)$;\; \Comment{Apply mutation strategies}
            \State $O_2 \gets$ \Func{Mutate}$(O_2, \mu)$;
            \State $\mathcal{Q}_g \gets \mathcal{Q}_g \cup \{O_1, O_2\}$
        \EndWhile

        \ForAll{$\pi_i \in \mathcal{Q}_g$} \Comment{Evaluate offspring}
            \State $(\mathrm{QoA}_i, \text{Cost}_i, \text{Latency}_i, \text{Energy}_i) \gets$ \Func{QueryOrPredict}$(\pi_i, \Func{PerfDB})$ \label{line:reeval}
            \If{$\text{Cost}_i > F_{\max}$ \textbf{OR} $\text{Latency}_i > L_{\max}$ \textbf{OR} $\text{Energy}_i > E_{\max}$ \textbf{OR} $\mathrm{QoA}_i < \mathrm{QoA}_{\min}$}
                 \State Mark $\pi_i$ as infeasible \Comment{Re-check feasibility}
            \EndIf
            \State $\pi_i.\mathrm{obj} \gets (-\text{Cost}_i, -\text{Latency}_i, -\text{Energy}_i, \mathrm{QoA}_i)$ \Comment{Update objectives}
        \EndFor

        \State $\mathcal{R}_g \gets \mathcal{P}_g \cup \mathcal{Q}_g$ \Comment{Combine parent and offspring}
        \State $\{\mathcal{F}_1, \mathcal{F}_2, \dots\} \gets$ \Func{FastNonDominatedSort}$(\mathcal{R}_g)$ \Comment{Rank combined population}
        \State $\mathcal{P}_{g+1} \gets$ \Func{SelectNextGeneration}$(\{\mathcal{F}_k\}, N)$ \Comment{Select based on rank and crowding}
    \EndFor

    \State \Return \Func{SelectByPreference}\bigl(\Func{GetFront}($\mathcal{P}_G, 1$),\,W\bigr)  \Comment{Select from best front using user weights}
  \end{algorithmic}
\end{algorithm}
\else
NSGA-II is a robust default for our discrete, non-convex search space (topology + model choices). Each individual encodes (i) a canonicalized DAG topology (\texttt{connectivity\_map}) and (ii) a model-assignment vector $\mathbf{m}$ (one LLM per node). Starting from an initial population of size $N$, NSGA-II iterates for $G$ generations using binary tournament selection (non-dominated rank, then crowding distance). With crossover probability $C_r$, we apply one-point crossover independently to the topology and assignment segments, then apply targeted mutations that mirror common plan edits: (i) edge flips (rewiring), (ii) node insertions (adding an operator), and (iii) model reassignments (changing $L_i$ on a non-blending node). After any structural edit we apply lightweight repair to enforce validity (e.g., a valid sink/output) and the blending constraint (any node with in-degree $>1$ is assigned the blending model), followed by canonicalization so isomorphic plans share a single encoding. We rank the merged parent+offspring pool with fast non-dominated sorting and constraint handling (feasible dominates infeasible; infeasible ranked by total violation), then select the next generation by taking successive fronts with crowding-distance tie-breaking. The first non-dominated front $\mathcal{F}_1$ approximates the Pareto set; we choose a single execution plan by applying $W$ to normalized objectives over $\mathcal{F}_1$.
\fi

\pseudosection{Dynamic Programming.}
\ifextended
The DP baseline uses the same canonical bit-plan representation, objective definitions, constraints, and final preference-based selection rule as \textsc{NSGA-II}. As summarized in Algorithm~\ref{alg:dp}, DP starts from a set of seed plans and incrementally constructs larger plans up to depth $k$ by applying valid structural and model-assignment edits. Each candidate is evaluated using \textsc{QueryOrPredict} over \textsc{PerfDB}, with the same objective tuple $(\text{Cost}, \text{Latency}, \text{Energy}, \text{QoA})$ and the same user-specified bounds $(F_{\max}, L_{\max}, E_{\max}, \mathrm{QoA}_{\min})$ used throughout the framework. 

To control state growth, we apply configurable post-expansion pruning at each depth (Algorithm~\ref{alg:dp}, lines~18--23). Our implementation supports three modes: no pruning, parent-specific $\Delta$-QoA progress pruning, and approximate dominance pruning via multiplicative $\varepsilon$-dominance following~\cite{trummer2014approximation}. Table~\ref{tab:dp-pruning} summarizes the resulting runtime--quality trade-offs. Unless otherwise stated, we use the parent-specific $\Delta$-QoA rule by default, retaining a child only if
\(
\mathit{QoA}(\pi_{\text{child}}) \;\ge\; \mathit{QoA}(\pi_{\text{parent}}) + \Delta,
\)
with $\Delta = 0.05$. This setting provided a strong runtime--quality trade-off in our analysis, substantially reducing runtime while preserving high hypervolume. After canonical deduplication and pruning, the final plan is selected from the first Pareto front using the user preference vector $W$.
\begin{algorithm}[!htb]
  \footnotesize
  \caption{Dynamic programming for multi-objective schedule optimization.}
  \label{alg:dp}
  \begin{algorithmic}[1]
    \Require Seed plans $\mathcal{S}_0$, maximum operations $k$, pruning mode $P \in \{\texttt{none}, \texttt{delta\_qoa}, \texttt{epsilon}\}$, \\
            \hspace{1.3em}optional pruning parameter $\Delta$ or $\varepsilon$, constraints $(F_{\max}, L_{\max}, E_{\max}, \mathrm{QoA}_{\min})$, \\
            \hspace{1.3em}performance database $\mathrm{PerfDB}$, preference vector $W = (w_C, w_L, w_E, w_{\mathrm{QoA}})$
    \Ensure Schedule selected from the final Pareto set according to $W$

    \State $\mathcal{B}_1 \gets \mathcal{S}_0$ \Comment{Initialize bucket of 1-operation plans}

    \For{$d \gets 1$ \textbf{TO} $k$}
        \ForAll{$\pi_i \in \mathcal{B}_d$}
            \State $(\mathrm{QoA}_i, \mathrm{Cost}_i, \mathrm{Latency}_i, \mathrm{Energy}_i) \gets$ \Func{QueryOrPredict}$(\pi_i,\mathrm{PerfDB})$
            \If{$\mathrm{Cost}_i > F_{\max}$ \textbf{OR} $\mathrm{Latency}_i > L_{\max}$ \textbf{OR} $\mathrm{Energy}_i > E_{\max}$ \textbf{OR} $\mathrm{QoA}_i < \mathrm{QoA}_{\min}$}
                \State Mark $\pi_i$ as infeasible
            \EndIf
            \State $\pi_i.\mathrm{obj} \gets (-\mathrm{Cost}_i, -\mathrm{Latency}_i, -\mathrm{Energy}_i, \mathrm{QoA}_i)$
        \EndFor

        \If{$d = k$}
            \State \textbf{break}
        \EndIf

        \State $\mathcal{B}_{d+1} \gets \emptyset$

        \ForAll{$\pi_p \in \mathcal{B}_d$} \Comment{Expand all parents at depth $d$}
            \State $\mathcal{C} \gets$ \Func{Expand}$(\pi_p)$
            \ForAll{$\pi_c \in \mathcal{C}$}
                \State $\pi_c \gets$ \Func{RepairAndCanonicalize}$(\pi_c)$
                \State $(\mathrm{QoA}_c, \mathrm{Cost}_c, \mathrm{Latency}_c, \mathrm{Energy}_c) \gets$ \Func{QueryOrPredict}$(\pi_c,\mathrm{PerfDB})$
                \If{$\mathrm{Cost}_c > F_{\max}$ \textbf{OR} $\mathrm{Latency}_c > L_{\max}$ \textbf{OR} $\mathrm{Energy}_c > E_{\max}$ \textbf{OR} $\mathrm{QoA}_c < \mathrm{QoA}_{\min}$}
                    \State \textbf{continue}
                \EndIf
                \State $\pi_c.\mathrm{obj} \gets (-\mathrm{Cost}_c, -\mathrm{Latency}_c, -\mathrm{Energy}_c, \mathrm{QoA}_c)$
                \State $\mathcal{B}_{d+1} \gets \mathcal{B}_{d+1} \cup \{(\pi_p,\pi_c)\}$ \Comment{Keep parent-child link for pruning}
            \EndFor
        \EndFor

        \State $\mathcal{B}_{d+1} \gets$ \Func{DeduplicateCanonical}$(\mathcal{B}_{d+1})$

        \If{$P = \texttt{delta\_qoa}$}
            \State $\mathcal{B}_{d+1} \gets$ \Func{DeltaQoAPrune}$(\mathcal{B}_{d+1}, \Delta)$
        \ElsIf{$P = \texttt{epsilon}$}
            \State $\mathcal{B}_{d+1} \gets$ \Func{EpsilonPrune}$(\mathcal{B}_{d+1}, \varepsilon)$
        \EndIf

        \State $\mathcal{B}_{d+1} \gets$ \Func{ParetoPrune}$(\mathcal{B}_{d+1})$
    \EndFor

    \State $\mathcal{R} \gets \bigcup_{d=1}^{k} \mathcal{B}_d$
    \State $\mathcal{F}_1 \gets$ \Func{GetFront}$(\mathcal{R}, 1)$
    \State \Return \Func{SelectByPreference}$(\mathcal{F}_1, W)$
  \end{algorithmic}
\end{algorithm}
\begin{table}[t]
\centering
\caption{Runtime--quality trade-off of DP pruning strategies at $k=5$, averaged over 10 topics. Hypervolume (HV) is reported as a percentage of the no-pruning baseline.}
\label{tab:dp-pruning}
\setlength{\tabcolsep}{4pt}
\renewcommand{\arraystretch}{1.15}
\begin{tabular}{c cc cc cc}
\toprule
\multirow{2}{*}{$\varepsilon=\Delta$} & \multicolumn{2}{c}{No pruning} & \multicolumn{2}{c}{$\varepsilon$-dominance} & \multicolumn{2}{c}{$\Delta$-QoA gate} \\
\cmidrule(lr){2-3}\cmidrule(lr){4-5}\cmidrule(lr){6-7}
& HV (\%) & Time (s) & HV (\%) & Time (s) & HV (\%) & Time (s) \\
\midrule
0.01 & 100.0 & 227.5 & 95.5 & 22.8 & 96.3 & 14.5 \\
0.02 & 100.0 & 227.5 & 95.3 & 20.2 & 95.8 & 9.5 \\
0.05 & 100.0 & 227.5 & 93.3 & 11.6 & 92.3 & 2.1 \\
0.10 & 100.0 & 227.5 & 90.3 & 7.7  & 78.8 & 0.2 \\
0.20 & 100.0 & 227.5 & 83.0 & 3.7  & 68.9 & 0.0 \\
\bottomrule
\end{tabular}
\end{table}

\else
DP provides a systematic baseline by enumerating canonical plan structures and propagating feasible assignment sets per structure. For plans with $k$ operations, the DP state is indexed by the canonical structure identifier \texttt{connectivity\_map}. Each bucket stores multiple candidate assignments $\mathbf{m}$ (one model per node) together with plan-level estimated costs computed via \textsc{PerfDB}-based estimators. Blending is enforced by construction: nodes with in-degree $>1$ must use the designated blending model, while nodes with in-degree $\le 1$ may use any base model. To build $DP[k]$ from $DP[k{-}1]$, we extend each canonical $(k{-}1)$-node structure by adding a new node and enumerating predecessor subsets $S\subseteq\{0,\ldots,k{-}2\}$, adding edges $(u\rightarrow k{-}1)$ for all $u\in S$. For each parent assignment, we enumerate admissible model choices for the new node, evaluate objectives, and canonicalize to merge isomorphic structures into same bucket.

To control state growth, we use pruning. In particular, we found a simple \emph{parent-specific $\Delta$-QoA progress gate} to provide the best runtime--quality trade-off: when expanding a parent into children, we retain a child only if
\(
\mathit{QoA}(\pi_{\text{child}}) \;\ge\; \mathit{QoA}(\pi_{\text{parent}}) + \Delta \).
We use $\Delta=0.05$ by default. (We also explored within-bucket approximate dominance pruning via multiplicative $\varepsilon$-dominance; see the extended version~\cite{optiqextended2026}.)
\fi

\pseudosection{Hill Climbing.}
\ifextended
Because DP can become state-intensive as $k$ grows, we also implement a randomized local-search optimizer that quickly finds high-quality trade-offs under a per-question time budget, following~\cite{trummer2016fast}. As summarized in Algorithm~\ref{alg:hc}, the method performs repeated random restarts: each restart samples a feasible plan $\pi$, then greedily hill-climbs through local plan edits until reaching a locally non-dominated solution. Collecting these local optima across restarts yields an approximate Pareto frontier.

The optimizer uses the same canonical bit-plan representation, objective definitions, constraints, and final preference-based selection rule as \textsc{NSGA-II} and DP. The neighborhood of a plan $\pi$ is defined by exhaustive single-step mutations: (i) \emph{model mutation}, which changes the base model at one non-blending node; (ii) \emph{edge flip}, which toggles one edge in the upper-triangular encoding and retains the result only if it remains a valid DAG; and (iii) \emph{node addition}, which appends one new node (up to the operation limit $k$) using an admissible extension that preserves a valid sink/output. After any structural mutation, we repair assignments to enforce the blending constraint, canonicalize the topology, and remap $\mathbf{m}$ under the canonical permutation so that isomorphic DAGs share a single encoding.

Each candidate is evaluated using \textsc{QueryOrPredict} over \textsc{PerfDB} and filtered using the same hard bounds $(F_{\max}, L_{\max}, E_{\max}, \mathrm{QoA}_{\min})$ as the other planners. A neighbor is accepted if it dominates the current plan, i.e., it is no worse on all objectives and strictly better on at least one. We scan the neighborhood and take the first dominating move, repeating until no neighbor dominates the current plan (Algorithm~\ref{alg:hc}, lines~9--19). As locally optimal candidates are discovered, we maintain a global frontier across restarts; optionally, this insertion step can apply the same parent-specific $\Delta$-QoA filter used in DP.
\begin{algorithm}[!htb]
  \footnotesize
  \caption{Hill climbing for multi-objective schedule optimization.}
  \label{alg:hc}
  \begin{algorithmic}[1]
    \Require Time budget $T$, maximum operations $k$, constraints $(F_{\max}, L_{\max}, E_{\max}, \mathrm{QoA}_{\min})$, \\
            \hspace{1.3em}performance database $\mathrm{PerfDB}$, preference vector $W = (w_C, w_L, w_E, w_{\mathrm{QoA}})$
    \Ensure Schedule selected from the final approximate Pareto frontier according to $W$

    \State $\mathcal{F} \gets \emptyset$ \Comment{Global frontier}
    \While{\Func{TimeRemaining}$(T)$}
        \State $\pi \gets$ \Func{SampleFeasiblePlan}$(k)$ \Comment{Random restart}
        \State $(\mathrm{QoA}, \mathrm{Cost}, \mathrm{Latency}, \mathrm{Energy}) \gets$ \Func{QueryOrPredict}$(\pi,\mathrm{PerfDB})$
        \State $\pi.\mathrm{obj} \gets (-\mathrm{Cost}, -\mathrm{Latency}, -\mathrm{Energy}, \mathrm{QoA})$

        \State improved $\gets$ \textbf{true}
        \While{improved}
            \State improved $\gets$ \textbf{false}
            \State $\mathcal{N} \gets$ \Func{GenerateNeighborhood}$(\pi, k)$
            \ForAll{$\pi' \in \mathcal{N}$}
                \State $\pi' \gets$ \Func{RepairAndCanonicalize}$(\pi')$
                \State $(\mathrm{QoA}', \mathrm{Cost}', \mathrm{Latency}', \mathrm{Energy}') \gets$ \Func{QueryOrPredict}$(\pi',\mathrm{PerfDB})$
                \If{$\mathrm{Cost}' > F_{\max}$ \textbf{OR} $\mathrm{Latency}' > L_{\max}$ \textbf{OR} $\mathrm{Energy}' > E_{\max}$ \textbf{OR} $\mathrm{QoA}' < \mathrm{QoA}_{\min}$}
                    \State \textbf{continue}
                \EndIf
                \State $\pi'.\mathrm{obj} \gets (-\mathrm{Cost}', -\mathrm{Latency}', -\mathrm{Energy}', \mathrm{QoA}')$
                \If{\Func{Dominates}$(\pi', \pi)$}
                    \State $\pi \gets \pi'$ \Comment{First improving move}
                    \State improved $\gets$ \textbf{true}
                    \State \textbf{break}
                \EndIf
            \EndFor
        \EndWhile

        \State $\mathcal{F} \gets$ \Func{InsertIntoFrontier}$(\mathcal{F}, \pi)$
    \EndWhile

    \State $\mathcal{F}_1 \gets$ \Func{GetFront}$(\mathcal{F}, 1)$
    \State \Return \Func{SelectByPreference}$(\mathcal{F}_1, W)$
  \end{algorithmic}
\end{algorithm}

\else
DP can become state-intensive as $k$ grows, so we also implement a randomized local-search optimizer that quickly finds high-quality tradeoffs under a per-question time budget, following~\cite{trummer2016fast}. The method performs repeated random restarts: each restart samples a feasible plan $\pi$, then greedily hill-climbs using \emph{Pareto-improving} moves until reaching a locally non-dominated solution; collecting these local optima across restarts yields an approximate frontier. We define the neighborhood of $\pi$ via exhaustive single-step mutations: (i) \emph{model mutation} changes the base model at one non-blending node, (ii) \emph{edge flip} toggles one edge in the upper-triangular encoding and keeps the result only if it remains a valid DAG, and (iii) \emph{node addition} appends one new node (up to the operation limit $k$) using an admissible extension that preserves a valid sink/output. After any structural mutation we repair assignments to satisfy the blending constraint, canonicalize the topology, and remap $\mathbf{m}$ under the canonical permutation so isomorphic DAGs share a single encoding. We discard neighbors whose estimated metrics violate the planning-time budgets. A neighbor is accepted if it dominates the current plan (no worse on all objectives and strictly better on at least one); we scan the neighborhood and take the first dominating move, repeating until no neighbor dominates the current plan. As candidates are discovered we maintain a global frontier (optionally applying the same $\Delta$-QoA filter from before when inserting plans).

\section{Opti-Q Implementation}
\label{sect:framework}
% \begin{figure}[t]  % one-column float (not figure*)
%   \centering
%   \includegraphics[width=\columnwidth]{figures/Final_Overall.png}
%   \caption{High-level overview of \system's implementation.}
%   \Description{High-level overview of \system's implementation}
%   \label{fig:arch}
% \end{figure}
We implement \system as a modular framework for flexible multi-LLM QA planning under user-defined constraints\footnote{Code, datasets, prompts, and results are available at~\url{https://github.com/Aamir7693/Opti-Q.git}.}. Our prototype integrates five widely used open-source LLMs:  Gemma-3:27B~\cite{gemma_2024}, LLaMA3-ChatQA (8B)\cite{liu2024chatqa}, Qwen2.5 (14B)\cite{qwen2025qwen25technicalreport}, Phi-4 (14B)\cite{abdin2024phi4technicalreport}, and Mistral (7B)\cite{jiang2023mistral7b}. We selected these models for their complementary strengths across benchmarks~\cite{ye2024benchmarking,wei2024measuringshortformfactualitylarge} (e.g., strong reasoning vs.\ conversational QA vs.\ efficiency). All models run locally via Ollama\footnote{ \url{https://ollama.com}}, which also simplifies adding or replacing models. For tuning system parameters, we construct a held-out configuration set by sampling 100 SimpleQA and 100 MMLU-Pro questions; this set is strictly disjoint from the test set in Section~\ref{sect:expr} to prevent leakage.

\pseudosection{Blending Model.}
\system supports pluggable blending models for combining outputs from parallel branches. We evaluated both the GenFuser component from LLM-Blender~\cite{jiang2023llmblenderensemblinglargelanguage} and a general-purpose LLM-based blender. In our experiments, Gemma-3:27B produced higher validation QoA than GenFuser on the held-out configuration set described above, so we use Gemma-3:27B as the designated blending model. The blender receives the original question, a blending prompt $(p_{bld})$, and the concatenated branch outputs, i.e., $(p_{bld} \oplus \text{Par}(Q,\dots))$, and produces a single consolidated answer.

\pseudosection{Question Processing and \textsc{PerfDB} Initialization.}
\system initializes \textsc{PerfDB} with per-model and per-topic planning parameters, including model size and empirical estimates of cost, latency, energy, and QoA. These priors are populated from public benchmarking sources, including ML.ENERGY~\cite{ml-energy-leaderboard} and Artificial Analysis\footnote{\url{https://artificialanalysis.ai/}}, and are refined as additional execution traces are collected.

At question time, \system extracts topics for an incoming question $Q$ using BERTopic~\cite{grootendorst2022bertopic}. Concretely, we embed $Q$ with a pretrained transformer encoder, reduce dimensionality with UMAP~\cite{mcinnes2020umapuniformmanifoldapproximation}, cluster with HDBSCAN~\cite{mcinnes2017hdbscan}, and derive representative topic keywords using c-TF-IDF~\cite{grootendorst2022bertopic}. The inferred topics are then added to the question tuple used by the planners and \textsc{PerfDB} lookups. 
% We first encode $Q$ with pre-trained transformed model into a dense vector, apply UMAP~\cite{mcinnes2020umapuniformmanifoldapproximation} for dimensionality reduction while preserving key semantic relationships. Then, we cluster embeddings via HDBSCAN~\cite{mcinnes2017hdbscan} into semantically coherent topic groups. Representative topic keywords  are derived using a modified class-based TF-IDF (c-TF-IDF) method~\cite{grootendorst2022bertopic}. The resulting topics are incorporated to the question tuple used by \system to guide optimized plan selection.

\pseudosection{Prompt Design.}
We consider four prompting strategies: \textit{Zero-Shot (ZS)~\cite{kojima2022large}},
% which provides only the question with minimal guidance, 
\textit{Few-Shot (FS)~\cite{brown2020language}},
% which augments the question with exemplars, 
and \textit{Chain-of-Thought (CoT)~\cite{wei2022chain}},
% which explicitly requires intermediate reasoning steps,
and a \textit{baseline} strategy (B) (i.e., just instructions without examples or reasoning). We also define two auxiliary prompts: a \textit{context prompt} $pt_{ctx}$ for sequential operations (ensuring the successor treats the predecessor output as evidence) and a \textit{blending prompt} $pt_{bld}$ for parallel operations (instructing the blender to reconcile redundant or conflicting answers). 
\ifextended
Table~\ref{tab:iqr_stats_test} summarizes the results of the prompting evaluation. \textit{CoT} achieved the highest QoA (median $0.25$, $+40$–$50\%$ over \textit{B}), particularly on deduction-heavy domains (e.g., History, Politics), but shows the largest variability (IQR $=0.27$). \textit{ZS} and \textit{FS} delivered moderate, more consistent gains (median QoA $0.22$ and $0.19$ vs. $0.18$ for \textit{B}) with lower variability (IQR $\approx 0.23$). All strategies degrade on more subjective domains (e.g., Sports, Music). Prompting also affect overheads: \textit{FS} was the most expensive (average cost $\$0.128$, latency $136.2$ s, energy $113.7$ J); \textit{ZS} was the most efficient (average $\$0.045$, $49$ s, $39.5$ J); \textit{CoT} and \textit{B} fall in between ($\sim\$0.05$, $\sim53$ s, $\sim43$ J). Within \system, these trade-offs motivate considering both CoT (higher quality, higher variance) and ZS (lower cost, higher stability). In our experiments, we use ZS since it offers the best trade-off between stability and efficiency, aligning with \system’s emphasis on mitigating the effects of inherent LLM nondeterminism.
\begin{table}[htbp] % Use standard float placement
    \centering
    \small % Control font size if needed
    \caption{Performance, QoA, of different prompting strategies.}
    \label{tab:iqr_stats_test} % Use a unique label for testing
    \vspace{-.1cm}
    \begin{tabular}{lcccccc} % l=left, c=center for stats
        \toprule
        \textbf{Prompting Strategy} & \textbf{Mean} & \textbf{Median} &  \textbf{Std Dev} \\
        \midrule
        Few-Shot (FS)  & 0.23 & 0.19 & 0.19  \\
        Zero-Shot (ZS) & 0.26 & 0.22 & 0.19 \\
        Chain-of-Thought (CoT) & 0.30 & 0.25 & 0.21 \\
        Baseline (B)  & 0.22 & 0.18 & 0.20 \\
        \bottomrule
    \end{tabular}
\end{table}

\else
Our prompting evaluation shows that \textit{ZS} provides the best trade-off between efficiency and stability (see the extended version~\cite{optiqextended2026}). We therefore use \textit{ZS} in the experiments of Section~\ref{sect:expr}.

\pseudosection{QoA Computation.}
We measure QoA using semantic similarity rather than lexical overlap (e.g., BLEU~\cite{papineni-etal-2002-bleu}, ROUGE~\cite{lin-2004-rouge}), since LLM outputs often vary stylistically while preserving meaning. For free-form QA, we embed generated and reference answers with the 384-dimensional  \textit{all-MiniLM-L6-v2} sentence transformer\footnote{\url{https://huggingface.co/sentence-transformers/all-MiniLM-L6-v2}}, which has demonstrated strong performance on the Massive Text Embedding Benchmark (MTEB)\footnote{\url{https://huggingface.co/spaces/mteb/leaderboard}} and is widely adopted.
% , with over 168 million downloads per month. 
The QoA is then computed as the cosine similarity between embeddings. For multiple-choice QA, we score single-answer questions (e.g., MMLU-Pro) by normalized exact match (binary in $\{0,1\}$). For multi-answer (``select all that apply'') questions, we normalize the predicted and gold option sets and compute a set-overlap score in $[0,1]$ (e.g., Jaccard similarity) so partially correct predictions receive partial credit.

\pseudosection{NSGA-II Hyperparameter Optimization.}
\ifextended
We tune our NSGA-II scheduler via grid search over
mutation rates $\boldsymbol{\mu}=\{\mu_1,\mu_2,\mu_3\}$ for (i) edge modifications, (ii) node adjustments, and (iii) LLM-driven adaptations, as well as population size ($\boldsymbol{P}$) and number of generations ($\boldsymbol{G}$), guided by prior recommendations~\cite{hassanat2019choosing}. 
Each configuration runs NSGA-II on a fixed set of 200 questions sampled from the benchmark used in our evaluation (more on this in Section~\ref{sect:expr}) without budget constraints. Following our protocol (Section~\ref{subsection:O-model}), \emph{QoA} is maximized and \emph{cost}, \emph{latency} and \emph{energy} are minimized. For cross-configuration comparison, we report a scalarized score that linearly combines QoA with normalized cost/latency/energy (each metric divided by its maximum observed value across the grid and weighted equally).\footnote{Within runs, we use standard NSGA-II non-dominated sorting and crowding distance; the scalarized score is only for comparing \emph{hyperparameter settings} across runs.} We evaluated population sizes $P\in\{75,100,200\}$, generations $G\in\{75,100,200\}$, and mutation rates $\mu_i\in\{0.1,0.3\}$ for all three mutation types.
The best-performing configuration by mean scalarized score in the converged population was $P=200$, $G=200$, and $\boldsymbol{\mu}=\{0.3,0.1,0.3\}$, which consistently outperformed other settings by $1.2\times$–$12\times$ across validation metrics adopting this configuration for all subsequent experiments.
\else
We tuned NSGA-II hyperparameters via grid search on a held-out stratified set (200 questions); we report the best setting and release configs/scripts. The best-performing configuration by mean scalarized score in the converged population was $P=200$, $G=200$, and $\boldsymbol{\mu}=\{0.3,0.1,0.3\}$, which outperformed other settings by $1.2\times$–$12\times$ across validation metrics, and we adopt this configuration for all subsequent experiments.
\fi

\section{Experimental Evaluation}
\label{sect:expr}

\noindent\textbf{Evaluation setup.}
Our evaluation targets two foundational QA paradigms using specialized benchmarks: \textsc{SimpleQA}, an open-ended factual QA dataset containing 4K questions, and \textsc{MMLU-Pro}, a domain-specific dataset comprising 12K multiple choice questions. Because exhaustive evaluation would entail ($\approx$3.2M plan executions across five models), we instead build a representative, computationally tractable testbed using stratified sampling: we select 10 diverse question types from each dataset (e.g., \emph{History}, \emph{Mathematics}, \emph{Sports}) and uniformly sample 10 questions per type, yielding 200 questions total (100 per benchmark) spanning factual, analytical, and reasoning-intensive tasks. Even under this stratified design, the evaluation remained computationally intensive, requiring roughly 1,000 GPU-hours overall.
Experiments run on a dedicated high-performance cluster equipped with $2\times$24-core Intel Xeon Gold 6240R (2.40\,GHz, 165\,W TDP) CPUs and x4 NVIDIA RTX~6000 GPUs, with shared central storage exceeding 3\,PB.
To account for variability in LLM outputs, we execute each question five times under fixed decoding configurations and report micro-averaged QoA, latency, energy, and financial cost with 95\% confidence intervals.
% To maintain computational feasibility, we randomly sampled 100 questions from each dataset, totaling 200 questions. Given five models and a maximum of five operations per question, our evaluation involved approximately 3.2 million plan executions, conducted on two dedicated high-performance computing clusters, one GPU-intensive and one CPU-intensive, with shared central storage exceeding 3 petabytes. The evaluation covers 20 question types (e.g. \emph{History}, \emph{Geography}, \emph{Mathematics}, \emph{Sports}, \emph{Science}) with 10 questions per type, producing a balanced and domain-rich testbed that covers factual, analytical, and reasoning-intensive tasks.
% We compare methods under matched prompts, GA hyperparameters, and budgets accounting to ensure a controlled, budget-aware assessment.

\noindent\textbf{Planner Comparison.} We compare \system's planner backends---NSGA-II, exact DP, pruned DP, Hill Climbing, and pruned Hill Climbing---as the maximum number of operations $k$ increases. The full comparison table is provided in 
\ifextended
\begin{table}[t]
\centering
\caption{Algorithm comparison across DAG sizes $k=1$--$5$, averaged over 10 topics (3 runs for stochastic algorithms). HV\% and IGD are computed w.r.t.\ the DP Pareto set when DP is tractable. Best values per $k$ are in \textbf{bold} (excluding the DP reference for HV\% and IGD). $\downarrow$: lower is better, $\uparrow$: higher is better.}
\label{tab:algorithm-comparison}
\footnotesize
\setlength{\tabcolsep}{3pt}
\renewcommand{\arraystretch}{1.05}
\begin{tabular}{clrrr}
\toprule
$k$ & Alg. & Time & HV\% & IGD \\
\midrule
\multirow{5}{*}{1}
& NSGA-II         & $1.3 \pm 0.0$  & $\mathbf{100.0 \pm 0.0}$ & $\mathbf{0.0000 \pm 0.0000}$ \\
& DP              & $\mathbf{0.0}$ & $100.0$                  & $0.0000$ \\
& DP+$\Delta$     & $\mathbf{0.0}$ & $100.0$                  & $0.0000$ \\
& HC              & $1.3 \pm 0.0$  & $\mathbf{100.0 \pm 0.0}$ & $\mathbf{0.0000 \pm 0.0000}$ \\
& HC+$\Delta$     & $1.3 \pm 0.0$  & $100.0 \pm 0.0$          & $0.0136 \pm 0.0000$ \\
\midrule
\multirow{5}{*}{2}
& NSGA-II         & $3.4 \pm 0.0$  & $\mathbf{100.0 \pm 0.0}$ & $\mathbf{0.0000 \pm 0.0000}$ \\
& DP              & $\mathbf{0.0}$ & $100.0$                  & $0.0000$ \\
& DP+$\Delta$     & $\mathbf{0.0}$ & $99.6$                   & $0.0953$ \\
& HC              & $3.4 \pm 0.0$  & $\mathbf{100.0 \pm 0.0}$ & $\mathbf{0.0000 \pm 0.0000}$ \\
& HC+$\Delta$     & $3.4 \pm 0.0$  & $99.3 \pm 0.0$           & $0.1907 \pm 0.0000$ \\
\midrule
\multirow{5}{*}{3}
& NSGA-II         & $10.2 \pm 0.1$ & $\mathbf{100.0 \pm 0.0}$ & $\mathbf{0.0000 \pm 0.0000}$ \\
& DP              & $\mathbf{0.4}$ & $100.0$                  & $0.0000$ \\
& DP+$\Delta$     & $0.1$          & $97.7$                   & $0.0955$ \\
& HC              & $10.3 \pm 0.0$ & $99.7 \pm 0.3$           & $0.0130 \pm 0.0076$ \\
& HC+$\Delta$     & $10.3 \pm 0.0$ & $97.3 \pm 0.3$           & $0.1690 \pm 0.0334$ \\
\midrule
\multirow{5}{*}{4}
& NSGA-II         & $10.1 \pm 0.1$ & $\mathbf{100.0 \pm 0.0}$ & $\mathbf{0.0001 \pm 0.0000}$ \\
& DP              & $8.0$          & $100.0$                  & $0.0000$ \\
& DP+$\Delta$     & $\mathbf{0.4}$ & $94.4$                   & $0.1191$ \\
& HC              & $10.2 \pm 0.0$ & $93.2 \pm 1.0$           & $0.0824 \pm 0.0159$ \\
& HC+$\Delta$     & $10.4 \pm 0.3$ & $91.2 \pm 1.1$           & $0.1575 \pm 0.0285$ \\
\midrule
\multirow{5}{*}{5}
& NSGA-II         & $21.0 \pm 0.3$ & $\mathbf{99.9 \pm 0.1}$  & $\mathbf{0.0043 \pm 0.0016}$ \\
& DP              & $208.5$        & $100.0$                  & $0.0000$ \\
& DP+$\Delta$     & $\mathbf{1.9}$ & $92.3$                   & $0.1123$ \\
& HC              & $21.1 \pm 0.1$ & $78.4 \pm 4.4$           & $0.1078 \pm 0.0175$ \\
& HC+$\Delta$     & $21.1 \pm 0.1$ & $76.9 \pm 4.3$           & $0.1352 \pm 0.0247$ \\
\bottomrule
\end{tabular}
\end{table}
\else
the extended version~\cite{optiqextended2026}. 
\fi
For each $k \in \{1,\dots,5\}$, we measure (i) planning time and (ii) Pareto-front quality using hypervolume (HV) and inverted generational distance (IGD), computed relative to the exact DP frontier (when tractable). For fair comparison under matched computational budgets, the hill-climbing planner is run with a per-question timeout equal to the maximum wall-clock time observed for NSGA-II on the same question under the same operation limit. Exact DP recovers the reference frontier but exhibits state-space explosion as $k$ grows (from near-zero planning time at small $k$ to over 200~s at $k{=}5$). Pruned DP substantially reduces planning time (down to seconds at $k{=}5$) at the cost of lower frontier quality. Hill Climbing, under the same time budget as NSGA-II, degrades more sharply at larger $k$ due to local optima; adding pruning further reduces frontier quality with little runtime benefit under timeout-bounded search. NSGA-II provides the best scalability--quality trade-off, maintaining near-reference frontier quality while keeping planning time in the low tens of seconds at $k{=}5$ (21~s), motivating it as the default backend for larger plan spaces. Unless noted otherwise, in subsequent experiments, we use pruned DP as the default DP-based planner and Hill Climbing with strict Pareto dominance (i.e., without the $\Delta$-QoA filter) as the default Hill Climbing variant, with per-question timeout matched to NSGA-II for the same question and operation limit, based on the above runtime--quality trade-offs.

\noindent\textbf{Baselines and scope.}
\system is designed for broad open-domain QA, while most existing cost-aware multi-LLM methods target classification tasks. To enable a comparison on MMLU-Pro, we treat each question as a classification and compare \system against \textsc{ThriftLLM}~\cite{10.14778/3749646.3749702}, \textsc{LLM-Ensemble}~\cite{chen2025harnessingmultiplelargelanguage}, \textsc{FrugalGPT}~\cite{chen2023frugalgptuselargelanguage} and \textsc{LLM-Blender}~\cite{jiang2023llmblenderensemblinglargelanguage}.
Because \textsc{LLM-Ensemble} lacks a budget constraint, we implement a budgeted variant that greedily selects top-\(K\) weighted models until the budget is met~\cite{chen2025harnessingmultiplelargelanguage}. For \textsc{LLM-Blender}, we adopt the latest \emph{PairRanker} and \emph{GenFuser} components with the author-recommended settings. 
On SimpleQA, we compare \system to \textsc{FrugalGPT} and \textsc{LLM-Blender} using the released configurations. Unless noted, all methods (\system and baselines) were evaluated under identical budget constraints. We omit \textit{LLM-TopLa} (repository errors), Shekhar et~al.\ (summarization task), and \textit{Palimpzest} (declarative AI analytics pipelines over document corpora) as they are not directly comparable under our QA-oriented orchestration setting.

\subsection{Baseline Comparison}
We evaluate all baseline systems under a unified cost–accuracy framework. Each method is configured to adhere to the same strict per-question budget as \system. However, \textit{LLM-Blender} is not budget-aware and invokes its full model set for every query, leading to different average costs across questions: 
\(\approx4.4{\times}10^{-5}\)~USD/query for the \textit{MMLU-Pro} classification benchmark and \(\approx1.6{\times}10^{-5}\)~USD/query for the \textit{SimpleQA} open-ended benchmark. 
Hence, it is shown in our figures for completeness but excluded from cost–accuracy comparisons, and compared against \system solely on the basis of QoA. Furthermore, we adapt FrugalGPT to operate under our strict per-query budget constraint rather than its native expected-cost scheduling mechanism to align its behavior with other baselines. 
All baseline systems were re-implemented using our selected model suite and benchmark datasets; we denote these adapted versions with an asterisk (e.g., \textit{FrugalGPT*}, \textit{ThriftLLM*}). 
Hereafter, all baseline names refer to these re-implemented variants. The evaluation spans five budget levels, \(b \in \{1,\ldots,5\}\), each scaling a baseline single-model usage cost of \(1.2\times10^{-5}\). 
This value represents the average per-question execution cost across all candidate LLMs. 
Thus, the maximum allowable financial cost per question is defined as \(F_{\text{max}}(b)=b\cdot1.2\times10^{-5}\).
\paragraph{Classification-reasoning QA (MMLU-PRO)}
 \begin{figure}[htb]
    \centering
    \includegraphics[width=1\linewidth]{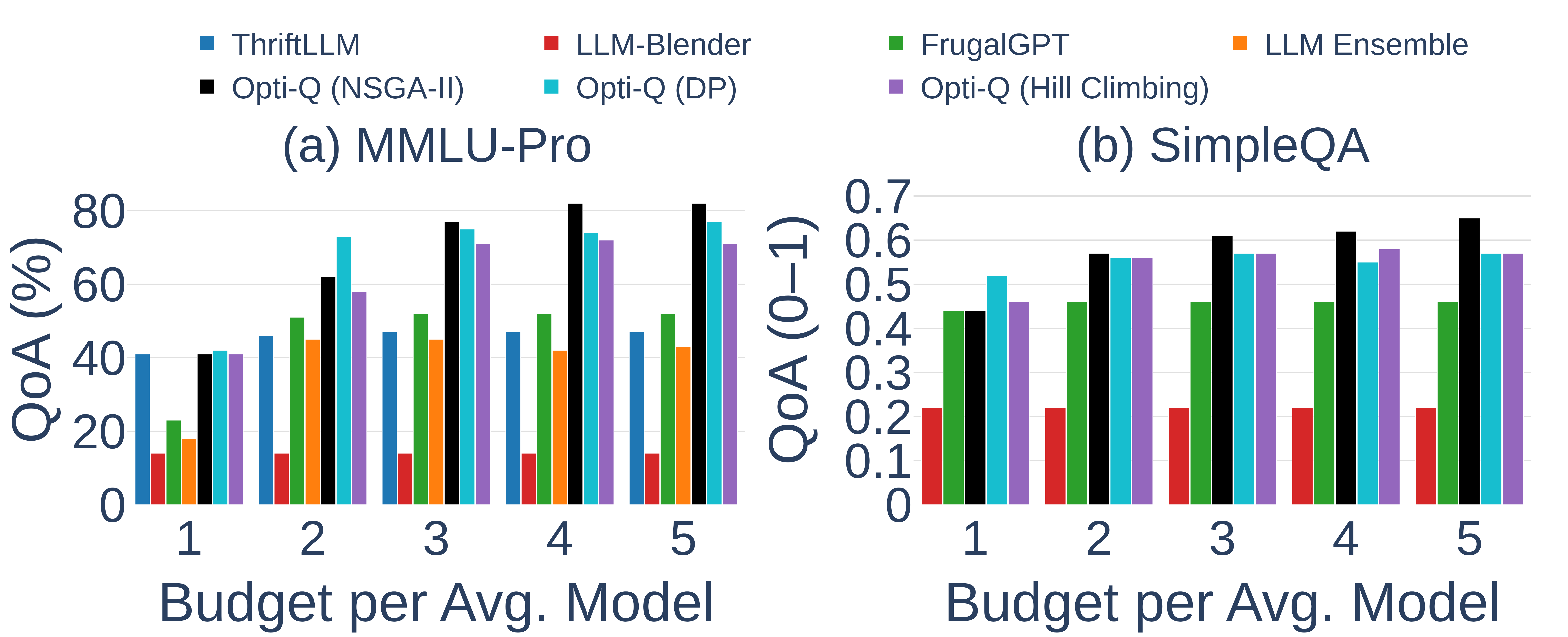}
    \caption{QoA versus budget across two benchmarks: (a) \textit{MMLU-Pro}, where QoA is measured as the percentage of questions answered correctly, and (b) \textit{SimpleQA}, where QoA is measured as the average quality per question.}
    \label{fig:mmlu_classif}
\end{figure}

We first evaluate \system on the \textit{MMLU-Pro} benchmark, a multiple-choice QA task where each question has fixed answer options, and performance is measured by accuracy under varying budget constraints. At the lowest budget (\(b{=}1\)), as shown in Figure~\ref{fig:mmlu_classif}(a), \textit{ThriftLLM*} is competitive with the \system planner variants (about 41\%), due to its routing mixture between Mistral and Qwen models and its per-question ensemble aggregation, which balance cost and model capability. \system (DP) is slightly higher at approximately 42\%. As the budget increases (\(b{=}2{-}3\)), all baseline systems converge around 45--47\% accuracy and subsequently plateau. This stagnation arises because these methods execute all models in parallel and deterministically collapse their predictions into a single label via confidence ranking or majority voting, thereby discarding contextual and complementary evidence among models. Consequently, additional model invocations increase the cost without improving the accuracy. In contrast, \system improves substantially with budget. At $b{=}2$, \system (DP) performs best ($\approx 73\%$), while at $b{=}3$, \system (NSGA-II) overtakes ($\approx 77\%$ versus $\approx 75\%$ for DP and $\approx 71\%$ for Hill Climbing).
Unlike deterministic aggregation strategies, \system constructs question-specific sequential/parallel/hybrid execution plans that preserve inter-model reasoning context and allocate budget adaptively across dependent and independent operations. This enables continued gains beyond the baseline plateau. Cost analysis shows that ThriftLLM* and FrugalGPT* consume, on average, 58--76\% across budget levels, while LLM-Ensemble* consumes 53--99\%. In contrast, \system uses 64--88\% of the available budget through selective plan allocation while achieving substantially higher accuracy. Overall, these results demonstrate that deterministic parallel ensembles exhibit early efficiency but limited scalability, whereas adaptive planning in \system breaks this plateau by dynamically composing sequential, parallel, and hybrid plans to achieve superior cost–accuracy efficiency on classification-style QA tasks. 
Compared to LLM-Blender, \system attains substantially higher QoA (0.82 vs.\ 0.14), underscoring the benefit of adaptive execution planning.

\paragraph{Open-Ended QA (SimpleQA)}
We next evaluate \system on the \textit{SimpleQA} benchmark~(Figure~\ref{fig:mmlu_classif}(b)), an open-ended  QA task that tests the model’s ability to synthesize and articulate factual knowledge. Performance is measured by QoA, range [0,1], computed as the cosine similarity between predicted and reference embeddings.
%  \begin{figure}[htb]
%     \centering
%     \includegraphics[width=0.9\linewidth]{figures/sqa.png}
%     \caption{Average Quality per question versus budget on the \textit{SimpleQA} benchmark.}
%     \Description{Accuracy vs cost on the \textit{SimpleQA} benchmark.}
%     \label{fig:simpleqa_classif}
%     \vspace{-0.2 cm}
% \end{figure}
Among the baselines, FrugalGPT* remains nearly flat at around 0.46 QoA across budgets, consuming on average 58--76\% of the budget (mean cost \(7.2{\times}10^{-6}{-}1.4{\times}10^{-5}\)). LLM-Blender* (shown for completeness but excluded from cost-normalized comparisons due to fixed full-ensemble invocation) incurs a nearly fixed cost of \(\approx1.6{\times}10^{-5}\)~USD/query and remains substantially lower at about 0.22 QoA. In contrast, all three \system planner variants outperform these baselines. At the lowest budget ($b{=}1$), \system (DP) performs best ($\approx 0.52$), followed by \system (Hill Climbing) ($\approx 0.46$) and \system (NSGA-II) ($\approx 0.44$). At moderate budgets ($b{=}2{-}3$), the three planners converge to similar QoA ($\approx 0.56$--$0.61$). At higher budgets ($b{=}4{-}5$), \system (NSGA-II) achieves the best QoA, improving to about 0.65 while utilizing 64--88\% of the available budget, whereas DP and Hill Climbing plateau near 0.55--0.58.

\ifextended
\subsection{Impact of Number of Operations}
\label{sect:exp_num_operations}
\ifextended
\else
We study how the maximum number of operations per plan ($k$) affects QoA and resource usage. For each question, we vary $k$ from 1 to 5 and aggregate results across the five \textsc{PerfDB} coverage levels. This experiment serves as a sensitivity analysis on plan-space size: increasing $k$ expands the set of candidate sequential/parallel/hybrid workflows available to the planner.
Due to space constraints, Figure~\ref{fig:k_sweep_mmlu} reports the detailed \textsc{NSGA-II} planner results for \textit{MMLU-Pro}, while the discussion covers the trends observed on both \textit{MMLU-Pro} and \textit{SimpleQA}. Table~\ref{tab:qoa-buckets-k-level} summarizes the corresponding QoA-bucket distributions for both datasets. The extended version~\cite{optiqextended2026} includes the figures for NSGA-II (SimpleQA) and corresponding $k$-sweep analyses for the DP and Hill Climbing baselines, all of which exhibit the same fundamental QoA--resource trade-offs as $k$ scales. The nature of these QoA gains, however, differs across benchmarks because of their evaluation structure. 
\fi
\begin{figure}[htb]
    \centering
    \includegraphics[width=\linewidth]{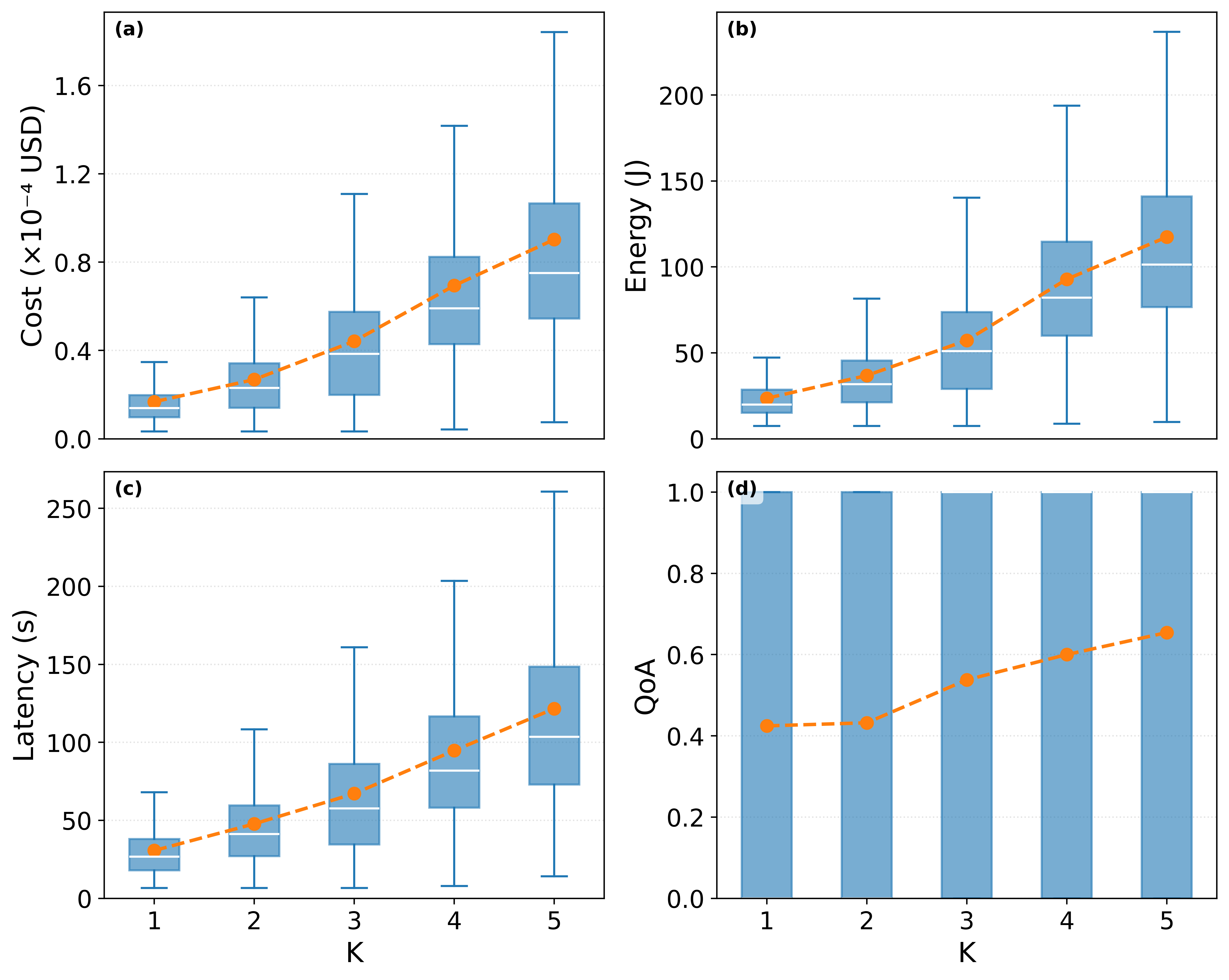} 
    \vspace{-0.5em}
    \caption{NSGA-II performance across operation limits $k$ on \textit{MMLU-Pro}. Panels show per-question distributions of (a) financial cost, (b) energy, (c) latency, and (d) QoA. Boxplots indicate the median and interquartile range; the dashed line with markers shows the mean. Increasing $k$ improves QoA but also increases cost, energy, and latency.}
    \label{fig:k_sweep_mmlu}
\end{figure}
\ifextended
\begin{figure}[htb]
    \centering
    \includegraphics[width=\linewidth]{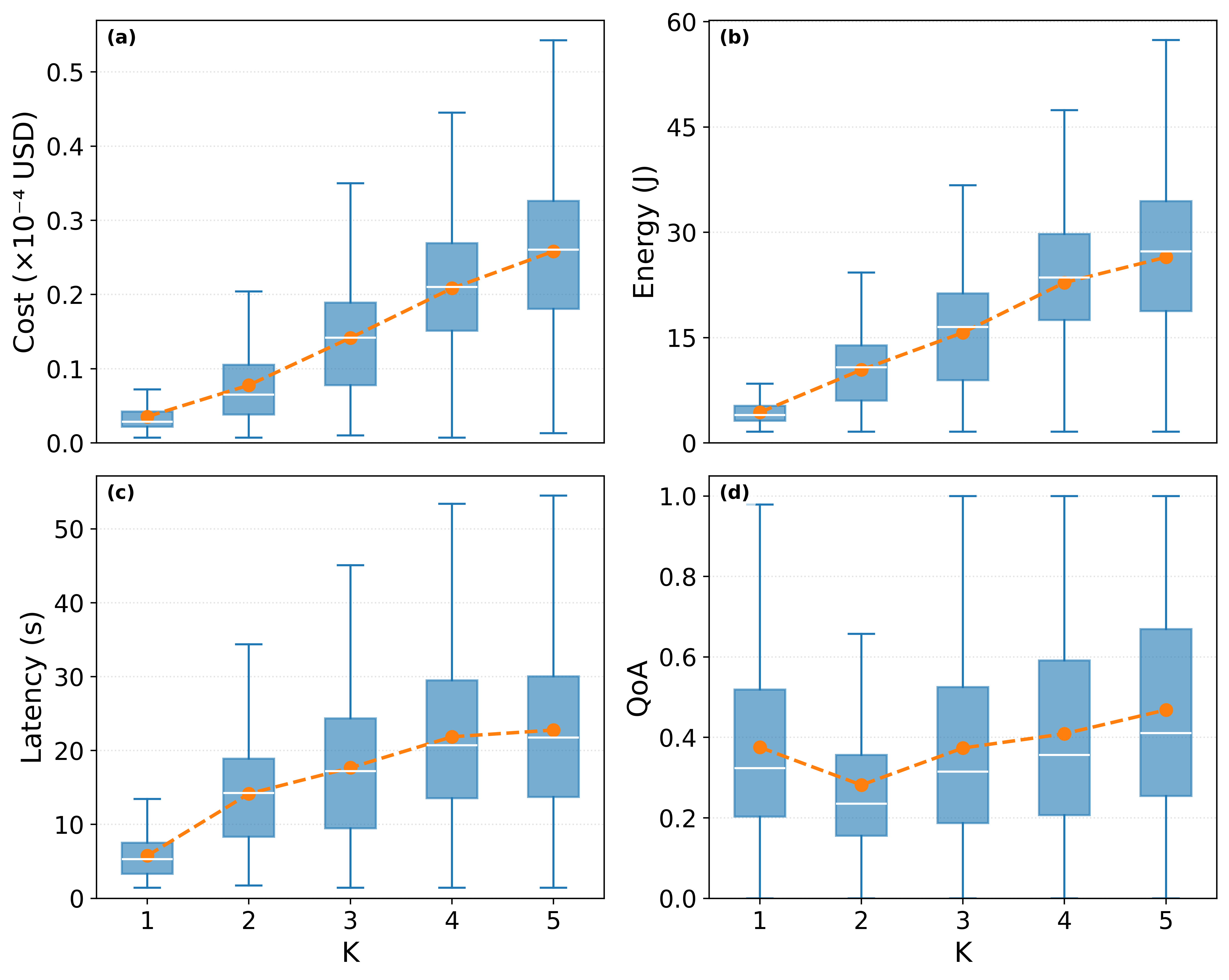} 
    \vspace{-0.5em}
    \caption{NSGA-II performance across operation limits $k$ on \textit{SimpleQA}. Panels show per-question distributions of (a) financial cost, (b) energy, (c) latency, and (d) QoA. Boxplots summarize dispersion (median and interquartile range); the dashed line with markers shows the mean. Increasing $k$ expands the planner's search space and improves QoA, but also raises cost, energy, and latency.}
    \label{fig:k_sweep_sqa}
\end{figure}
\else
\fi
% We study how the maximum number of operations per plan ($k$) affects QoA and resource usage. For each question, we vary $k$ from 1 to 5 and aggregate results across the five \textsc{PerfDB} coverage levels. This provides a sensitivity analysis of plan-space size, since larger $k$ values enable the planner to explore a broader set of sequential, parallel, and hybrid workflows. We report the full $k$-sweep analysis for \textsc{NSGA-II}, DP, and Hill Climbing on both \textit{MMLU-Pro} and \textit{SimpleQA}. Across all three planners, larger $k$ generally yields higher QoA, but with predictable increases in latency, energy, and financial cost. The nature of these QoA gains, however, differs across benchmarks because of their evaluation structure. 

For \textit{MMLU-Pro}, which relies on exact-match multiple-choice accuracy, the QoA distribution is effectively binary (the $0.5$--$0.8$ bucket remains empty). Here, increasing $k$ from 1 to 5 systematically shifts outcomes from the low-QoA bucket ($57.55\% \to 34.58\%$) to the high-QoA bucket ($42.45\% \to 65.42\%$). This demonstrates that larger plans effectively correct previous misclassifications, leading to strong, consistent improvements. In contrast, \textit{SimpleQA} presents a more nuanced, non-monotonic trend due to its open-ended nature. After an initial QoA degradation at $k=2$ (where the low-QoA bucket spikes to $86.36\%$), performance steadily improves. By $k=5$, the low-QoA bucket shrinks to $57.23\%$, while the middle and high-QoA buckets expand to $27.61\%$ and $12.86\%$, respectively. This indicates that while larger plans consistently enhance semantic answer quality, these gains are more gradual and highly sensitive to plan composition compared to rigid exact-match tasks.
\fi
\ifextended
\else
% \begin{table}[t]
% \centering
% \setlength{\tabcolsep}{3pt}
% \renewcommand{\arraystretch}{0.98}
% \footnotesize
% \begin{tabular}{c l r r r}
% \toprule
% \textbf{K} & \textbf{Benchmark} & \textbf{0--0.5} & \textbf{0.5--0.8} & \textbf{0.8--1 } \\
% \midrule
% 1 & MMLU-Pro & 57.55 & 0.00 & 42.45 \\
% 1 & SimpleQA  & 71.63 & 20.00 & 5.12 \\
% \midrule
% 2 & MMLU-Pro & 56.79 & 0.00 & 43.21 \\
% 2 & SimpleQA  & 86.36 & 9.63 & 1.87 \\
% \midrule
% 3 & MMLU-Pro & 46.22 & 0.00 & 53.78 \\
% 3 & SimpleQA  & 71.74 & 20.22 & 6.01 \\
% \midrule
% 4 & MMLU-Pro & 40.00 & 0.00 & 60.00 \\
% 4 & SimpleQA  & 65.22 & 24.51 & 8.48 \\
% \midrule
% 5 & MMLU-Pro & 34.58 & 0.00 & 65.42 \\
% 5 & SimpleQA  & 57.23 & 27.61 & 12.86 \\
% \bottomrule
% \end{tabular}
% \caption{Plan QoA distribution (\%) across buckets for different operation limits $K$ on \textit{MMLU-Pro} and \textit{SimpleQA}. Larger $K$ generally shifts mass toward higher-QoA buckets.}
% \label{tab:qoa_bucket_mmlu_simpleqa}
% \end{table}
% \vspace{-0.05cm}

\fi
Evaluating the raw trade-offs highlights the diverging cost-benefit profiles between the two benchmarks. For SimpleQA, scaling from $k=1$ to $k=5$ exhibits moderate returns: a $23.6\%$ relative QoA gain drives cost, energy, and latency up by roughly $6.5\times$, $6\times$, and $3.8\times$, respectively. Conversely, MMLU-Pro demonstrates a much stronger return on resource investment at higher k. For this dataset, a substantial $54.7\%$ QoA improvement is achieved alongside $4\times$ to $5.3\times$ increases across the resource metrics, justifying the expanded plan space. Crucially, as $k$ grows, the planner increasingly favors concurrent execution. Hybrid and parallel plans account for $\approx36\%$, $77\%$, and $82\%$ of all executions at $k=3, 4$, and $5$, respectively (leaving sequential plans at just $\approx14\%$ by $k=5$). This architectural shift explains the higher resource consumption at larger $k$: $k=3$ captures most benefits for semantic QA, while classification tasks can justify larger $k$ for higher QoA.
\ifextended

We next examine whether this $k$-dependent trend is robust across alternative planner backends. Figures~\ref{fig:k_sweep_dp_mm}, ~\ref{fig:k_sweep_dp_sqa}, ~\ref{fig:k_sweep_hc_mm} and~\ref{fig:k_sweep_hc_sqa} demonstrate that both DP and Hill Climbing exhibit the same qualitative pattern as NSGA-II: expanding the operations limit ($k$) improves QoA but predictably incurs higher financial cost, energy consumption, and latency. The primary distinction lies in the absolute QoA ceilings achieved at higher k. Hill Climbing remains highly competitive, reaching a mean QoA of $0.56$ on MMLU-Pro and $\approx0.47$ on SimpleQA at $k=5$. In contrast, DP plateaus earlier and yields lower top-end quality, peaking at $\approx0.48$ on MMLU-Pro and $\approx0.44$ on SimpleQA. The QoA-band distributions corroborate this performance gap. For example, DP on MMLU-Pro sees its high-QoA band ($0.8$--$1.0$) grow from $39.38\%$ at $k=1$ to $48.35\%$ at $k=5$, while its low-QoA band ($0$--$0.5$) drops from $60.63\%$ to $51.65\%$. On SimpleQA, DP's high-QoA band only reaches $10.29\%$ by $k=5$. These results confirm that the fundamental trade-off—improving QoA through expanded plan spaces at the cost of diminishing efficiency returns beyond moderate $k$—is universal across algorithms. However, they also clearly highlight that the capacity to efficiently translate a larger search space into high-quality outcomes is highly dependent on the planner's underlying search strategy.
\fi

\ifextended
\begin{figure}[htb]
    \centering
    \includegraphics[width=\linewidth]{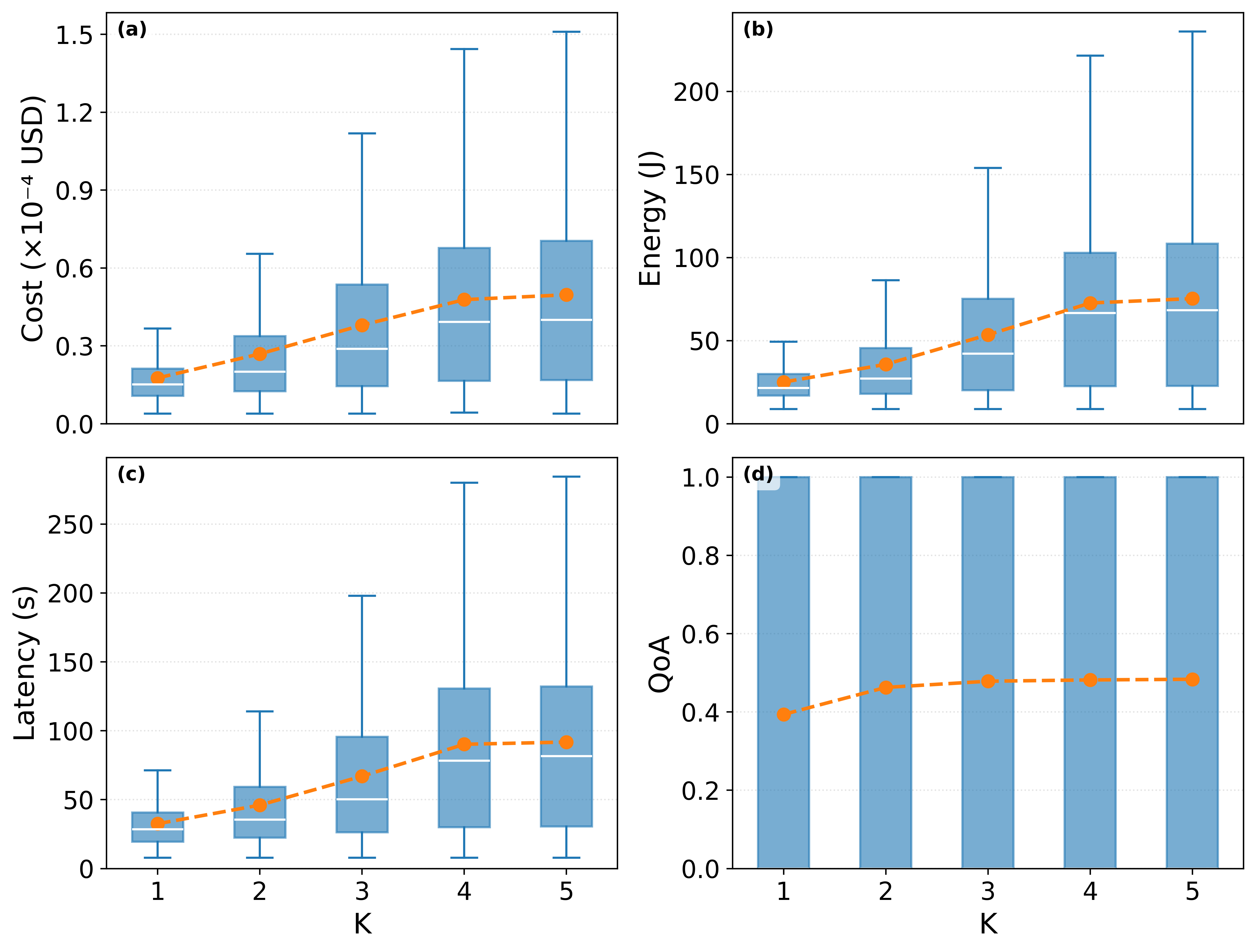} % DP+$\Delta$ k-sweep
    \vspace{-0.5em}
    \caption{DP performance across operation limits $k$ on \textit{MMLU-Pro}. Panels show per-question distributions of (a) financial cost, (b) energy, (c) latency, and (d) QoA. Boxplots summarize dispersion (median and interquartile range); the dashed line with markers shows the mean. Increasing $k$ expands the planner's search space and improves QoA.}
    \label{fig:k_sweep_dp_mm}
\end{figure}

\begin{figure}[htb]
    \centering
    \includegraphics[width=\linewidth]{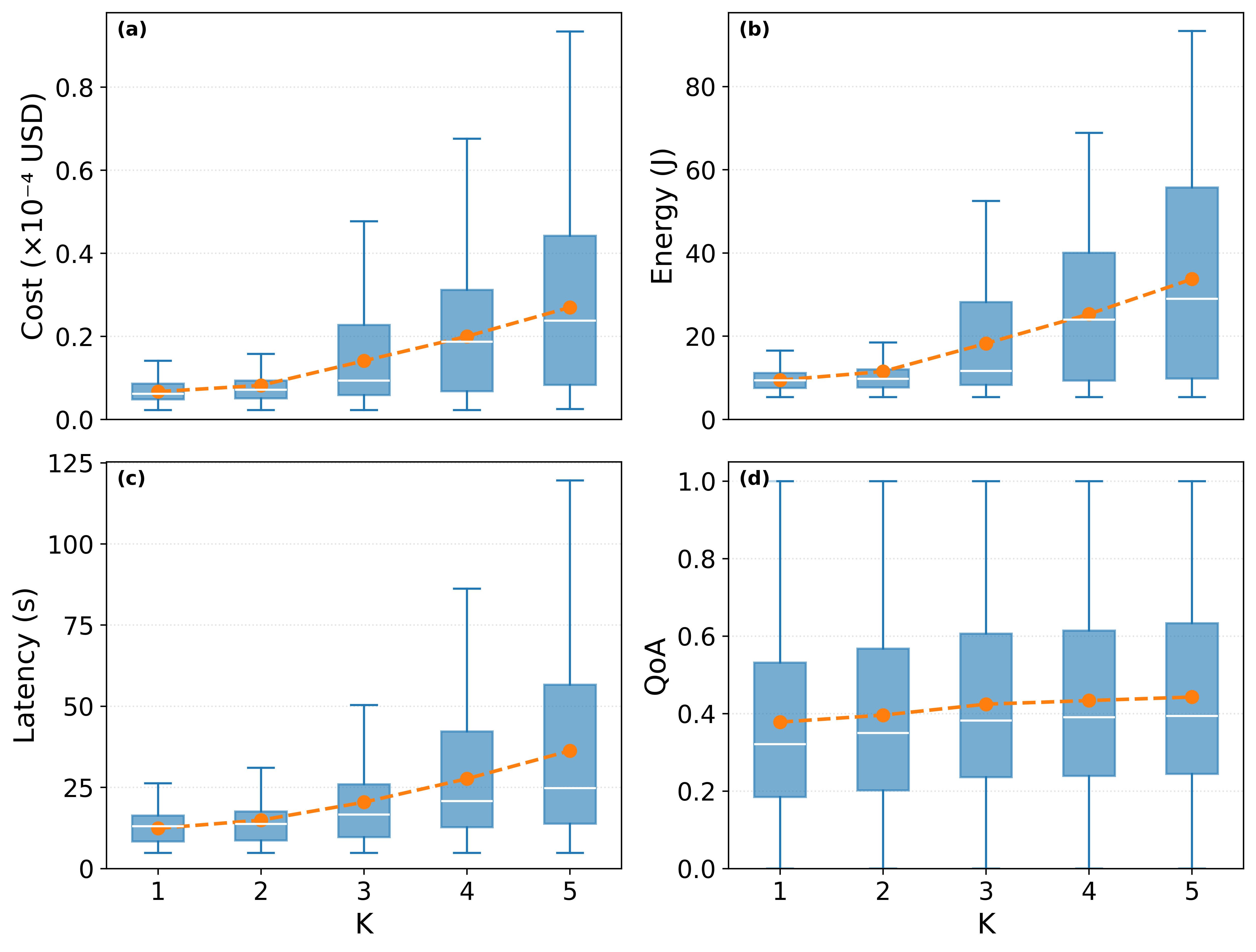} % Hill Climbing k-sweep
    \vspace{-0.5em}
    \caption{DP performance across operation limits $k$ on \textit{SimpleQA}. Panels show per-question distributions of (a) financial cost, (b) energy, (c) latency, and (d) QoA. Boxplots summarize dispersion (median and interquartile range); the dashed line with markers shows the mean. Increasing $k$ expands the planner's search space and improves QoA.}
    \label{fig:k_sweep_dp_sqa}
\end{figure}

\begin{figure}[htb]
    \centering
    \includegraphics[width=\linewidth]{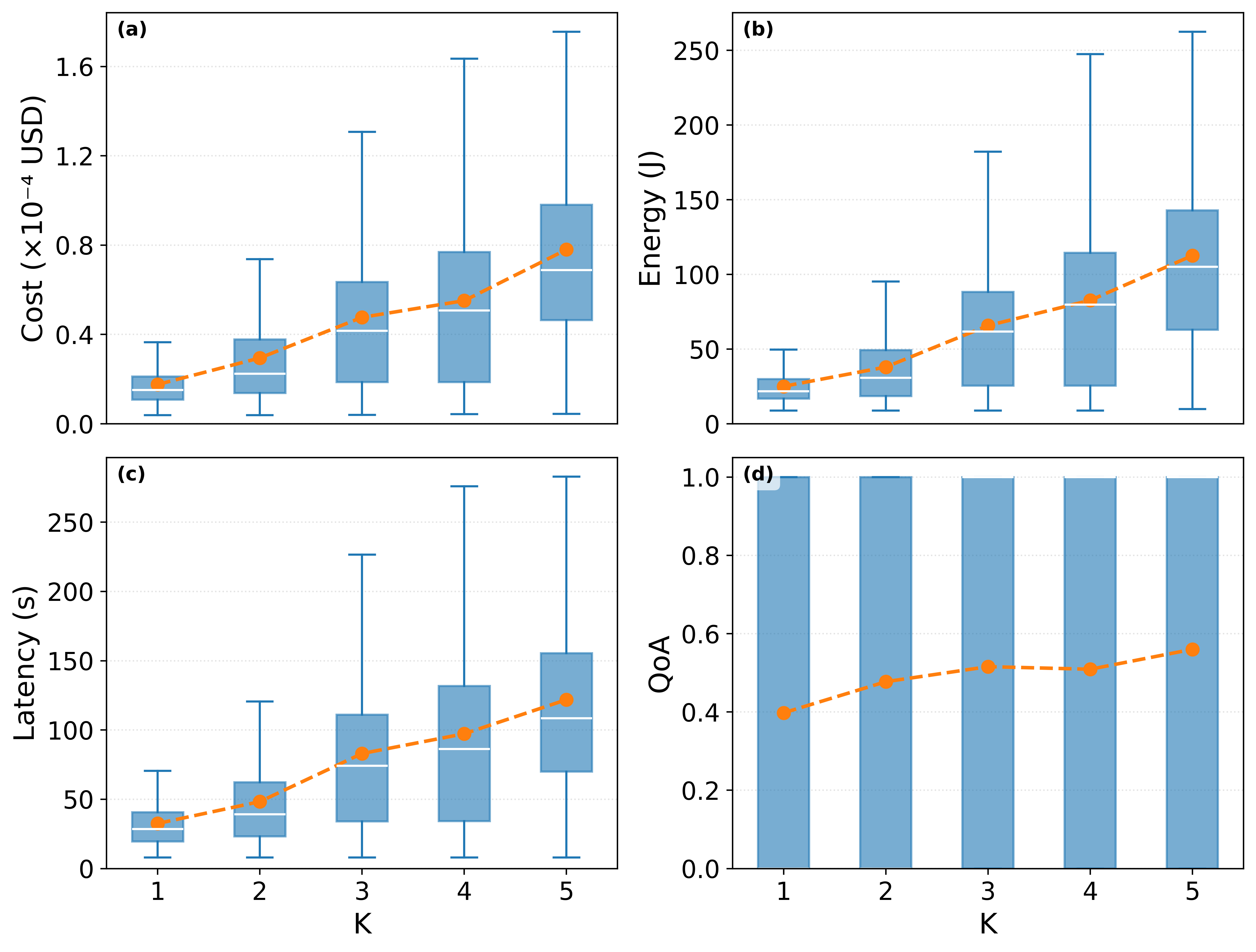} % DP+$\Delta$ k-sweep
    \vspace{-0.5em}
    \caption{Hill Climbing performance across operation limits $k$ on \textit{MMLU-Pro}. Panels show per-question distributions of (a) financial cost, (b) energy, (c) latency, and (d) QoA. Boxplots summarize dispersion (median and interquartile range); the dashed line with markers shows the mean. Increasing $k$ expands the planner's search space and improves QoA.}
    \label{fig:k_sweep_hc_mm}
\end{figure}

\begin{figure}[htb]
    \centering
    \includegraphics[width=\linewidth]{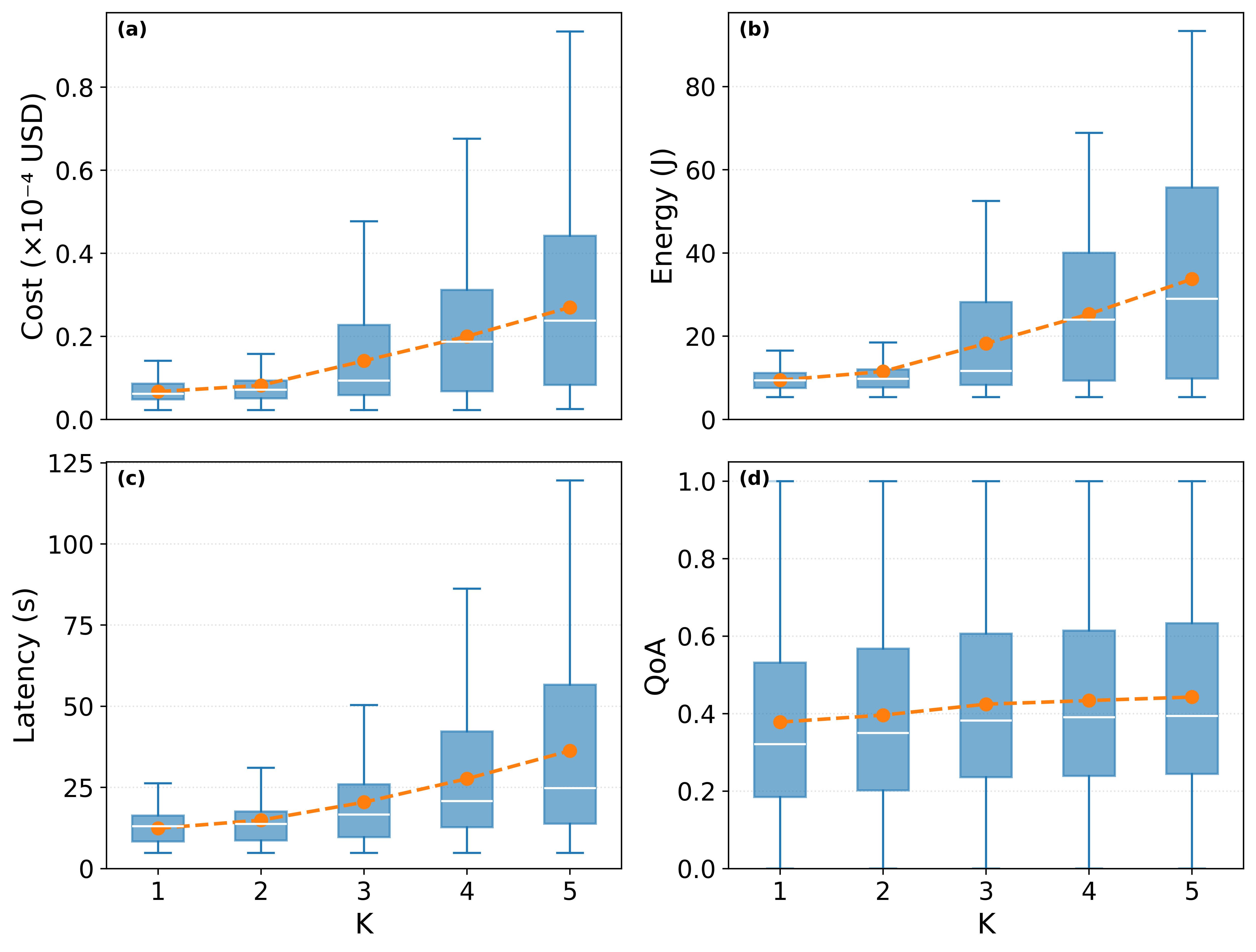} % Hill Climbing k-sweep
    \vspace{-0.5em}
    \caption{Hill Climbing performance across operation limits $k$ on \textit{SimpleQA}. Panels show per-question distributions of (a) financial cost, (b) energy, (c) latency, and (d) QoA. Boxplots summarize dispersion (median and interquartile range); the dashed line with markers shows the mean. Increasing $k$ expands the planner's search space and improves QoA.}
    \label{fig:k_sweep_hc_sqa}
\end{figure}
\fi
% \vspace{-0.8em}
\subsection{Robustness to Limited Historical Data in \textsc{PerfDB}}
\label{sect:exp_hist_complexity}
We assess how \system behaves when \textsc{PerfDB} has limited historical coverage. This experiment directly studies cold-start and sparse-statistics settings by varying the amount of empirical execution metadata available to the planner. We define five coverage levels ($0$--$4$). Level~0 contains only static model priors, representing a cold-start setting with no empirical execution history. Levels~1--4 progressively enrich \textsc{PerfDB} with empirical traces collected from \emph{disjoint training questions} that share the same topical distribution as the evaluation set. This setup enables the planner to generalize from related but unseen instances while avoiding data leakage. The cumulative numbers of recorded executions available in \textsc{PerfDB} at Levels~1--4 are $135$, $985$, $1{,}735$, and $2{,}360$, respectively. These counts denote execution traces used to populate \textsc{PerfDB}, not evaluation questions. For each question, the planner can compose up to five operations ($k{=}5$), as identified in the previous experiment.

\ifextended
We report the full level-sweep analysis across planners, including \textsc{NSGA-II}, DP, and Hill Climbing, for both \textit{MMLU-Pro} and \textit{SimpleQA}. Across all planners, richer \textsc{PerfDB} coverage consistently improves planning quality and downstream QoA, with the largest gains occurring when moving from sparse to moderate coverage. For \textsc{NSGA-II}, Figures~\ref{fig:fig_lvl_nsga_mm} and~\ref{fig:fig_lvl_nsga_sqa} compare predicted versus actual plan metrics across coverage levels for \textit{MMLU-Pro} and \textit{SimpleQA}, respectively, showing improved prediction fidelity and a shift toward higher-QoA regions as coverage increases. Table~\ref{tab:qoa-buckets-k-level} further summarizes how the plan-QoA distribution evolves across coverage levels for each benchmark.

\begin{figure}[htb]
    \centering
    \begin{subfigure}[t]{0.98\linewidth}
        \centering
        \includegraphics[width=\linewidth]{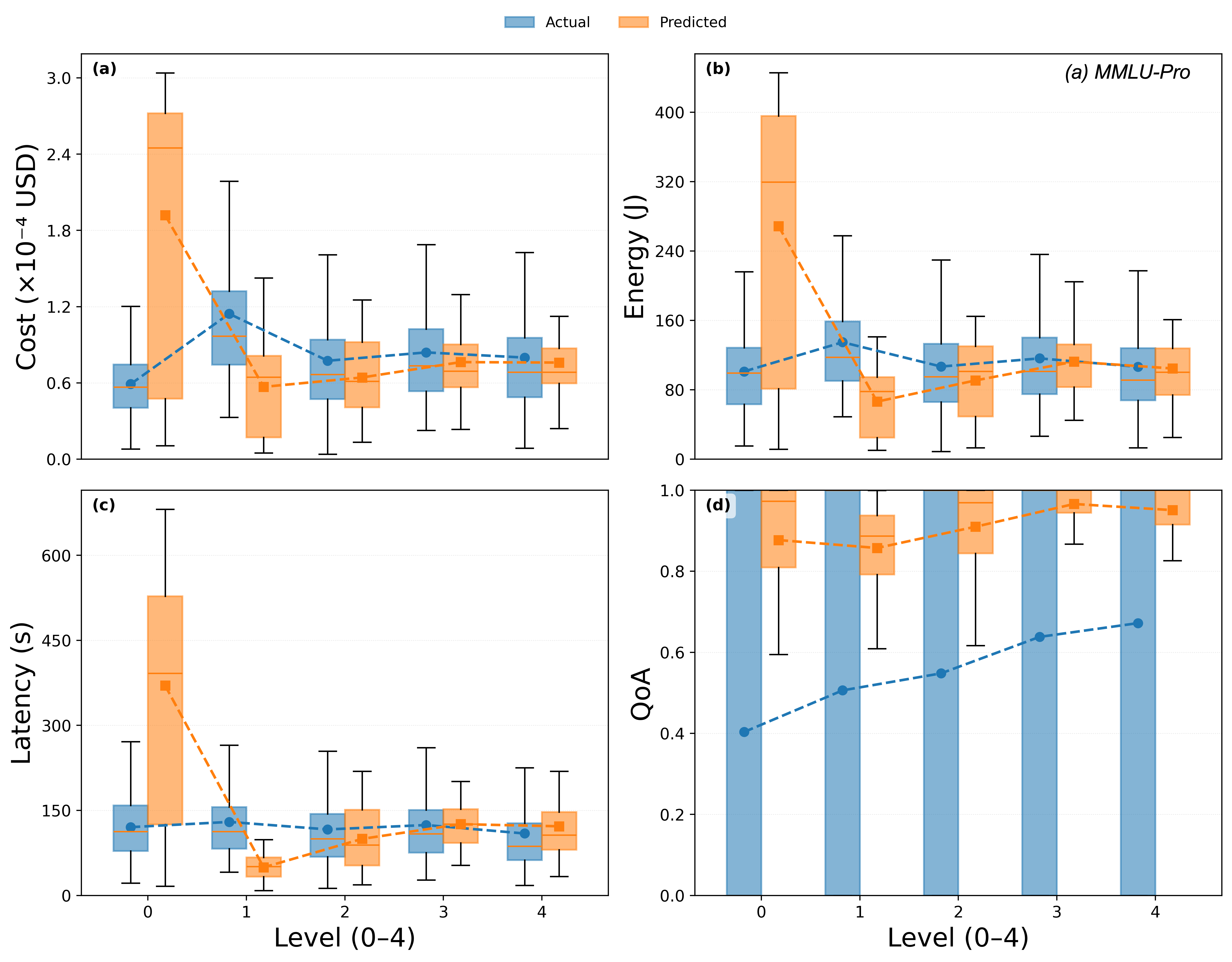}
        \caption{\textit{MMLU-Pro}}
        \label{fig:fig_lvl_nsga_mm}
    \end{subfigure}
    
    \vspace{0.3em}
    
    \begin{subfigure}[t]{0.98\linewidth}
        \centering
        \includegraphics[width=\linewidth]{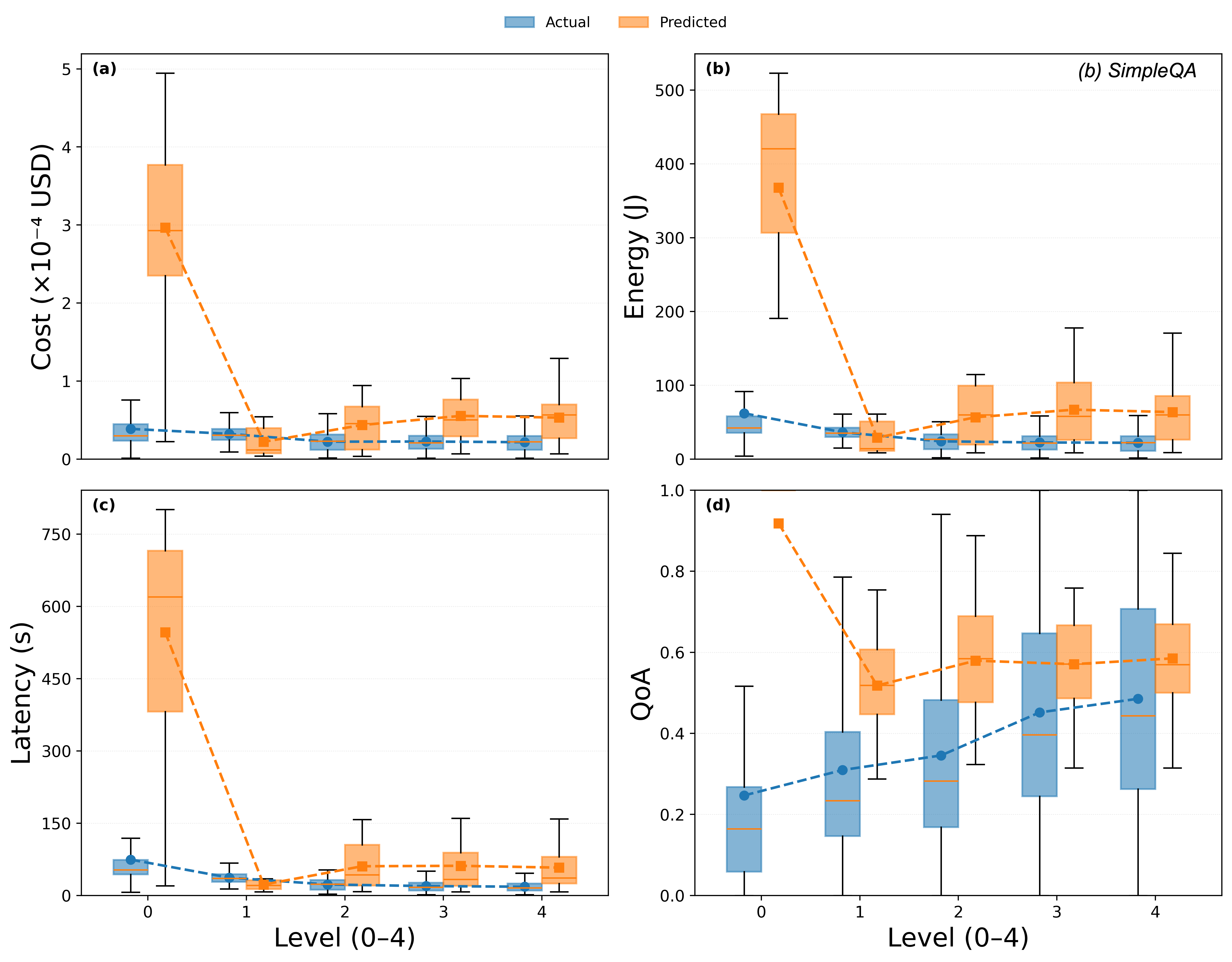}
        \caption{\textit{SimpleQA}}
        \label{fig:fig_lvl_nsga_sqa}
    \end{subfigure}
    
    \caption{Actual vs. predicted plan-level cost, energy, latency, and QoA across \textsc{PerfDB} coverage levels (0–4) for NSGA-II on MMLU-Pro (a) and SimpleQA (b). In both, calibration improves with historical coverage: early overestimation under sparse data (Level 0) diminishes at higher levels, while predictions preserve the level-wise trends needed for planner ranking decisions.}
    \label{fig:fig_lvl_nsga_combined}
    \vspace{-0.25cm}
\end{figure}

\else

Due to space constraints, the main paper reports NSGA-II results for \textit{MMLU-Pro} and \textit{SimpleQA} (Figures~\ref{fig:fig_lvl_nsga_mm} and~\ref{fig:fig_lvl_nsga_sqa}); the extended version~\cite{optiqextended2026} includes the corresponding level-sweep analysis for DP and Hill Climbing, which show the same qualitative trend. Figures~\ref{fig:fig_lvl_nsga_mm} and~\ref{fig:fig_lvl_nsga_sqa} compare predicted and actual plan-level cost, energy, latency, and QoA across \textsc{PerfDB} coverage levels, while Table~\ref{tab:qoa-buckets-k-level} summarizes the resulting QoA-bucket shifts.

% \begin{figure}[htb]
%     \centering
%     \begin{subfigure}[t]{0.98\linewidth}
%         \centering
%         \includegraphics[width=\linewidth]{figures/levels_fig/nsga_Levels_comp_mmlu.png}
%         % \caption{\textit{MMLU-Pro}}
%         \label{fig:fig_lvl_nsga_mm}
%     \end{subfigure}
    
%     \vspace{-1.2em} 
    
%     \begin{subfigure}[t]{0.98\linewidth}
%         \centering
%         \includegraphics[width=\linewidth]{figures/levels_fig/nsga_Levels_comp_simpleqa.png}
%         % \caption{\textit{SimpleQA}}
%         \label{fig:fig_lvl_nsga_sqa}
%     \end{subfigure}
\begin{figure}[t]
    \centering
    \begin{subfigure}[t]{0.98\linewidth}
        \centering
        \includegraphics[width=\linewidth]{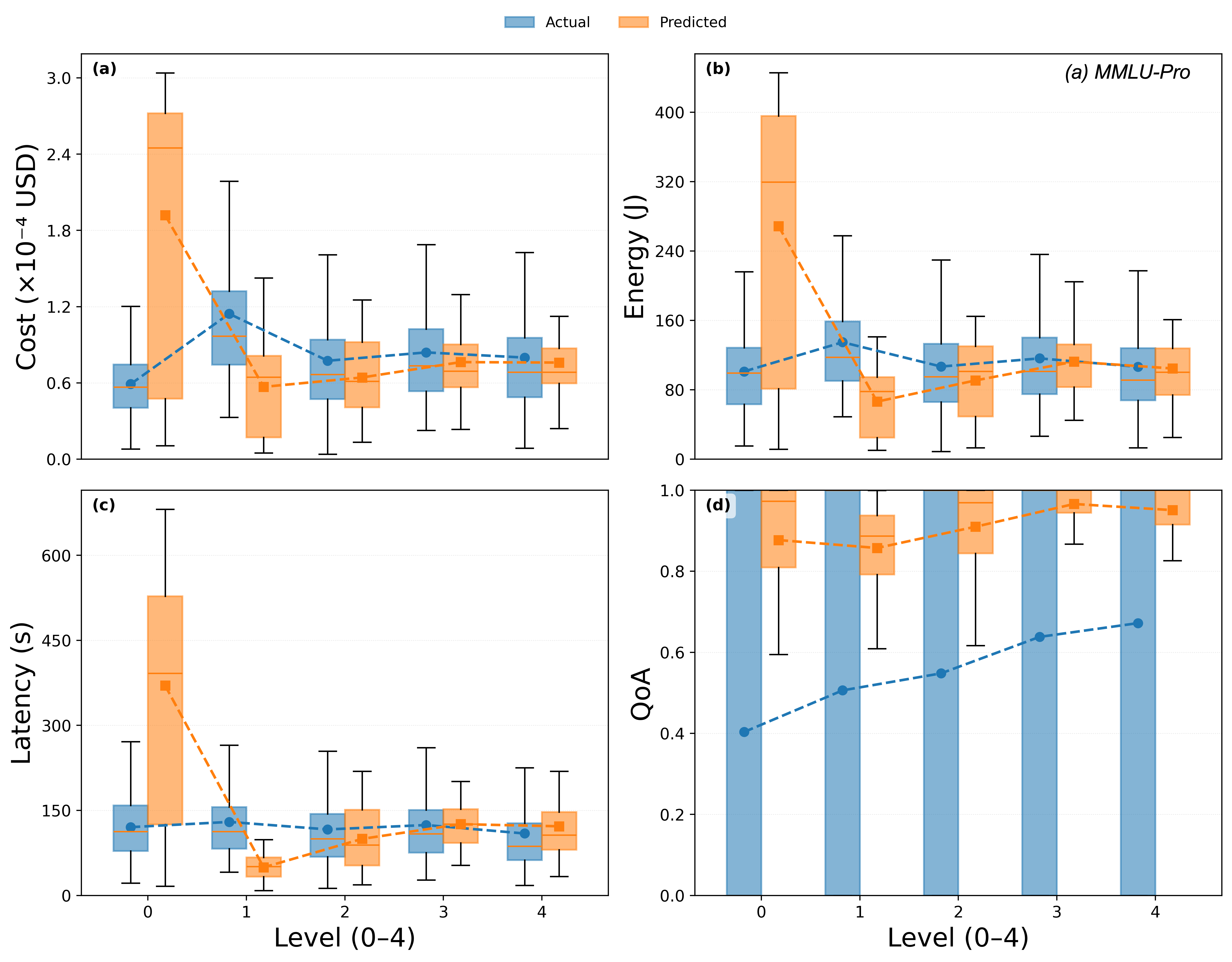}
        \phantomcaption
        \label{fig:fig_lvl_nsga_mm}
    \end{subfigure}
    
    \vspace{-1.5em}
    
    \begin{subfigure}[t]{0.98\linewidth}
        \centering
        \includegraphics[width=\linewidth]{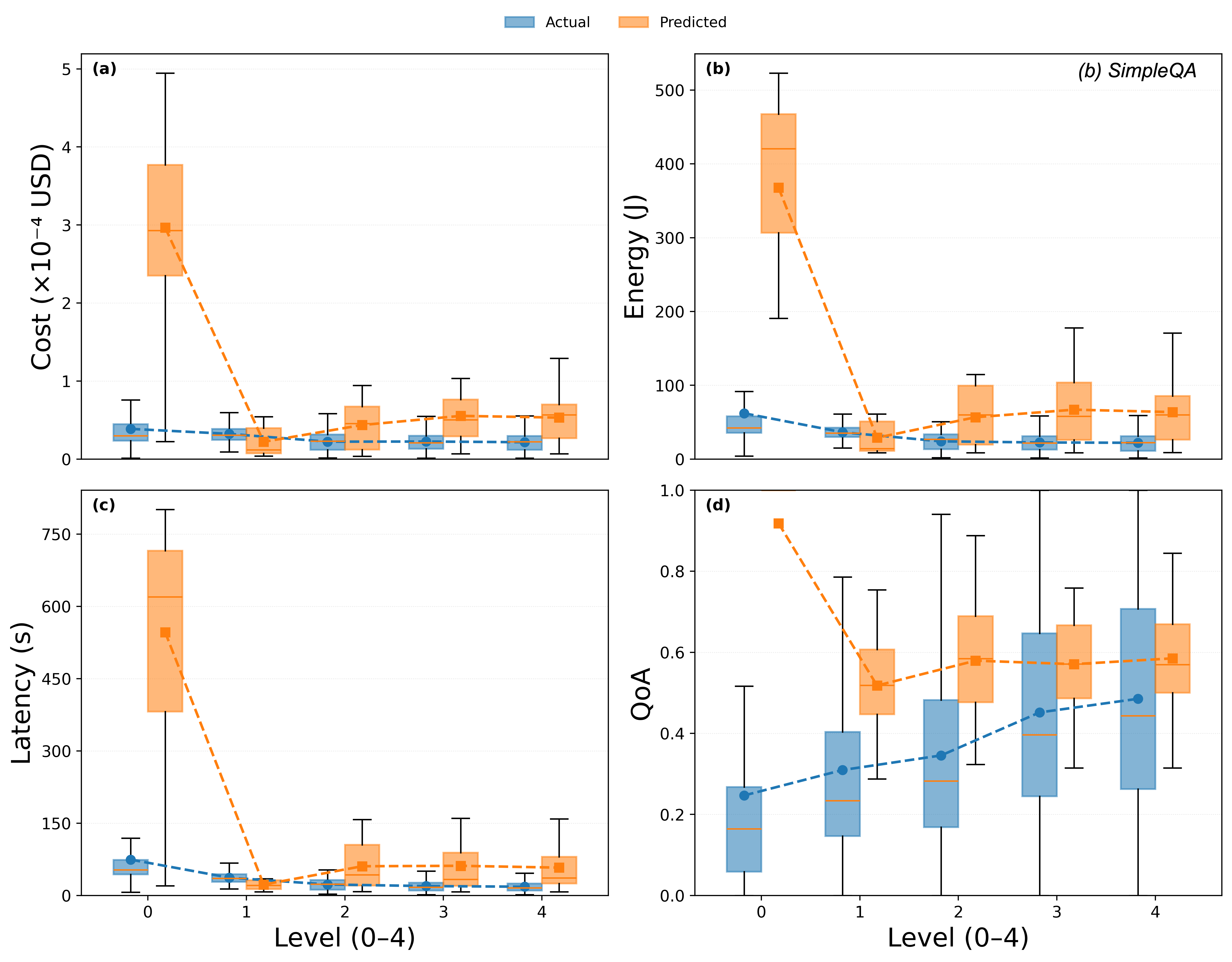}
        \phantomcaption
        \label{fig:fig_lvl_nsga_sqa}
    \end{subfigure}
    \vspace{-0.5cm}
    \caption{Actual vs. predicted plan-level cost, energy, latency, and QoA across \textsc{PerfDB} coverage levels (0--4) for \textsc{NSGA-II} on \textit{MMLU-Pro} (a) and \textit{SimpleQA} (b). As historical coverage increases, resource estimates become better calibrated and QoA estimates provide a stronger selection signal for planning.}
    \label{fig:fig_lvl_nsga_combined}
    \vspace{-0.25cm}
\end{figure}
% \begin{figure}[htb]
%     \centering
%     \includegraphics[width=0.95\linewidth]{figures/levels_fig/nsga_Levels_comp_mmlu.png}
%     \caption{Actual vs.\ predicted plan-level cost, energy, latency, and QoA across \textsc{PerfDB} coverage levels (0--4) for NSGA-II on \textit{MMLU-Pro}. Calibration improves with historical coverage: early overestimation under sparse data (Level~0) diminishes at higher levels, while predictions preserve the level-wise trends needed for planner ranking decisions.}
%     \label{fig:fig_lvl_nsga_mm}
%     \vspace{-0.25cm}
% \end{figure}

% \begin{figure}[htb]
%     \centering
%     \includegraphics[width=0.95\linewidth]{figures/levels_fig/nsga_Levels_comp_simpleqa.png}
%     \caption{Actual vs.\ predicted plan-level cost, energy, latency, and QoA across \textsc{PerfDB} coverage levels (0--4) for NSGA-II on \textit{SimpleQA}. Calibration improves with historical coverage: early overestimation under sparse data (Level~0) diminishes at higher levels, while predictions preserve the level-wise trends needed for planner ranking decisions.}
%     \label{fig:fig_lvl_nsga_sqa}
%     \vspace{-0.25cm}
% \end{figure}

\fi
Richer \textsc{PerfDB} coverage consistently improves plan quality. For \textit{MMLU-Pro}, QoA increases monotonically from $0.40$ at Level~0 to $0.67$ at Level~4 ($+0.27$, $\approx66.7\%$ relative improvement), while resource usage stabilizes after an early Level~1 calibration increase. Mean cost rises from $5.9{\times}10^{-5}$ at Level~0 to $1.14{\times}10^{-4}$ at Level~1, then settles in the $7.7{\times}10^{-5}$--$8.4{\times}10^{-5}$ range at Levels~2--4; energy and latency show the same pattern, peaking at Level~1 ($134.8$\,J, $129.7$\,s) and then remaining broadly stable (roughly $106$--$116$\,J and $109$--$124$\,s). For \textit{SimpleQA}, QoA increases from $0.25$ to $0.48$ ($+0.24$, $\approx96.7\%$ relative improvement), while resource usage decreases substantially: cost drops from $3.9{\times}10^{-5}$ to $2.2{\times}10^{-5}$, energy from $62.1$\,J to $22.1$\,J, and latency from $74.4$\,s to $18.1$\,s (Levels~0 to 4). Thus, empirical history enables the planner to select plans that are not only higher quality but also more efficient, especially for open-ended QA where cold-start priors are conservative.
% For \textit{MMLU-Pro}, QoA improves monotonically from $0.40$ at Level~0 to $0.67$ at Level~4 (an absolute gain of $+0.27$, i.e., $\approx66.7\%$ relative improvement). Resource usage is comparatively stable after an early increase at Level~1: mean cost rises from $5.9{\times}10^{-5}$ (Level~0) to $1.14{\times}10^{-4}$ (Level~1), then settles in the $7.7{\times}10^{-5}$--$8.4{\times}10^{-5}$ range (Levels~2--4). Energy and latency show the same pattern, peaking at Level~1 ($134.8$\,J, $129.7$\,s) and then remaining broadly stable (roughly $106$--$116$\,J and $109$--$124$\,s). This indicates that richer history primarily improves plan selection quality on MMLU-Pro without requiring sustained additional resource expenditure beyond early calibration. For \textit{SimpleQA}, QoA also increases monotonically, from $0.25$ at Level~0 to $0.48$ at Level~4 (an absolute gain of $+0.24$, i.e., $\approx96.7\%$ relative improvement). In contrast to MMLU-Pro, resource usage \emph{decreases} substantially as coverage improves: mean cost drops from $3.9{\times}10^{-5}$ to $2.2{\times}10^{-5}$, energy from $62.1$\,J to $22.1$\,J, and latency from $74.4$\,s to $18.1$\,s (Levels~0 to 4). This suggests that empirical history enables the planner to select not only better but also more efficient plans for open-ended QA, where the initial cold-start priors appear especially conservative. 

Prediction error decreases sharply as \textsc{PerfDB} coverage increases. For \textit{MMLU-Pro}, MAE drops from Level~0 to Level~4 for cost ($1.37{\to}0.33{\times}10^{-4}$ USD), energy ($173.8{\to}41.5$\,J), latency ($257.0{\to}48.8$\,s), and QoA ($0.58{\to}0.34$). The trend is even stronger for \textit{SimpleQA}: cost MAE decreases from $2.58{\to}0.32{\times}10^{-4}$ USD, energy from $307.3{\to}41.6$\,J, latency from $474.0{\to}39.6$\,s, and QoA from $0.72{\to}0.27$. Thus, empirical traces rapidly correct cold-start bias in resource estimates, while QoA remains harder to calibrate but still improves with coverage. 
\begin{table}[!ht]
\ifextended
\caption{Plan QoA distribution (\%) of solutions generated by \textsc{NSGA-II} across QoA buckets as a function of operation limit $K$ and \textsc{PerfDB} coverage level on MMLU-Pro and SimpleQA. Larger $K$ and higher coverage levels generally shift the distribution toward higher-QoA buckets.}
\else
\caption{Plan QoA distribution (\%) across QoA buckets as a function of operation limit $K$ and \textsc{PerfDB} coverage level on MMLU-Pro and SimpleQA. Larger $K$ and higher levels generally shift mass toward higher-QoA buckets.}
\fi
\label{tab:qoa-buckets-k-level}
\centering
\scriptsize
\setlength{\tabcolsep}{3.5pt}
\renewcommand{\arraystretch}{1.05}
\begin{tabular}{llccc|ccc}
\toprule
\multirow{2}{*}{Param} & \multirow{2}{*}{Val} &
\multicolumn{3}{c|}{MMLU-Pro QoA bucket (\%)} &
\multicolumn{3}{c}{SimpleQA QoA bucket (\%)} \\
& & 0--0.5 & 0.5--0.8 & 0.8--1.0 & 0--0.5 & 0.5--0.8 & 0.8--1.0 \\
\midrule
\multicolumn{8}{l}{\textbf{Operation limit} $K$} \\
$K$ & 1 & 57.55 & 0.00 & 42.45 & 71.63 & 20.00 & 5.12 \\
$K$ & 2 & 56.79 & 0.00 & 43.21 & 86.36 & 9.63  & 1.87 \\
$K$ & 3 & 46.22 & 0.00 & 53.78 & 71.74 & 20.22 & 6.01 \\
$K$ & 4 & 40.00 & 0.00 & 60.00 & 65.22 & 24.51 & 8.48 \\
$K$ & 5 & 34.58 & 0.00 & 65.42 & 57.23 & 27.61 & 12.86 \\
\midrule
\multicolumn{8}{l}{\textbf{\textsc{PerfDB} coverage level}} \\
Level & 0 & 59.67 & 0.00 & 40.33 & 87.86 & 0.43  & 11.71 \\
Level & 1 & 49.40 & 0.00 & 50.60 & 80.40 & 13.60 & 6.00 \\
Level & 2 & 45.20 & 0.00 & 54.80 & 76.20 & 17.40 & 6.40 \\
Level & 3 & 36.20 & 0.00 & 63.80 & 59.00 & 27.40 & 13.60 \\
Level & 4 & 32.80 & 0.00 & 67.20 & 56.00 & 28.00 & 16.00 \\
\bottomrule
\end{tabular}
\end{table}
Table~\ref{tab:qoa-buckets-k-level} shows that improved calibration translates into better plan selection. For \textit{MMLU-Pro}, the high-QoA bucket ($0.8$--$1.0$) grows from $40.33\%$ at Level~0 to $67.20\%$ at Level~4, while the low-QoA bucket drops from $59.67\%$ to $32.80\%$. The middle bucket remains empty because exact-match multiple-choice scoring produces effectively binary QoA outcomes. For \textit{SimpleQA}, the shift is more gradual: the low-QoA bucket decreases from $87.86\%$ to $56.00\%$, the middle bucket expands from $0.43\%$ to $28.00\%$, and the high-QoA bucket increases from $11.71\%$ to $16.00\%$. This reflects the open-ended nature of \textit{SimpleQA}, where richer history often moves plans from poor to moderate semantic quality before reaching the highest QoA band. Overall, these results show that \textsc{PerfDB} provides an actionable selection signal for multi-objective planning even when absolute QoA magnitudes remain imperfect.

\ifextended

\begin{figure}[htb]
    \centering
    \begin{subfigure}[t]{0.98\linewidth}
        \centering
        \includegraphics[width=\linewidth]{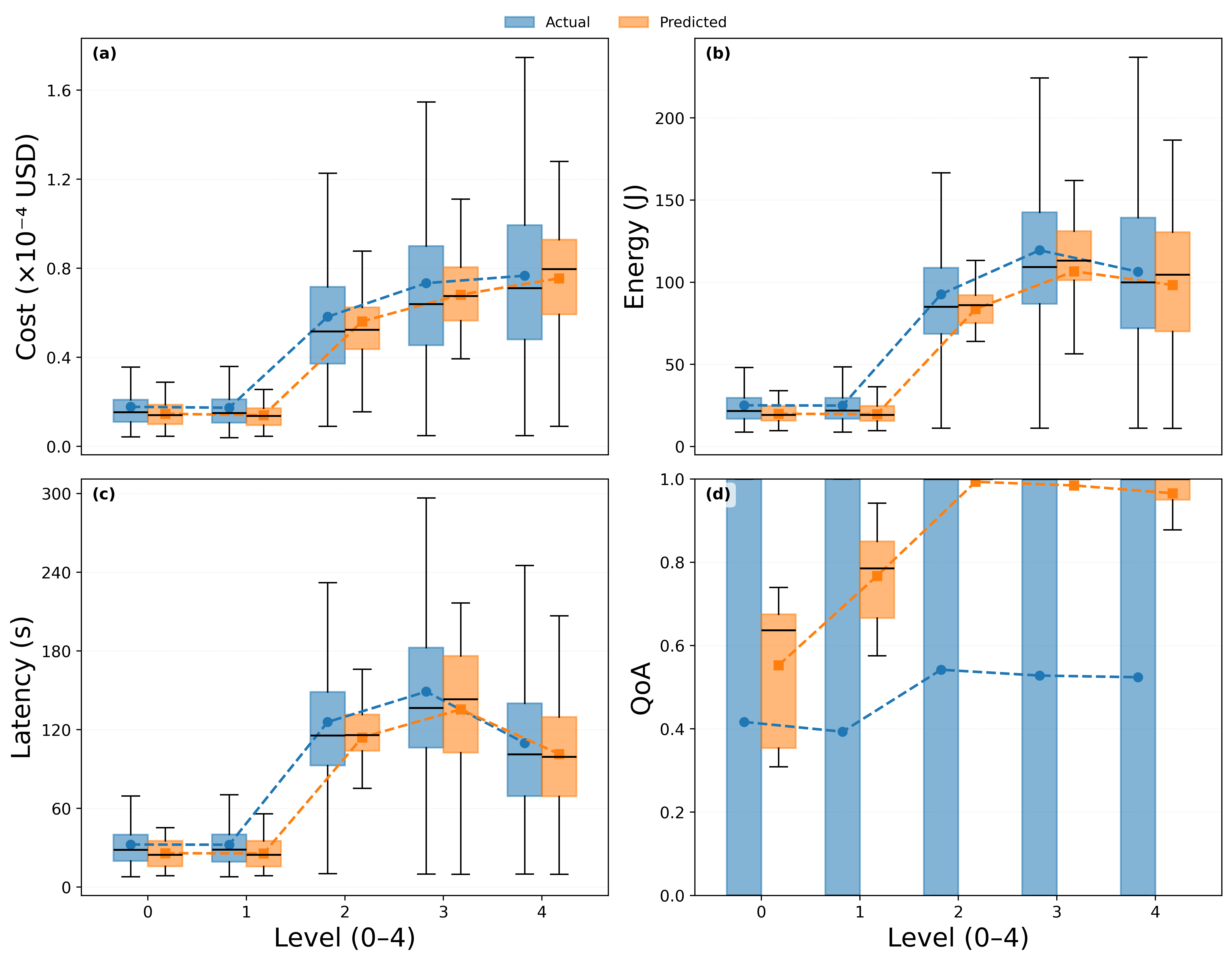}
        \caption{\textit{MMLU-Pro}}
        \label{fig:fig_lvl_dp_mm}
    \end{subfigure}
    
    \vspace{0.3em}
    
    \begin{subfigure}[t]{0.98\linewidth}
        \centering
        \includegraphics[width=\linewidth]{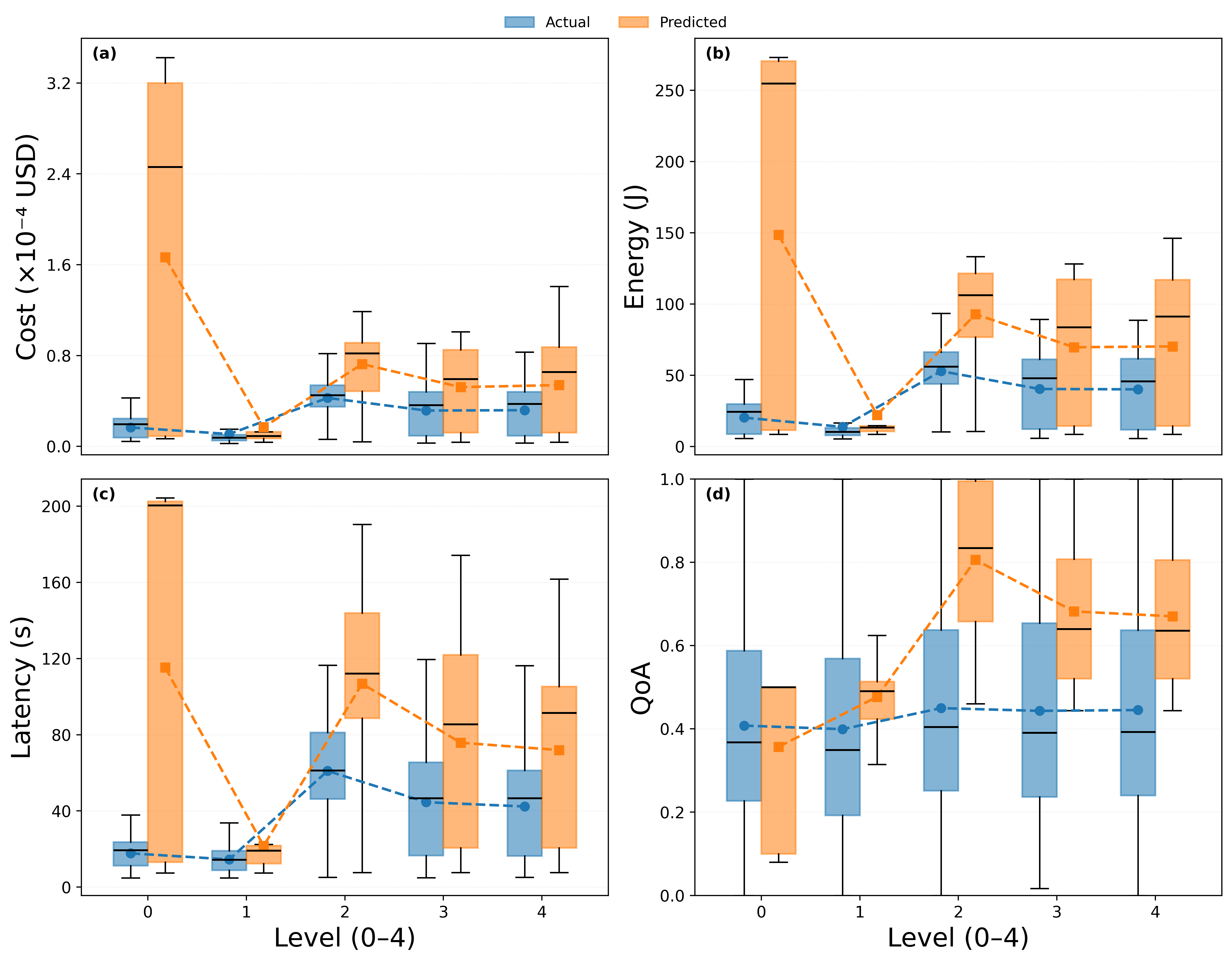}
        \caption{\textit{SimpleQA}}
        \label{fig:fig_lvl_dp_sqa}
    \end{subfigure}
    
    \caption{Actual vs. predicted plan-level cost, energy, latency, and QoA across \textsc{PerfDB} coverage levels (0–4) for DP on MMLU-Pro (a) and SimpleQA (b). In both, calibration improves with historical coverage: early overestimation under sparse data (Level 0) diminishes at higher levels, while predictions preserve the level-wise trends needed for planner ranking decisions.}
    \label{fig:fig_lvl_dp_combined}
    \vspace{-0.25cm}
\end{figure}

\begin{figure}[htb]
    \centering
    \begin{subfigure}[t]{0.98\linewidth}
        \centering
        \includegraphics[width=\linewidth]{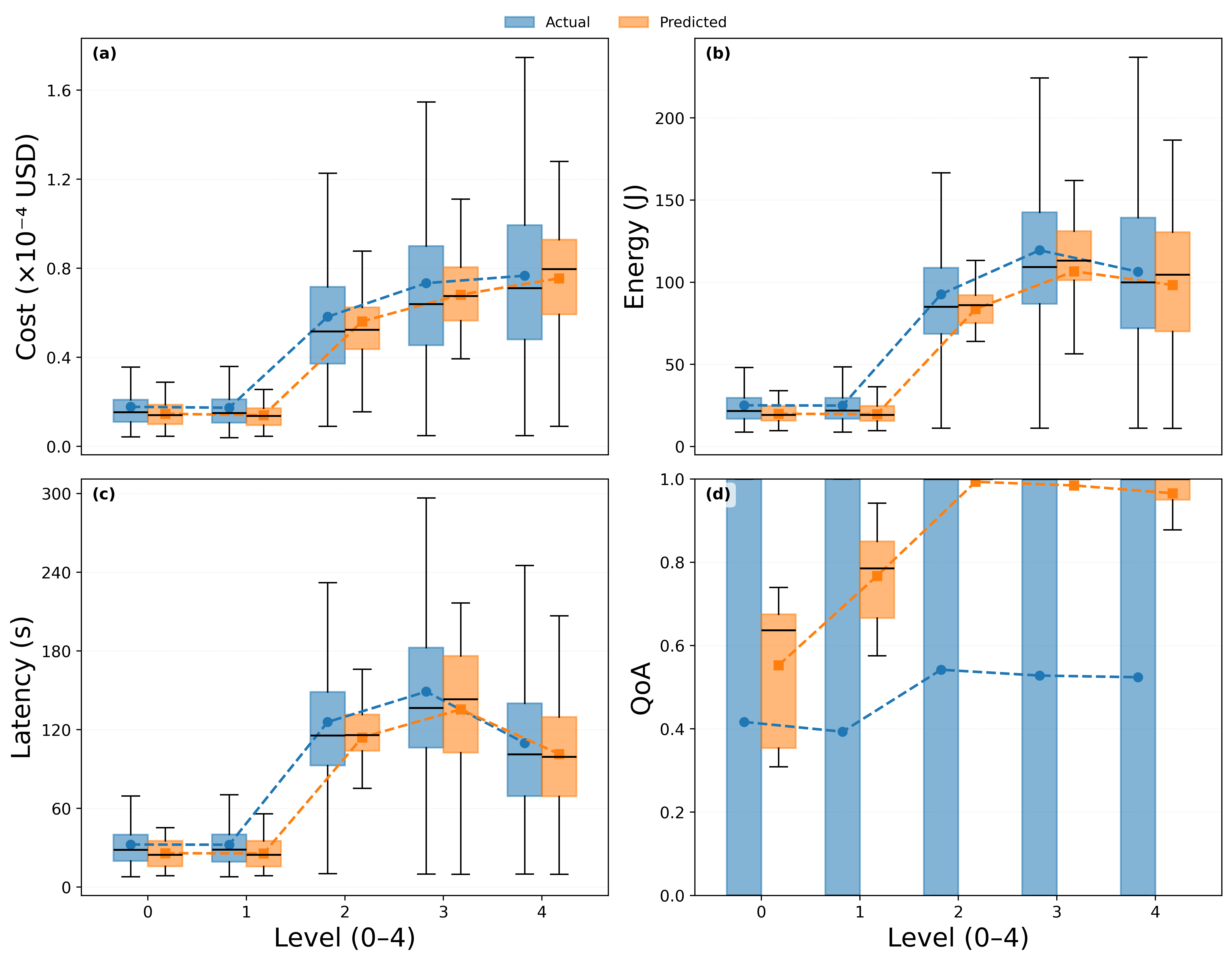}
        \caption{\textit{MMLU-Pro}}
        \label{fig:fig_lvl_hc_mm}
    \end{subfigure}
    
    \vspace{0.3em}
    
    \begin{subfigure}[t]{0.98\linewidth}
        \centering
        \includegraphics[width=\linewidth]{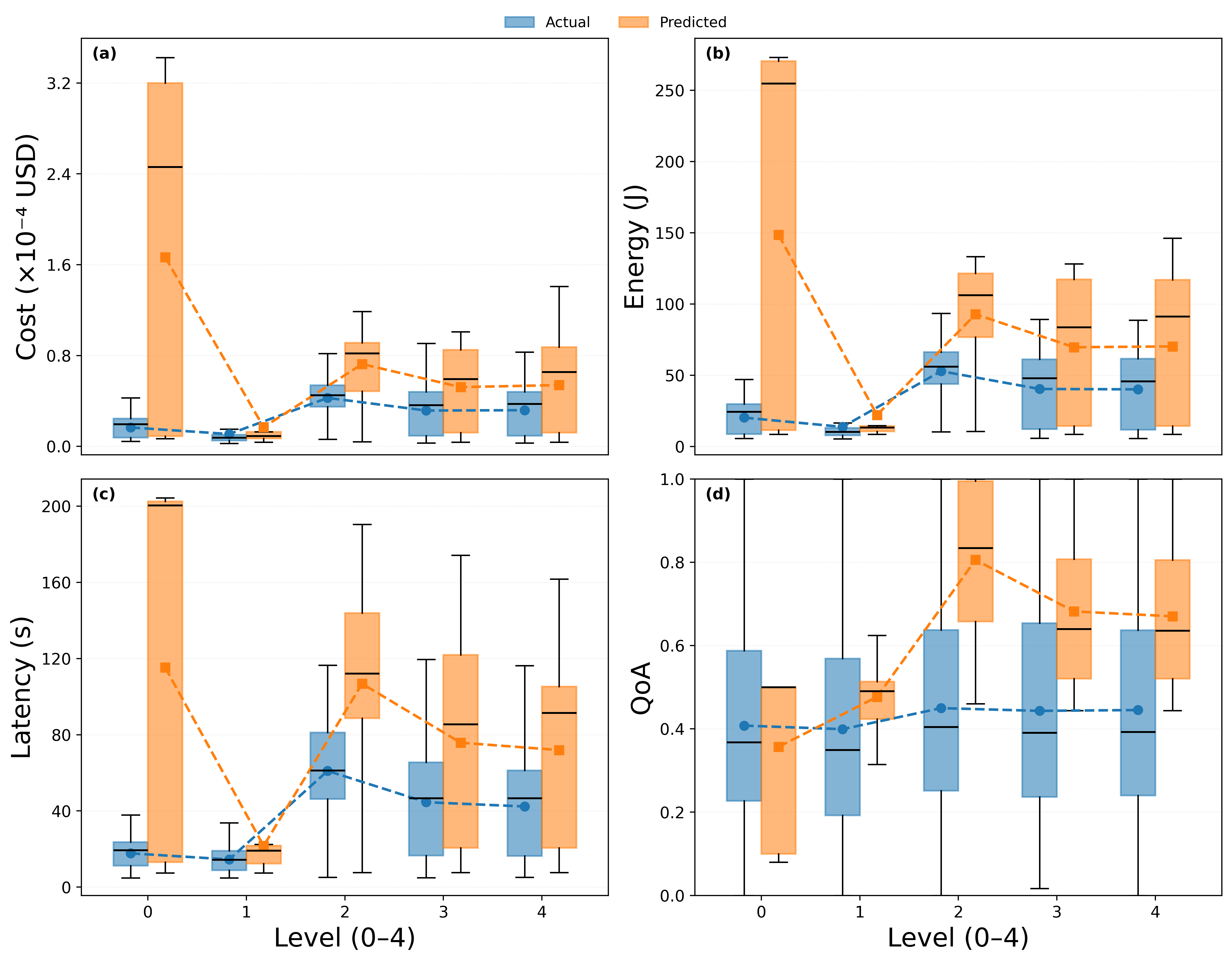}
        \caption{\textit{SimpleQA}}
        \label{fig:fig_lvl_hc_sqa}
    \end{subfigure}
    
    \caption{Actual vs. predicted plan-level cost, energy, latency, and QoA across \textsc{PerfDB} coverage levels (0–4) for Hill Climbing on MMLU-Pro (a) and SimpleQA (b). In both, calibration improves with historical coverage: early overestimation under sparse data (Level 0) diminishes at higher levels, while predictions preserve the level-wise trends needed for planner ranking decisions.}
    \label{fig:fig_lvl_hc_combined}
    \vspace{-0.25cm}
\end{figure}

We next examine whether this level-dependent \emph{calibration} trend, observed for NSGA-II above (Figures~\ref{fig:fig_lvl_nsga_mm} and~\ref{fig:fig_lvl_nsga_sqa}), is robust across alternative planner backends. Figures~\ref{fig:fig_lvl_dp_mm}--\ref{fig:fig_lvl_dp_sqa} and Figures~\ref{fig:fig_lvl_hc_mm}--\ref{fig:fig_lvl_hc_sqa} compare predicted versus actual plan metrics for DP and Hill Climbing across \textsc{PerfDB} coverage levels. In both benchmarks, Level~0 exhibits the largest miscalibration: predicted cost/energy/latency are systematically biased relative to observed distributions, and QoA estimates are overly optimistic due to reliance on static priors. As coverage increases (Levels~1--4), the predicted distributions move closer to the observed ones and track the level-wise trends more faithfully, confirming that empirical traces improve metric calibration beyond NSGA-II. QoA remains the hardest metric to calibrate---especially on the open-ended \textit{SimpleQA} task---but the prediction gap narrows with coverage and, importantly, the estimator preserves the relative ordering across levels, which is sufficient for reliable plan ranking during multi-objective search.
\fi

\ifextended
\subsection{Impact of Model Diversity} 

We evaluate the effect of \emph{LLM diversity} (i.e., the number of distinct models invoked in a plan)  on QoA and resource usage.
Based on the results in Section~\ref{sect:exp_num_operations}, we set the maximum number of operations $K{=}5$, with each operation allowed to call any model from the pool $L$). Since NSGA-II provides the best scalability--quality trade-off among the planners in our earlier comparisons, we use NSGA-II as the default planner for this experiment. We define a plan's diversity level, $d$, as the count of unique models invoked across all operations within it, which can range from $1$ (one LLM used) to $\min(k,|L|)$ (where $k$ is the number of operations and $|L|$ the LLM pool's size).
We report micro-averaged metrics across diversity levels using \emph{Level-4} PerfDB coverage aggregated across all budget settings, and analyze marginal efficiency ratios $\Delta\!\text{QoA}/\Delta\!\text{Financial}$, $\Delta\!\text{QoA}/\Delta\!\text{Energy}$ and $\Delta\!\text{QoA}/\Delta\!\text{Latency}$ to quantify trade-offs (see Table~\ref{tab:delta_metrics} and Figure~\ref{fig:div_box_plots}).

\label{fig:iqr_diversity}
\begin{figure}[htb]
    \centering
    \includegraphics[width=0.9\linewidth]{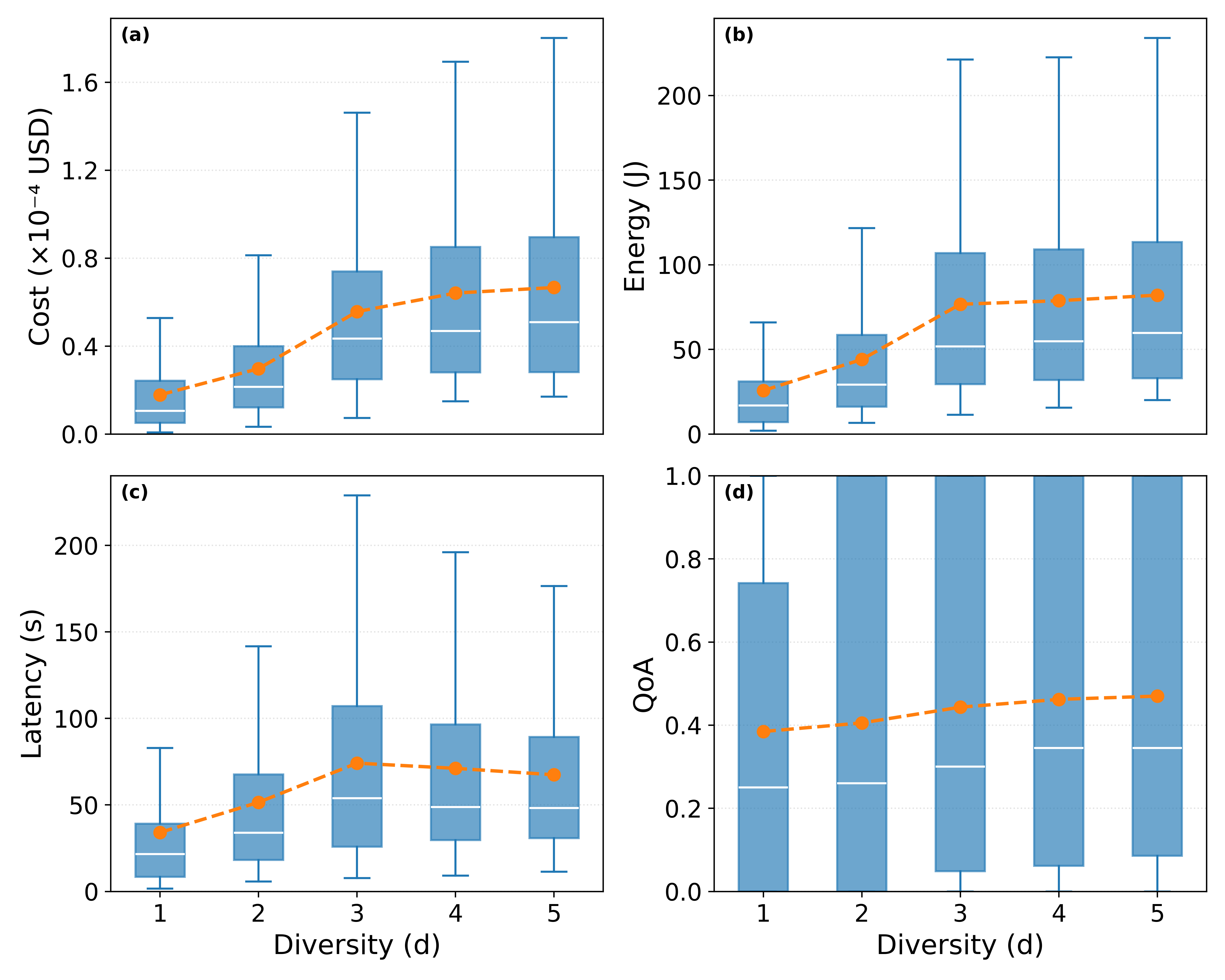}
    \caption{Distribution of QoA across diversity levels.}
    \label{fig:div_box_plots}
\end{figure}

\begin{table}[htb]
\centering
\caption{Incremental changes in QoA, cost, latency, and energy for each step $d{-}1 \to d$. Cost is reported in units of $10^{-5}$ USD.}
\label{tab:delta_metrics}
\scriptsize
\setlength{\tabcolsep}{4pt}
\renewcommand{\arraystretch}{1.05}
\begin{tabular}{lrrrr}
\hline
Metric & $1{\to}2$ & $2{\to}3$ & $3{\to}4$ & $4{\to}5$ \\
\hline
$\Delta$QoA & 0.021 & 0.038 & 0.019 & 0.008 \\
$\Delta$Cost ($10^{-5}$ USD) & 1.20 & 2.60 & 0.85 & 0.25 \\
$\Delta$Latency (s) & 17.419 & 22.634 & -2.995 & -3.732 \\
$\Delta$Energy (J) & 18.373 & 32.601 & 2.099 & 3.287 \\
$\Delta$QoA/$\Delta$Cost (per $10^{-4}$ USD) & 0.175 & 0.146 & 0.223 & 0.314 \\
$\Delta$QoA/$\Delta$Latency (per s) & 0.001 & 0.002 & -0.006 & -0.002 \\
$\Delta$QoA/$\Delta$Energy (per J) & 0.001 & 0.001 & 0.009 & 0.002 \\
\hline
\end{tabular}
\end{table}

Figure~\ref{fig:div_box_plots} shows that the effect of model diversity is non-linear. QoA improves with diversity, with the largest gain from $d{=}2{\to}3$, after which the improvements taper off. Cost and energy increase substantially up to $d=3$ and continue to increase modestly for $d>3$. Latency peaks at $d=3$ and declines thereafter (indicating diminishing QoA returns but improved time efficiency beyond a moderate level of diversity, $d=3$). Marginal efficiency peaks at the $d{=}2{\to}3$ transition for latency and at the $d{=}3{\to}4$ for energy, suggesting that $d\in{3,4}$ balances QoA and resources. Dataset effects also differ. In MMLU-Pro (knowledge-intensive), QoA generally increases with $d$ but at higher resource cost: e.g., from $d=1$ to $d=5$, QoA increased in Business ($0.44{\to}0.54$), Computer Science ($0.30{\to}0.64$), and Physics ($0.22{\to}0.28$), with Biology showing the largest gain ($0.80{\to}0.96$). At the same time, for instance, in \emph{Biology} financial cost, latency, and energy increased 3.2$\times$, 1.2$\times$, and 2.4$\times$, respectively. In contrast, SimpleQA's QoA is flat or declines with increasing $d$ (e.g., \emph{Art} $0.34{\to}0.29$), with only modest gains in a few categories (e.g., \emph{TV Shows} $0.32{\to}0.35$, \emph{Video Games} $0.31{\to}0.42$), which are often outpaced by cost/energy growth ($>3\times$). We next compare sequential and hybrid/parallel plans as model diversity increases. For $d \ge 2$,  hybrid/parallel plans consistently achieve a higher mean QoA than purely sequential plans (e.g., at $d=2$, $0.463$ vs.\ $0.384$), but consume more resources (higher latency and energy usage). Overall, moderate diversity ($d=3$) is a robust operating point, achieving strong QoA with favorable marginal efficiency; increasing diversity to $d \in \{4,5\}$ raises financial cost, latency, and energy without commensurate QoA gains.
\fi
% \begin{table}[htb]
% \centering
% \caption{Incremental changes in QoA, Cost, Latency, and Energy for each step $d-1 \to d$ .}
% \label{tab:delta_metrics}
% \small
% \begin{tabular}{@{}l|rrrr@{}}
% &Step &Step  &Step  &Step  \\
% Metric & (1→2) & (2→3) & (3→4) & (4→5) \\
% \midrule
% $\Delta$QoA & 0.021 & 0.038 & 0.019 & 0.008 \\
% $\Delta$Cost (USD) & 0.000 & 0.000 & 0.000 & 0.000 \\
% $\Delta$Latency (s) & 17.419 & 22.634 & -2.995 & -3.732 \\
% $\Delta$Energy (J) & 18.373 & 32.601 & 2.099 & 3.287 \\
% $\Delta$QoA/$\Delta$Cost (per \$0.0001) & 0.175 & 0.146 & 0.223 & 0.314 \\
% $\Delta$QoA/$\Delta$Latency (per s) & 0.001 & 0.002 & -0.006 & -0.002 \\
% $\Delta$QoA/$\Delta$Energy (per J) & 0.001 & 0.001 & 0.009 & 0.002
% \end{tabular}
% \end{table}

% Overall, model diversity enhances QoA up to $d=3$, where hybrid and parallel plans achieve the best trade-off between quality and resource consumption. More diversity yields diminished QoA returns and higher cost, latency, and energy, particularly in knowledge-intensive datasets, showing the planner’s ability to efficiently adapt the ensemble composition.

\subsection{Impact of Budget}
We evaluate execution under planning-time financial and latency constraints, leaving $E_{\max}$ and $QoA_{\min}$ unconstrained\footnote{We budget only \emph{Financial} and \emph{Latency}—the primary levers for fast-and-cheap interactive QA; QoA is maximized (not budgeted) and Energy closely tracks latency. Focusing on $(F_{\max},L_{\max})$ captures the user-facing trade-off while keeping the analysis simple.}. A plan $\pi$ is considered feasible at planning time when its estimated financial cost and latency satisfy $\mathrm{Financial}(\pi)\le F_{\max}$ and $\mathrm{Latency}(\pi)\le L_{\max}$; the planner may use up to $K{=}5$ operations. Unless noted otherwise, we use NSGA-II as the planning backend in this experiment, since it provides the best scalability--quality trade-off in our planner comparison.
Budgets are instantiated at five levels using the model-averaged anchors \((\$0.000012,19.47\,\mathrm{s})\), and scaled as \((F_{\max}(b),L_{\max}(b))=b\cdot(\$0.000012,19.47\,\mathrm{s})\) for \(b\in\{1,\ldots,5\}\).
The anchor tuple specifies the mean observed financial cost and latency across the five distinct models, establishing a standardized baseline for resource allocation. Budgets are then generated by scaling this composite anchor, ensuring that each constraint is systematically derived from the aggregated typical resource usage of the system. In this setup, \system runs NSGA-II with feasibility dominance and prunes candidates using plan-time estimates of cost and QoA.
For each selected plan, we label it as \emph{budget-adherent} if budget constraints are met after its execution and as an \emph{overrun} otherwise.
For each of the five budgets, we report the \emph{budget-adherence breakdown} (see Table~\ref{tab:budget-summary}) as well as the QoA on budget-adherent runs (see Figure~\ref{fig:budget_figure}).

\begin{table}[htb]
\centering
\scriptsize
\caption{Budget-adherence breakdown per budget level.}
\label{tab:budget-summary}
\begin{tabular}{r|rrrrr}
\textbf{$b$} & \textbf{\makecell{Budget-adherence \\ (\%)} } & \textbf{\makecell{Overrun: \\ Fin. (\%)} } & \textbf{\makecell{Overrun: \\ Lat. (\%)} } & \textbf{\makecell{Overrun: \\ Both (\%)} } \\
\midrule
1 & 96.6 & 2.8 & 0 & 0.6 \\
2 & 91.4 & 2.6 & 3.3 & 2.7 \\
3 & 95.8 & 1.5 & 1.2 & 1.5 \\
4 & 88.8 & 5 & 2.3 & 3.9 \\
5 & 88.6 & 4.3 & 2.5 & 4.6
\end{tabular}
\end{table}

The \emph{adherence rate} remains high across budgets (96.6\% at $b{=}1$ to 88.6\% at $b{=}5$).
Overruns are predominantly cost-driven (1.5--5.0\%), with latency-only overruns at
0.0--3.3\%, and joint cost+latency overruns at 0.6--4.7\%.
Mean QoA on budget-adherent selections shows a clear knee at $b{=}3$ (0.45), with only marginal
changes thereafter (0.41 at $b{=}4$, 0.44 at $b{=}5$), indicating diminishing returns beyond moderate budgets. Dispersion increases with budget (IQR $0.31{\to}0.52$ from $b{=}1{\to}5$), reflecting higher upside but greater variability under looser constraints.
Budget utilization is conservative: budget-adherent plans consume 0.32 -- 0.38 of the financial cost budget and  0.25--0.30 of the latency budget, indicating that the planner typically meets the targets without saturating them. 
Overall, the planner maintains high budget adherence across all levels, with overruns mainly cost-driven. QoA increases sharply at moderate budgets ($b=3$) and then plateaus, while dispersion increases under looser budgets, indicating a trade-off between upside and variability. 
% \begin{figure}[htb]
%     \centering
% \includegraphics[width=1\linewidth]{figures/fig_budget_qoa.png}
%     \caption{QoA vs. budget (\emph{budget-adherent plans only}). }
%     \label{fig:budget_figure}
%     % \vspace{-0.3 cm}
% \end{figure}

\begin{figure}[htb]
\centering
\includegraphics[width=1\linewidth, height=0.42\linewidth, keepaspectratio=false]{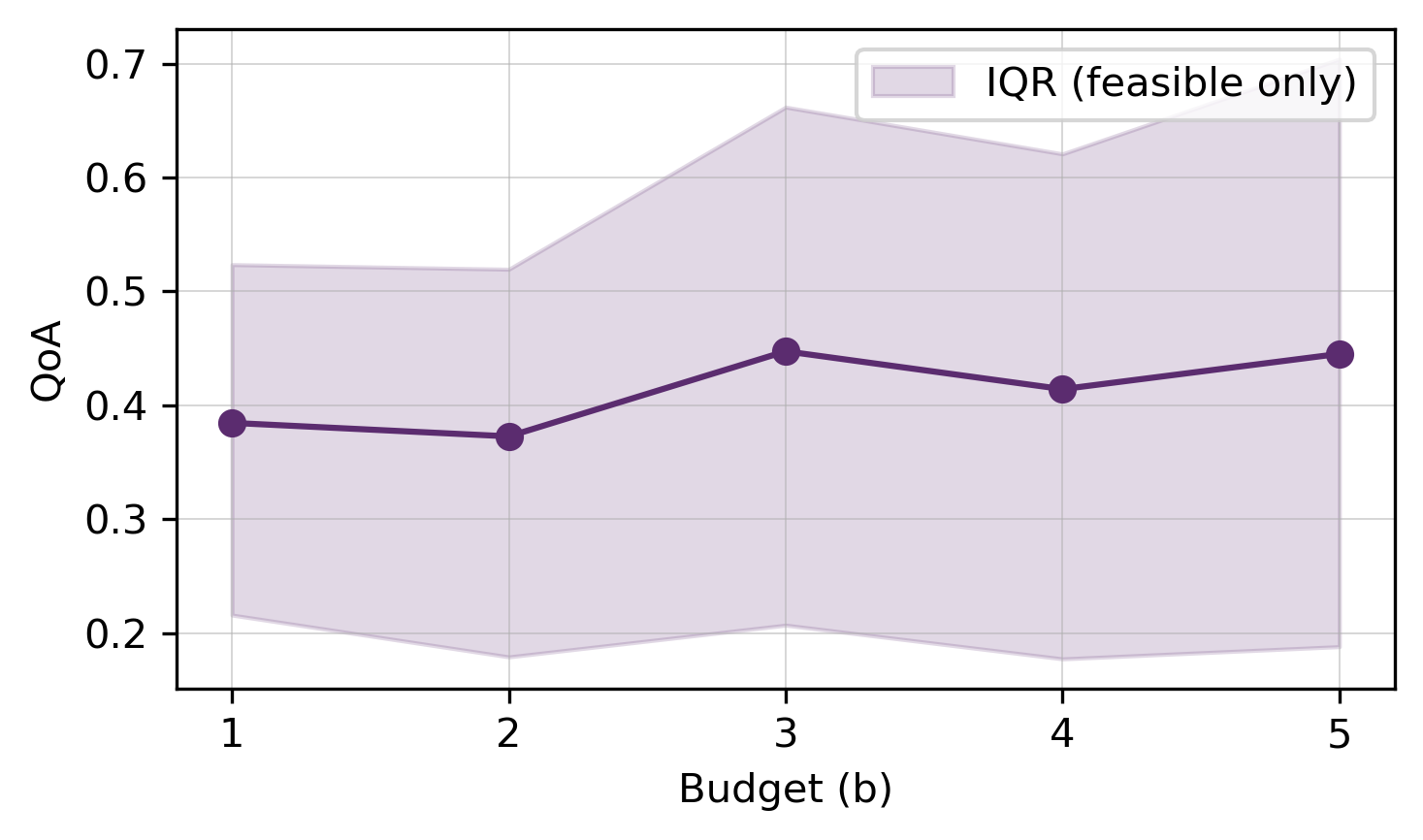}
\caption{QoA vs. budget (\emph{budget-adherent plans only}).}
\vspace{-0.5cm}
\label{fig:budget_figure}
\end{figure}

\subsection{Quality--Cost Trade-offs Against Commercial API Models}
\label{sect:exp_strong_single_model}

We evaluate whether \system remains useful when compared against high-capability state-of-the-art commercial models accessed through their respective provider APIs (i.e., Claude Opus 4.6, GPT 5.4, and Gemini 3.5 Flash).\footnote{API model versions and prices were taken from provider documentation at evaluation time: Anthropic~\url{https://docs.anthropic.com/}, OpenAI~\url{https://platform.openai.com/docs/models}, and Google Gemini~\url{https://ai.google.dev/gemini-api/docs}.}
% These models represent major commercial model families and provide strong single-call reference points with different quality--cost profiles. 
Each commercial model is invoked five times per question using the same benchmark prompt and fixed decoding configuration; we report average QoA and cost across runs. Financial cost is computed from provider-reported token usage and public pricing. We also include the best local single model per benchmark to isolate the benefit of planning over simply choosing the strongest local model. Since commercial APIs run on provider-managed backends with hidden hardware, batching, and serving policies, their latency and energy are not directly comparable to our local models served through Ollama. We therefore compare only consistently measurable metrics across both settings: QoA and financial cost. All methods use the same sampled \textit{MMLU-Pro} and \textit{SimpleQA} questions and the QoA metrics defined earlier.

Table~\ref{tab:strong_single_call_cost} reports the results of the quality-cost comparison. Note that \system provides a substantially cheaper quality--cost operating point than strong commercial API models. Local models (including the ones in \system) where run using Ollama in our server. Hence, the cost is $0$. We report in parenthesis also the cost of running the local models in an external server. Even in that case, the cost compared to commercial LLMs is minimal. W.r.t. quality, on \textit{SimpleQA}, \system achieves higher QoA than Claude Opus~4.6 and GPT~5.4 while costing, if executed in an external server,~$37.8{\times}$ and $14.5{\times}$ less, respectively.  Gemini~3.5 Flash outperforms \system (+$0.14$) but it is~$95.2{\times}$ more expensive. On \textit{MMLU-Pro} we observe similar performance. Hence, \system is complementary to commercial API models. When maximum raw QoA is required, a high-capability single-call model may be preferable, but under cost-sensitive deployment, optimizer-guided local composition provides a substantially cheaper operating point with competitive (if not higher) quality.

\begin{table}[t]
\centering
\caption{Quality--cost comparison against strong commercial API LLM baselines.}
\label{tab:strong_single_call_cost}
\scriptsize
\setlength{\tabcolsep}{3pt}
\renewcommand{\arraystretch}{1.05}
\begin{tabular}{llcc}
\hline
Dataset & Method & QoA & Cost \\
&        &     & ($10^{-4}$ USD) \\
\hline
\textit{SimpleQA} & Claude Opus 4.6 & 0.597 & 9.59 \\
& GPT 5.4 & 0.605 & 3.68 \\
& Gemini 3.5 Flash & 0.790 & 24.18 \\
& Best local single (\emph{LLaMA3-ChatQA (8B)}) & 0.354 & $0$ ($0.071$) \\
& \system (\textsc{NSGA-II}) & 0.65 & $0$ ($0.254$) \\
\hline
\textit{MMLU-Pro} & Claude Opus 4.6 & 0.498 & 18.72 \\
& GPT 5.4 & 0.53 & 8.64 \\
& Gemini 3.5 Flash & 0.871 & 39.85 \\
& Best local single (\emph{Phi-4-14B}) & 0.74 & $0$ ($0.121$) \\
& \system (\textsc{NSGA-II}) & 0.82 & $0$ ($0.909$) \\
\hline
\end{tabular}
\vspace{-0.15cm}
\end{table}

\ifextended
\else

\subsection{Additional Sensitivity Analyses}
Due to space constraints, we report two additional sensitivity analyses in the extended version~\cite{optiqextended2026}. First, we evaluate the impact of model diversity and find that performance gains saturate at moderate diversity: the largest marginal QoA improvement occurs from $d=2$ to $d=3$ distinct models ($\Delta\text{QoA} = 0.038$), whereas the gain from $d=4$ to $d=5$ drops to just $0.008$. Second, we evaluate the impact of the number of operations and find that increasing the number of operations from $k=1$ to $k=5$ improves QoA by $23.6\%$ on SimpleQA and $54.7\%$ on MMLU-Pro; at $k=5$, parallel and hybrid plans account for $82\%$ of the plans selected by Opti-Q.
\fi

% \section{Discussion}
% \label{sect:disc}
% \input{sections/disc}

\section{Conclusion and Future Work}
\label{sect:conclusions}
% We presented \system, a framework for optimizing multi-LLM QA via a multi-objective formulation. \system systematically constructs execution plans that balance QoA, cost, latency, and energy under explicit user-defined budget and constraints. Our formulation unifies sequential, parallel, and hybrid orchestration strategies within a single optimization space, enabling plan evaluation prior to execution. \system’s ``pluggable optimization engine'', guided by a statistical performance database efficiently estimates the cost of the plan and the QoA while exploring a large search space with a tractable overhead. Across QA benchmarks, \system consistently achieves higher accuracy at a comparable cost relative to baselines, demonstrating that structured optimization can outperform heuristic or static orchestration. 
% % Reducing planning overhead was not the focus of this paper.
% In future work, we aim to explore mechanisms, such as caching previous results, to minimize the cost of planning while maximizing the estimated performance of the selected plans.
We presented \system, a database-inspired optimizer for per-question multi-LLM QA planning. \system represents LLM invocations as physical operators in a DAG, estimates QoA and resource usage using \textsc{PerfDB}, and searches for Pareto-efficient sequential, parallel, and hybrid plans under user-specified constraints. Across MMLU-Pro and SimpleQA, \system achieves higher QoA at comparable cost than strong routing, cascading, and ensembling baselines, demonstrating that plan-before-execute optimization can improve quality--resource trade-offs in multi-LLM QA. These results show that structured, statistics-driven planning provides a practical foundation for adaptive LLM orchestration as model choices, budgets, and deployment backends continue to diversify.

A natural next step is to extend the same plan-before-execute abstraction beyond QA to richer RAG and agentic workflows. Because such pipelines can be expressed as DAGs over a broader operator vocabulary (e.g., retrieval, reranking, verification, and tool execution alongside generation and blending) each carrying its own quality and resource profile in \textsc{PerfDB}, \system's planning machinery should transfer with little conceptual change. Realizing this will require characterizing these new operators' cost--benefit behavior and validating that the estimators remain reliable as plans grow longer and more heterogeneous, which we leave to future work.

% \section{Acknowledgments}
% Identification of funding sources and other support, and thanks to
% individuals and groups that assisted in the research and the
% preparation of the work should be included in an acknowledgment
% section, which is placed just before the reference section in your
% document.

\bibliographystyle{IEEEtran}
\bibliography{IEEEabrv,bibliography}

\clearpage

% \appendix
% \input{sections/appendix} 

\end{document}
\endinput
%%
%% End of file `sample-sigconf.tex'.